\documentclass[journal]{IEEEtran}

\usepackage[utf8]{inputenc} 
\usepackage[T1]{fontenc}    
\usepackage{hyperref}       
\usepackage{url}            
\usepackage{booktabs}       
\usepackage{amsfonts}       
\usepackage{nicefrac}       
\usepackage{microtype}      
\usepackage{xcolor}         
\usepackage{graphicx}
\usepackage{wrapfig}
\usepackage{cite}

\usepackage{balance}
\usepackage{adjustbox}

\title{Is it enough to optimize CNN architectures on ImageNet?}

\author{%
  Lukas Tuggener$^{1,2}$, Jürgen Schmidhuber$^{2,3,4}$, Thilo Stadelmann$^{1,5}$ \\
  $^1$ZHAW Centre for Artificial Intelligence, Winterthur, Switzerland, $^2$University of Lugano, Switzerland, \\
  $^3$The Swiss AI Lab IDSIA, Switzerland, $^4$ AI Initiative, King Abdullah University of Science and Technology (KAUST), Thuwal, Saudi Arabia, $^5$Fellow of the European Centre for Living Technology, Venice, Italy\\
  \texttt{\{tugg,stdm\}@zhaw.ch, juergen@idsia.ch} \\
}

\begin{document}

\maketitle

\begin{abstract}


Classification performance based on ImageNet is the de-facto standard metric for CNN development. In this work we challenge the notion that CNN architecture design solely based on ImageNet leads to generally effective convolutional neural network (CNN) architectures that perform well on a diverse set of datasets and application domains. To this end, we investigate and ultimately improve ImageNet as a basis for deriving such architectures. We conduct an extensive empirical study for which we train $500$ CNN architectures, sampled from the broad AnyNetX design space, on ImageNet as well as $8$ additional well known image classification benchmark datasets from a diverse array of application domains. We observe that the performances of the architectures are highly \emph{dataset dependent}. Some datasets even exhibit a negative error correlation with ImageNet across all architectures. We show how to significantly increase these correlations by \emph{utilizing ImageNet subsets restricted to fewer classes}. These contributions can have a profound impact on the way we design future CNN architectures and help alleviate the tilt we see currently in our community with respect to over-reliance on one dataset.
\end{abstract}


\section{Introduction}
\label{chp:experiments}

\label{chp:intro}
Deep convolutional neural networks (CNNs) are the core building block for most modern visual recognition systems and lead to major breakthroughs in many domains of computer perception in the past several years. Therefore, the community has been searching the high dimensional space of possible network architectures for models with desirable properties. Important milestones such as DanNet \cite{DBLP:conf/cvpr/CiresanMS12}, AlexNet  \cite{DBLP:conf/nips/KrizhevskySH12}, VGG  \cite{DBLP:journals/corr/SimonyanZ14a}, HighwayNet \cite{DBLP:journals/corr/SrivastavaGS15}, and ResNet \cite{DBLP:conf/cvpr/HeZRS16} (a HighwayNet with open gates) can be seen as update steps in this stochastic optimization problem and stand testament that the manual architecture search works. It is of great importance that the right metrics are used during the search for new neural network architectures. Only when we measure performance with a truly meaningful metric is it certain that a new high-scoring architecture is also fundamentally better. So far, the metric of choice in the community has often been the performance on the most well-known benchmarking dataset---ImageNet \cite{DBLP:journals/corr/RussakovskyDSKSMHKKBBF14}.

More specifically, it would be desirable to construct such a metric from a solid theoretical understanding of deep CNNs. Due to the absence of a solid theoretical basis novel neural network designs are tested in an empirical fashion. Traditionally, model performance has been judged using accuracy point estimates \cite{DBLP:conf/nips/KrizhevskySH12, DBLP:conf/eccv/ZeilerF14,DBLP:journals/corr/SimonyanZ14a}. This simple measure ignores important aspects such as model complexity and speed. Newer work addresses this issue by reporting a curve of the accuracy at different complexity settings of the model, highlighting how well a design deals with the accuracy versus complexity tradeoff \cite{DBLP:conf/cvpr/XieGDTH17, DBLP:conf/cvpr/ZophVSL18}. 

Very recent work strives to improve the quality of the empiric evaluation even further. There have been attempts to use extensive empirical studies to discover general rules on neural network design \cite{DBLP:journals/corr/abs-1712-00409,DBLP:conf/iclr/RosenfeldRBS20, DBLP:journals/corr/abs-2001-08361,tuggener2020design}, instead of simply showing the merits of a single neural network architecture. Another line of research aims to improve empiricism by sampling whole populations of models and comparing error distributions instead of individual scalar errors  \cite{DBLP:conf/iccv/Radosavovic0XLD19}. 

\begin{figure}
    \centering
    \includegraphics[width=0.45\textwidth]{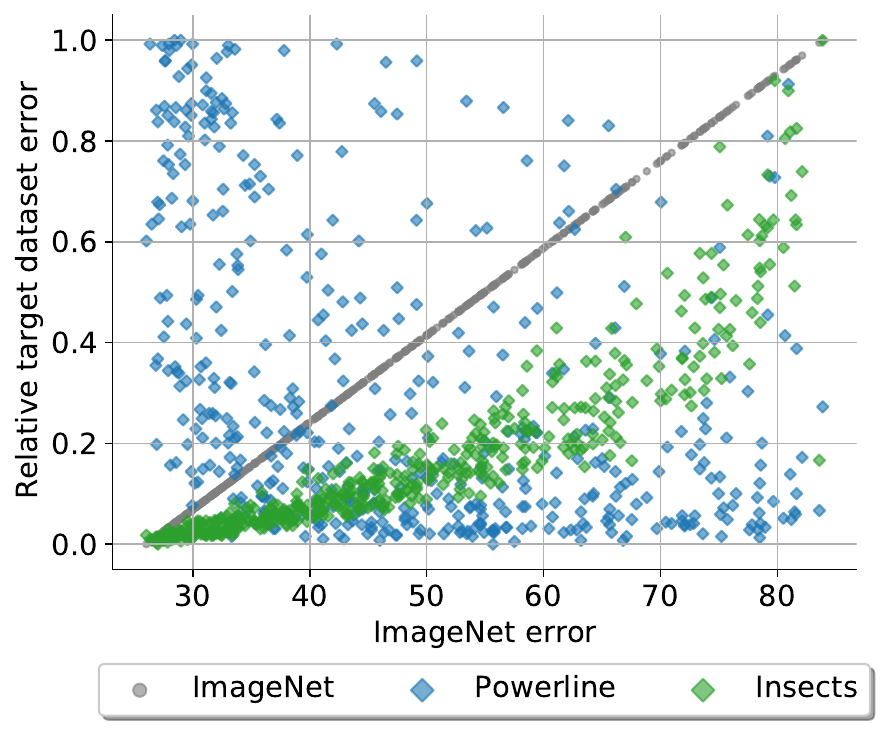} 
    \caption{Is a CNN \emph{architecture} that performs well on ImageNet automatically a good choice for a different vision dataset? This plot suggests otherwise: It displays the relative test errors of $500$ randomly sampled CNN architectures on three datasets (ImageNet, Powerline, and Insects) plotted against the test error of the same architectures on ImageNet. The architectures have been trained from scratch on all three datasets. Architectures with low errors on ImageNet also perform well on Insects, on Powerline the opposite is the case.} 
    \label{fig:intro}
\end{figure}

We acknowledge the importance of the above-mentioned improvements in the empirical methods used to test neural networks, but identify a weak spot that runs trough the above-mentioned work: the heavy reliance on ImageNet \cite{DBLP:journals/corr/RussakovskyDSKSMHKKBBF14} (and to some extent the very similar Cifar100 \cite{krizhevsky2009learning}). In 2011, Torralba and Efros already pointed out that visual recognition datasets that were built to represent the visual world tend to become a small world in themselves \cite{DBLP:conf/cvpr/TorralbaE11}. Objects are no longer in the dataset because they are important, they are important because they are in the dataset. 
\emph{In this paper, we investigate how well ImageNet represents a diverse set of visual classification datasets---and present methods to improve said representation, such that CNN architectures optimized on ImageNet become more effective on visual classification beyond ImageNet.} Specifically, our contributions are: (a) an extensive empirical study showcasing the fitness of ImageNet as a basis for generally effective CNN architectures; (b) we show how class-wise subsampled versions of ImageNet in conjunction with the original datasets yield a $2.5$-fold improvement in average error correlations with other datasets (c) we identify cumulative block depth and width as the architecture parameters most sensitive to changing datasets.


As a tool for this investigation we introduce the notion of architecture and performance relationship (APR). The  performance of a CNN architecture does not exist in a vacuum, it is only defined in relation to the dataset on which it is used. This dependency is what we call APR induced by a dataset. 
We study the change in APRs between datasets by sampling $500$ neural network architectures and training all of them on a set of datasets\footnote{Since we only sample models in the complexity regime of $340$ mega flops (MF) to $400$MF (ResNet-152 has $11.5$GF) we could complete the necessary $7500$ model trainings within a moderate $85$ GPU days on Tesla V100-SXM2-32GB GPUs.}. We then compare errors of the same architectures across datasets, revealing the changes in APR (see Figure \ref{fig:intro}). This approach allows us to study the APRs induced by different datasets on a whole population of diverse network designs rather than just a family of similar architectures such as the ResNets \cite{DBLP:conf/cvpr/HeZRS16} or MobileNets \cite{DBLP:journals/corr/HowardZCKWWAA17}. 


All of our code, sampled architectures and additional figures are available at \url{https://github.com/tuggeluk/pycls/tree/tnnls}.

\section{Related Work}
\label{chp:related}

\textbf{Neural network design.}
With the introduction of the first deep CNNs \cite{DBLP:conf/cvpr/CiresanMS12,DBLP:conf/nips/KrizhevskySH12} the design of neural networks immediately became an active research area. In the following years many improved architectures where introduced, such as VGG \cite{DBLP:journals/corr/SimonyanZ14a}, Inception \cite{DBLP:conf/cvpr/SzegedyLJSRAEVR15}, HighwayNet \cite{DBLP:journals/corr/SrivastavaGS15}, ResNet \cite{DBLP:conf/cvpr/HeZRS16} (a HighwayNet with open gates), ResNeXt \cite{DBLP:conf/cvpr/XieGDTH17}, or MobileNet \cite{DBLP:journals/corr/HowardZCKWWAA17}. These architectures are the result of manual search aimed at finding new design principles that improve performance, for example increased network depth and skip connections. More recently, reinforcement learning \cite{DBLP:conf/cvpr/ZophVSL18}, evolutionary algorithms \cite{DBLP:conf/aaai/RealAHL19} or gradient descent \cite{DBLP:conf/iclr/LiuSY19} have been successfully used to find suitable network architectures automatically. Our work relates to manual and automatic architecture design because it adds perspective on how stable results based on one or a few datasets are. 

\textbf{Empirical studies.}
In the absence of a solid theoretical understanding, large-scale empirical studies are the best tool at our disposal to gain insight into the nature of deep neural networks. These studies can aid network design \cite{DBLP:journals/tnn/GreffSKSS17, DBLP:conf/iclr/CollinsSS17, DBLP:conf/iclr/NovakBAPS18} or be employed to show the merits of different approaches, for example that the classic LSTM \cite{DBLP:journals/neco/HochreiterS97} architecture can outperform more modern models \cite{DBLP:conf/iclr/MelisDB18}, when it is properly regularised. More recently, empirical studies have been used to infer more general rules on the behaviour of neural networks such as a power-law describing the relationship between generalization error and dataset size \cite{DBLP:journals/corr/abs-1712-00409} or scaling laws for neural language models \cite{DBLP:journals/corr/abs-2001-08361}.

\textbf{Generalization in neural networks.}
Despite their vast size have deep neural networks shown in practice that they can generalize extraordinarily well to unseen data stemming from the same distribution as the training data. Why neural networks generalize so well is still an open and very active research area \cite{DBLP:journals/corr/abs-1710-05468,DBLP:conf/icml/DinhPBB17,DBLP:conf/iclr/ZhangBHRV17}. 
This work is not concerned with the generalization of a trained network to new data, but with the generalization of the architecture design progress itself. Does an architecture designed for a certain dataset, e.g. natural photo classification using ImageNet, work just as well for medical imaging? There has been work investigating the generalization to a newly collected test set, but in this case the test set was designed to be of the same distribution as the original training data \cite{DBLP:conf/icml/RechtRSS19}.

\textbf{Neural network transferability}
It is known that the best architecture for ImageNet is not necessarily the best  base architecture for other applications such as semantic segmentation \cite{DBLP:conf/cvpr/LongSD15} or object detection \cite{DBLP:conf/nips/ChenYZMXS19}. Researchers who computed a taxonomy of multiple visions tasks identified that the simmilarities between tasks did not depend on the used architecture \cite{DBLP:conf/ijcai/ZamirSSGMS19}.
Research that investigates the relation between model performance on ImageNet and new classification datasets in the context of transfer learning \cite{DBLP:conf/cvpr/RazavianASC14, DBLP:conf/icml/DonahueJVHZTD14} suggests that there is a strong correlation which is also heavily dependent on the training regime used \cite{DBLP:conf/cvpr/KornblithSL19}. 
Our work differs form the ones mentioned above in that we are not interested in the transfer of learned features but transfer of the architecture designs and therefore we train our networks from scratch on each dataset. Moreover do we not only test transferability on a few select architectures but on a whole network space.

\textbf{Neural network design space analysis.}
Radosavovic et al. \cite{DBLP:conf/iccv/Radosavovic0XLD19} introduced network design spaces for visual recognition. They define a design space as a set of architectures defined in a parametric form with a fixed base structure and architectural hyperparameters that can be varied, similar to the search space definition in neural architecture search \cite{DBLP:conf/cvpr/ZophVSL18,DBLP:conf/aaai/RealAHL19,DBLP:conf/iclr/LiuSY19}. The error distribution of a given design space can be computed by randomly sampling model instances from it and computing their training error. We use a similar methodology but instead of comparing different design spaces, we compare the results of the same design space on different datasets.

\section{Datasets}
\label{chp:data}
\begin{figure}
    \centering
    \includegraphics[width=0.98\columnwidth]{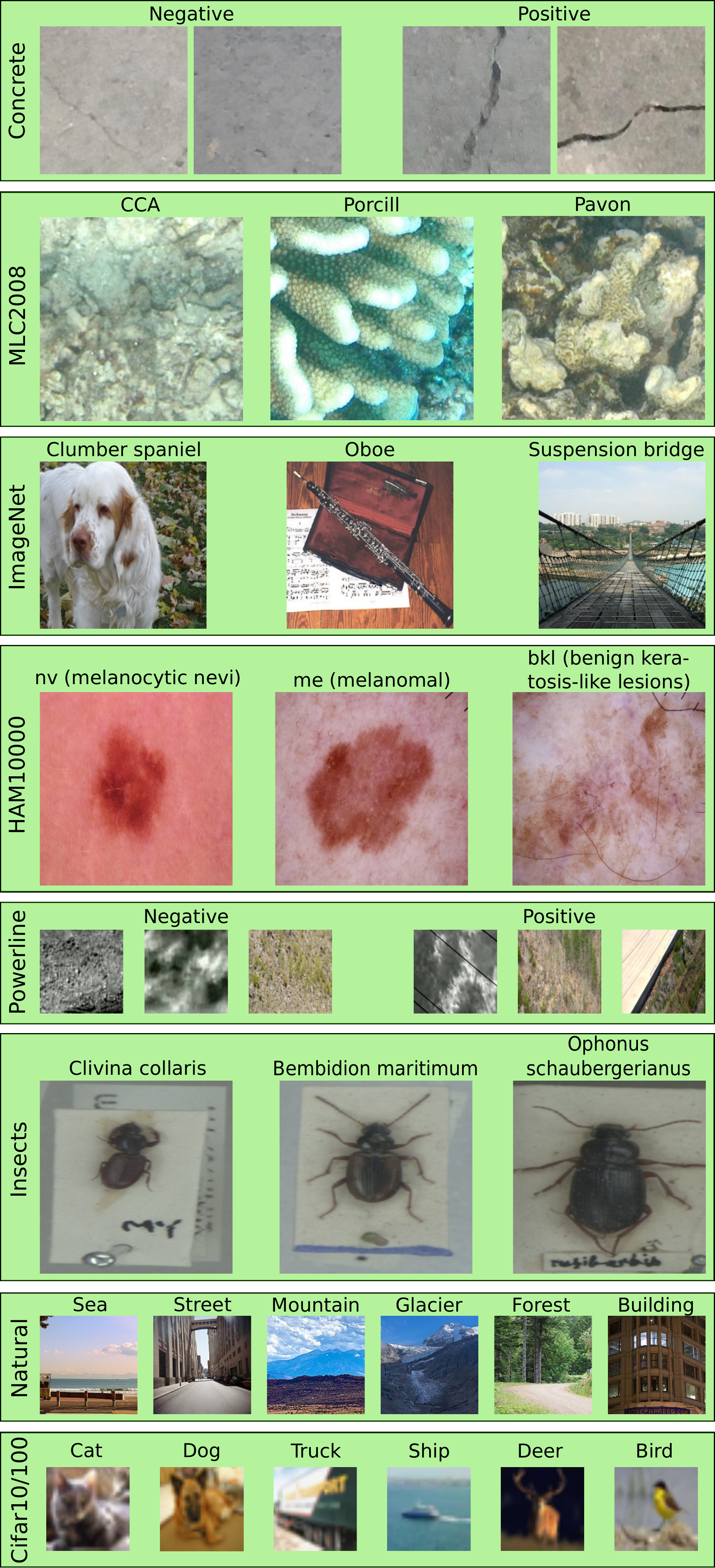}
    \caption{Example images from each dataset. Images of Cifar10/100 are magnified fourfold, the rest are shown in their original resolution (best viewed by zooming into the digital document).
    \label{fig:data_examples}}
\end{figure}

To enable cross dataset comparison of APRs we assembled a corpus of datasets. We chose datasets according to the following principles: (a) include datasets from a wide spectrum of application areas, such that generalization is tested on a diverse set of datasets; (b) only use datasets that are publicly available to anyone to ensure easy reproducibility of our work. Figure \ref{fig:data_examples} shows examples and Table \ref{tbl:datasets} lists meta-data of the chosen datasets. More detailed dataset specific information is given in the remainder of this chapter.

\begin{table}
    \caption{Meta data of the used datasets.}
    \label{tbl:datasets}
    \begin{center}
        \begin{small}
            \begin{sc}
                \begin{tabular}{lcccc}
                    \toprule
                    Dataset & No. Images & No. Classes & Img. Size  \\
                    \midrule
                    Concrete   & $40$K & $2$ & $227 \times 227$ \\
                    MLC2008    & $43$K & $9$ & $312 \times 312$ \\
                    ImageNet   & $1.3$M & $1000$ & $256 \times 256$  \\
                    HAM10000   & $10$K & $7$ & $296 \times 296$ \\
                    Powerline  & $8$K &  $2$ & $128 \times 128$ \\
                    Insects    & $63$K & $291$ & $296 \times 296$\\
                    Natural    & $25$k & $6$ & $150 \times 150$ \\
                    Cifar10    & $60$k & $10$ & $32 \times 32$ \\
                    Cifar100   & $60$k & $100$ & $32 \times 32$\\
                    \bottomrule
                \end{tabular}
            \end{sc}
        \end{small}
      
    \end{center}
    \vskip -0.4in
\end{table}

\textbf{Concrete}
\cite{ozgenel2018performance} contains $40$ thousand image snippets produced from $458$ high-resolution images that have been captured from various concrete buildings on a single campus. It contains two classes, positive (which contains cracks in the concrete) and negative (with images that show intact concrete). With $20$ thousand images in both classes the dataset is perfectly balanced.

\textbf{MLC2008}
\cite{DBLP:journals/remotesensing/ShihavuddinGGGG13} contains $43$ thousand image snippets taken form the MLC dataset \cite{DBLP:conf/cvpr/BeijbomEKMK12}, which is a subset of the images collected at the Moorea Coral Reef Long Term Ecological Research site. It contains images from three reef habitats and has nine classes. The class distribution is very skewed with crustose coralline algae (CCA) being the most common by far (see Figure \ref{fig:class_distributions} in Appendix \ref{chp:robustness}).

\textbf{ImageNet}
\cite{DBLP:journals/corr/RussakovskyDSKSMHKKBBF14} (ILSVRC 2012) is a large scale dataset containing $1.3$ million photographs sourced from flickr and other search engines. It contains $1000$ classes and is well balanced with almost all classes having exactly $1300$ training and $50$ validation samples.

\textbf{HAM10000}
\cite{DBLP:journals/corr/abs-1803-10417} ("Human Against Machine with $10000$ training images") is comprised of dermatoscopic images, collected from different populations and by varied modalities. It is a representative collection of all important categories of pigmented lesions that are categorized into seven classes. It is imbalanced with an extreme dominance of the  melanocytic nevi (nv) class (see Figure \ref{fig:class_distributions} in Appendix \ref{chp:robustness}).

\textbf{Powerline}
\cite{yetgin2017ground} contains images taken in different seasons as well as weather conditions from $21$ different regions in Turkey. It has two classes, positive (that contain powerlines) and negative (which do not). The dataset contains $8000$ images and is balanced with $4000$ samples per classes. Both classes contain $2000$ visible-light and $2000$ infrared images.

\textbf{Insects}
\cite{hansen_oskar_liset_pryds_2019_3549369} contains $63$ thousand images of $291$ insect species. The images have been taken of the collection of British carabids from the Natural History Museum London. The dataset is not completely balanced but the majority of classes have $100$ to $400$ examples.

\textbf{Intel Image Classification}
\cite{natural} dataset (``natural'') is a natural scene classification dataset containing $25$ thousand images and $6$ classes. It is very well balanced with all classes having between $2.1$ thousand and $2.5$ thousand samples.

\textbf{Cifar10 and Cifar100}
\cite{krizhevsky2009learning} both consist of $60$ thousand images. The images are sourced form the $80$ million tiny images dataset \cite{DBLP:journals/pami/TorralbaFF08} and are therefore of similar nature (photographs of common objects) as the images found in ImageNet, bar the much smaller resolution. Cifar10 has $10$ classes with $6000$ images per class, Cifar100 consists of $600$ images in $100$ classes, making both datasets perfectly balanced.



\newpage
\section{Experiments and Results}
\subsection{Experimental setup}
\label{chp:experiments}
\begin{figure}
    \centering
    \includegraphics[width=1\columnwidth]{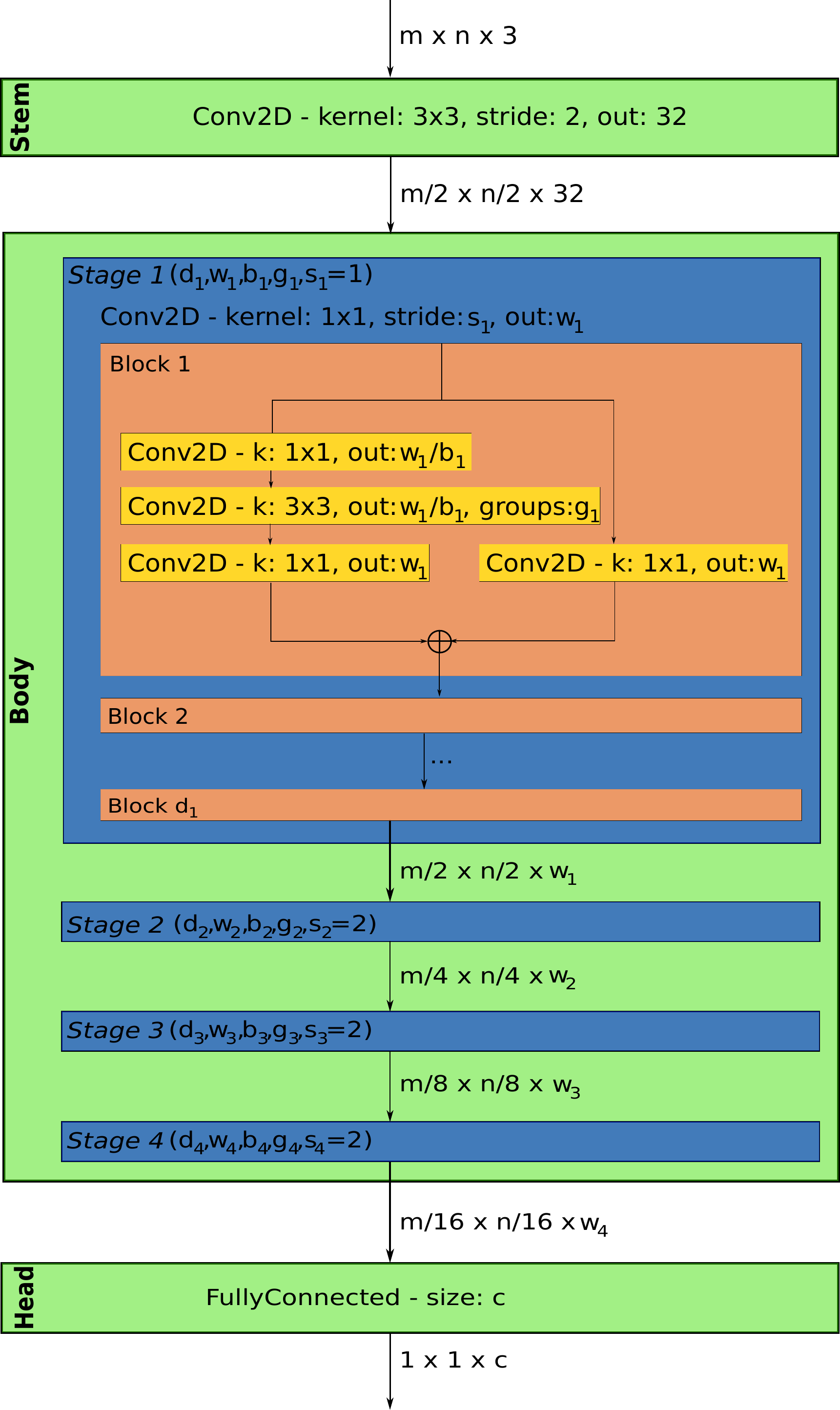}
    \caption{The structure of models in the AnyNetX design space, with a fixed stem and a head, consisting of one fully-connected layer of size $c$, (where c is the number of classes). Each stage $i$ of the body is parametrised by $d_i,w_i,b_i,g_i$, the strides of the stages are fixed with $s_1=1$ and $s_i=2$ for the remainder.}
    \label{fig:AnyNetX}
\end{figure}
We sample our architectures form the very general AnyNetX \cite{DBLP:conf/cvpr/RadosavovicKGHD20} parametric network space. The networks in AnyNetX consist of a stem, a body, and a head. The body performs the majority of the computation, stem and head are kept fixed across all sampled models. The body consists of four stages, each stage $i$ starts with a $1 \times 1$ convolution with stride $s_i$, the remainder is a sequence of $d_i$ identical blocks. The blocks are standard residual bottleneck blocks with group convolution \cite{DBLP:conf/cvpr/XieGDTH17}, with a total block width $w_i$, bottleneck ratio $b_i$ and a group width $g_i$ (into how many parallel convolutions the total width is grouped into). Within a stage, all the block parameters are shared. See Figure \ref{fig:AnyNetX} for a comprehensive schematic.

The AnyNetX design space has a total of $16$ degrees of freedom, having $4$ stages with $4$ parameters each. We obtain our model instances by performing log-uniform sampling of $d_i \leq 16$, $w_i \leq 1024$ and divisible by $8$, $b_i \in {1,2,4}$, and $g_i \in  {1,2, ..., 32}$. The stride $s_i$ is fixed with a stride of $1$ for the first stage and a stride of $2$ for the rest. We repeatedly draw samples until we have obtained a total of $500$ architectures in our target complexity regime of $360$ mega flops (MF) to $400$ MF. We use a very basic training regime that consists of only flipping and cropping of the inputs in conjunction with SGD plus momentum and weight decay. 

The same $500$ models are trained on each dataset until the loss is reasonably saturated. The exact number of epochs has been determined in preliminary experiments and depends on the dataset (see Table \ref{tbl:training}). For extensive ablation studies ensuring the empirical stability of our experiments with respect to Cifar10 performance, training duration, training variability, top-1 to top-5 error comparisons, overfitting and class distribution see chapters \ref{chp:cifar_stability} to \ref{chp:ds_distrib} in Appendix \ref{chp:robustness}. Supplementary results on the effect of pretraining and the structure of the best performing architectures can be found in chapters \ref{chp:pretrain} and \ref{chp:top_arch} in Appendix \ref{chp:ablation}.

\begin{table}[t]
    \caption{Dataset-specific experimental settings.}
    \label{tbl:training}
    \begin{center}
        \begin{small}
            \begin{sc}
                \begin{tabular}{lcccr}
                    \toprule
                    Dataset & No. training epochs & Eval. error  \\
                    \midrule
                    concrete   & $20$ & top-1\\
                    MLC2008   & $20$ & top-1\\
                    Imagenet   & $10$ & top-5\\
                    HAM10000   & $30$ & top-1\\
                    Powerline  & $20$ & top-1\\
                    Insects    & $20$ & top-5\\
                    Natural    & $20$ & top-1\\
                    Cifar10    & $30$ & top-1\\
                    Cifar100   & $30$ & top-5\\
                    \bottomrule
                \end{tabular}
            \end{sc}
        \end{small}
    \end{center}
\end{table}

\subsection{Experimental results}

\begin{figure}[h]
    \centering
    \includegraphics[width=0.22\textwidth]{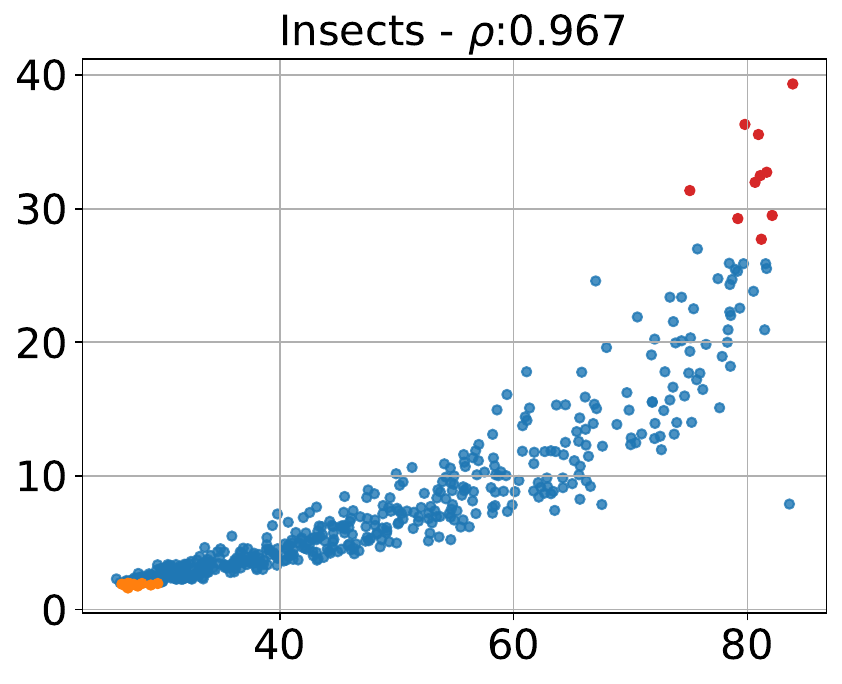}
    \includegraphics[width=0.22\textwidth]{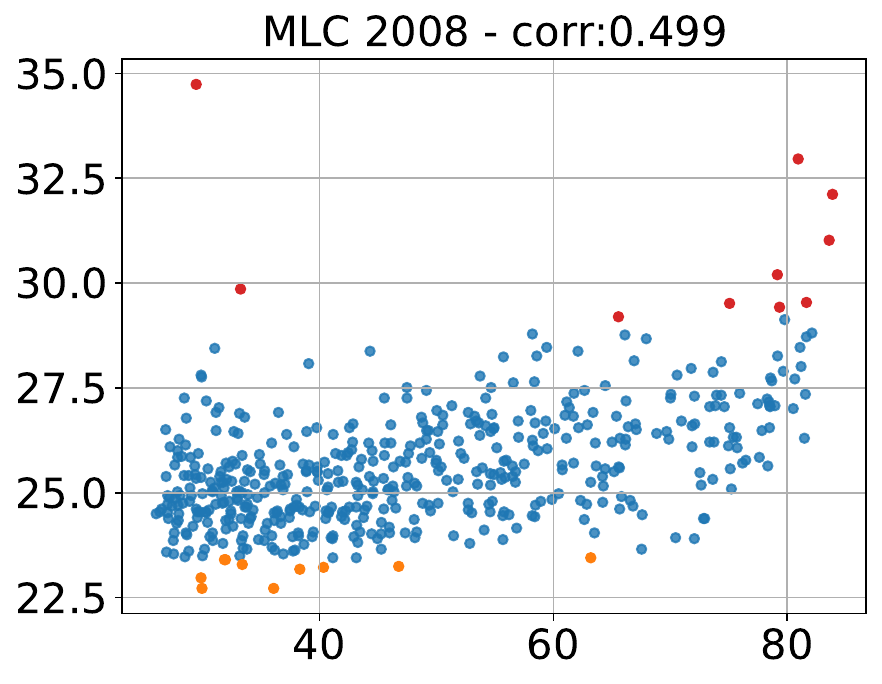} 
    \includegraphics[width=0.22\textwidth]{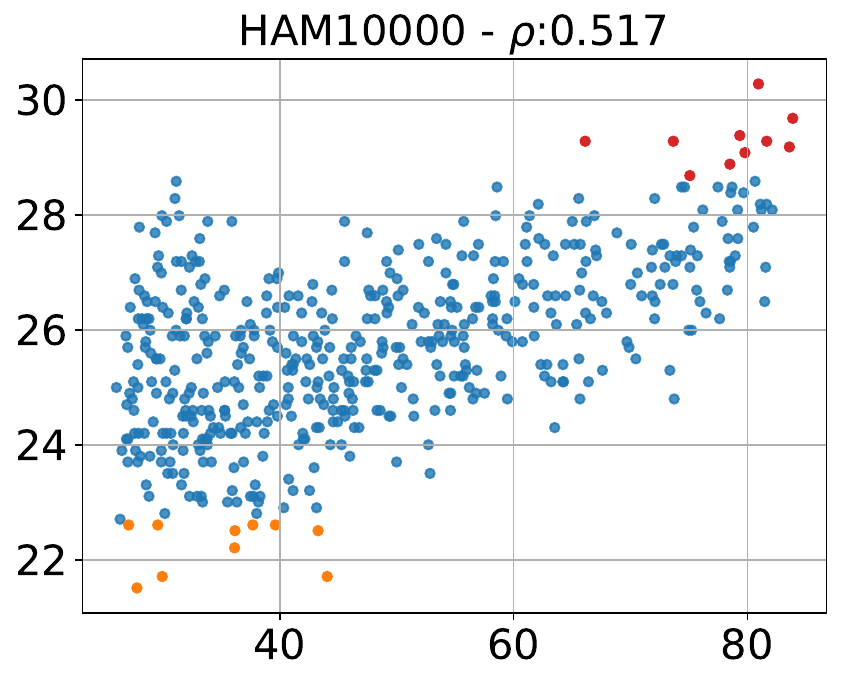}
    \includegraphics[width=0.22\textwidth]{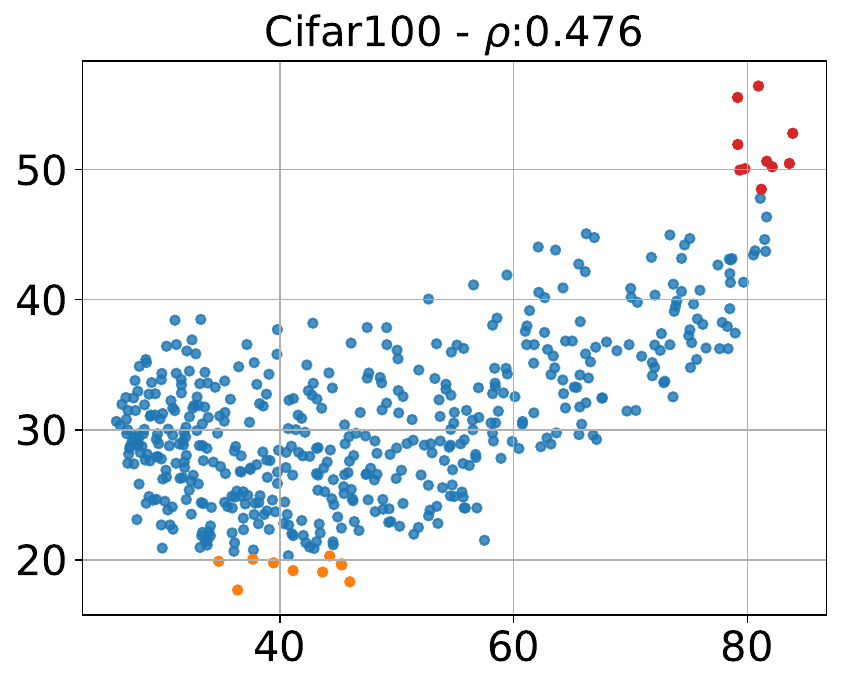} 
    \\
    \ 
     \includegraphics[width=0.22\textwidth]{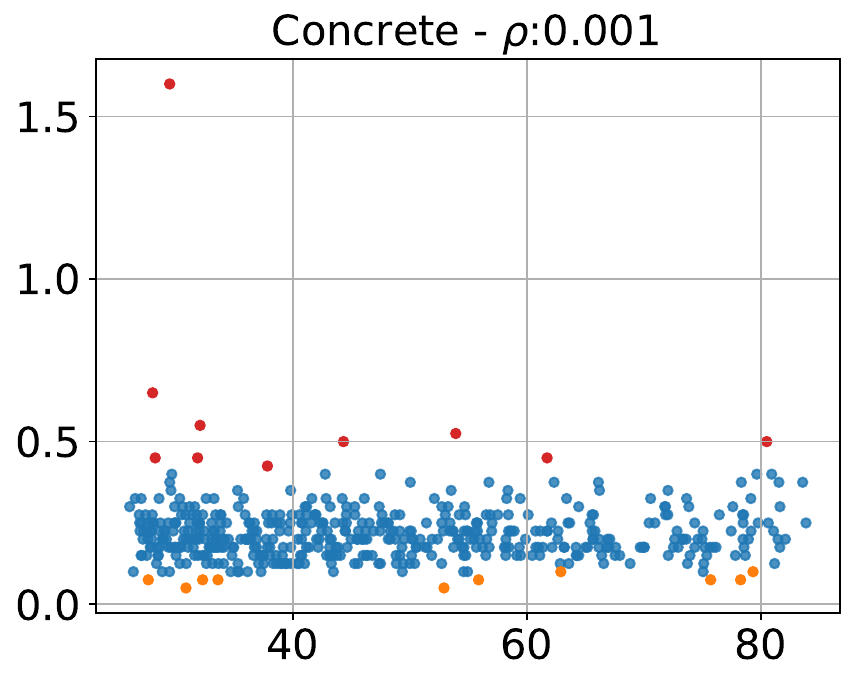} 
    \includegraphics[width=0.22\textwidth]{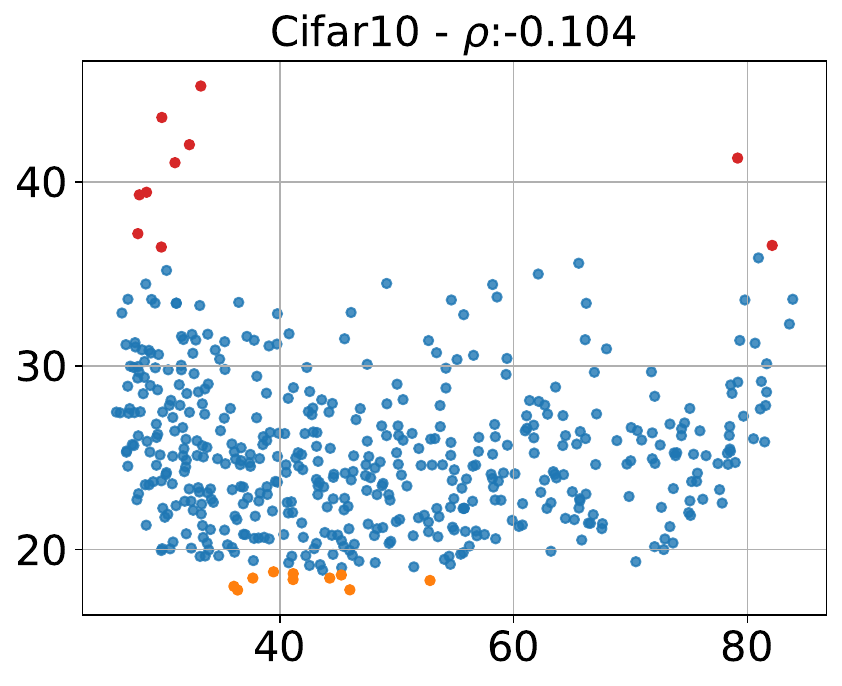} 
    \includegraphics[width=0.22\textwidth]{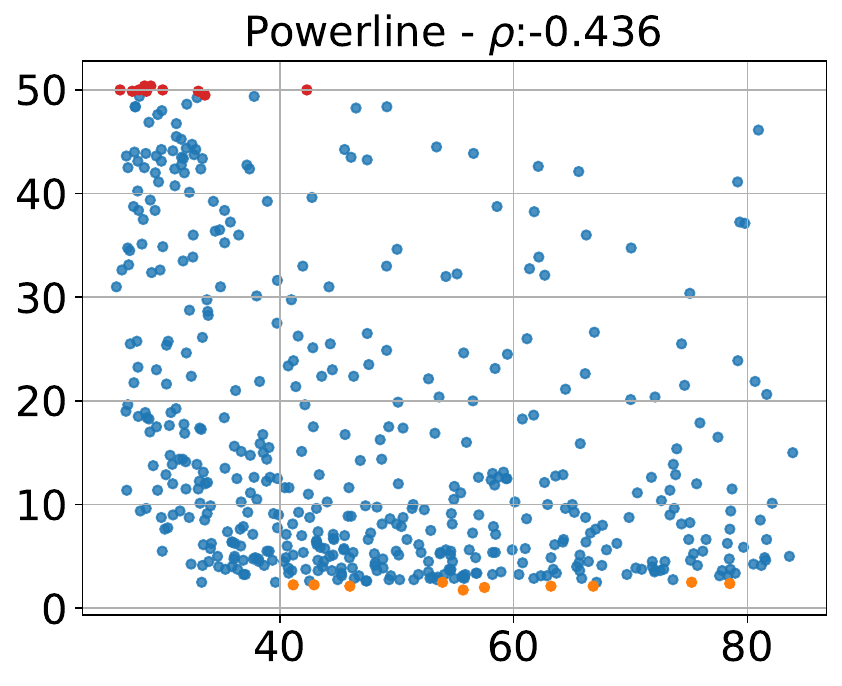}
    \includegraphics[width=0.22\textwidth]{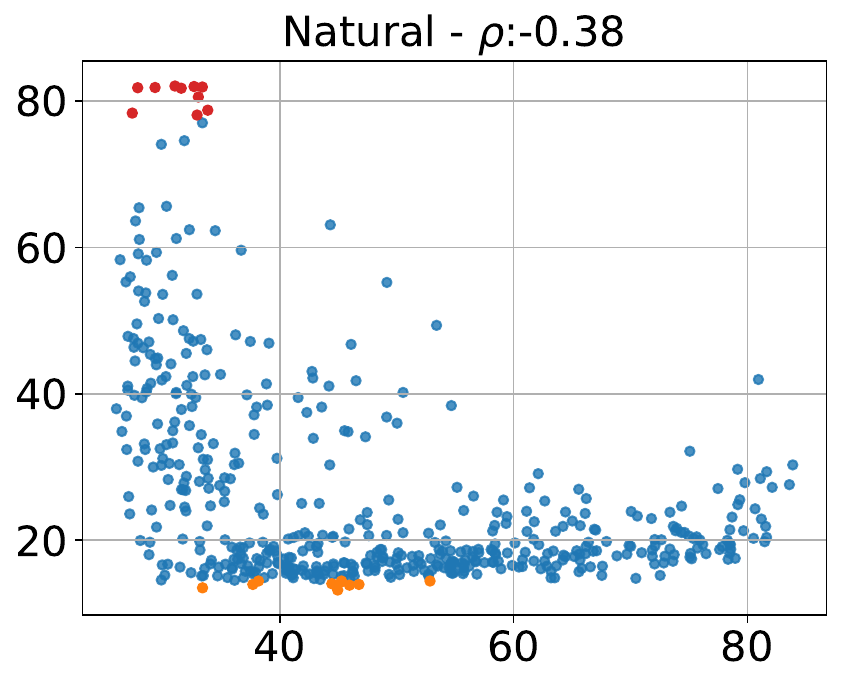} 
    \\
    \caption{Test errors of all 500 sampled architectures on target datasets (y-axis) plotted against the test errors of the same architectures (trained and tested) on ImageNet (x-axis). The top 10 performances on the target datasets are plotted in orange and the worst 10 performances in red.}
    \label{fig:scatter_performance}
\end{figure}

We analyze the architecture-performance relationship (APRs) in two ways. For every target dataset (datsets which are not ImageNet) we plot the test error of every sampled architecture against the test error of the same architecture (trained and tested) on ImageNet, visualizing the relationship of the target dataset's APR with the APR on ImageNet. Second, we compute Spearman's $\rho$ rank correlation coefficient \cite{freedman2007statistics}. It is a nonparametric measure for the strength of the relation between two variables (here the error on the target datasets with the error of the same architecture on ImageNet). Spearman's $\rho$ is defined on $[-1,1]$, where $0$ indicates no relationship and $-1$ or $1$ indicates that the relationship between the two variables can be fully described using only a monotonic function. \par

Figure \ref{fig:scatter_performance} contains the described scatterplots with the corresponding correlation coefficients in the title. The datasets plotted in the top two rows show a strong (Insects) or medium (MLC2008, HAM10000, Cifar100) error correlation with ImageNet. This confirms that many classification tasks have an APR similar to the one induced by ImageNet, which makes ImageNet performance a decent architecture selection indicator for these datasets. The errors on Concrete are independent of the their corresponding ImageNet counterparts since the accuracies are almost saturated with errors between $0$ and $0.5$. This has implications for practical settings, where in such cases suitable architectures should be chosen according to computational and model complexity considerations rather than ImageNet performance, and reinforces the idea that practical problems may lie well outside of the ImageNet visual world \cite{stadelmann2018deep}. The most important insight from Figure \ref{fig:scatter_performance}, however, is that some datasets have a slight (Cifar10) or even strong (Powerline, Natural) \emph{negative error correlation} with ImageNet. Architectures which perform well on ImageNet tend perform sub-par on these datasets. A visual inspection shows that some of the very best architectures on ImageNet perform extraordinarily poor on these three datasets. We can conclude that the \emph{APRs can vary wildly between datasets and high performing architectures on ImageNet do not necessarily work well on other datasets}.

An analysis of the correlations between all datasets (see Figure \ref{fig:scatter_all} in Appendix \ref{chp:ablation}) reveals that Powerline and Natural not only have low correlation with ImageNet but also with most of the other datasets making these two truly particular datasets. Interestingly is the correlation between Powerline and Naural relatively high, which suggests that there is a common trait that makes these two datasets behave differently. MLC 2008, HAM10000 and Cifar100 have a correlation of $0.69$ with each other which indicates that they induce a very similar APR. This APR seems to be fairly universal since MLC 2008, HAM10000 and Cifar100 have a moderate to high correlation with all other datasets.

\subsection{Impact of the Number of Classes}
Having established that APR varies heavily between datasets, leaves us width the questions if it is possible to identify properties of the datasets themselves that influences its APR and if it is possible to control these factors to reduce the APR differences. \par
ImageNet has by far the largest number of classes among all the datasets. Insects, which is the dataset with the second highest class count, also shows the strongest similarity in APR to ImageNet. This suggests that the number of classes might be an important property of a dataset with respect to APR. We test this hypothesis by running an additional set of experiments on subsampled versions of ImageNet. We create new datasets by randomly choosing a varying number of classes from ImageNet and deleting the rest of the dataset. This allows us to isolate the impact of the number of classes while keeping all other aspects of the data itself identical. We create four subsampled ImageNet versions with $100$, $10$, $5$, and $2$ classes, which we call ImageNet-100, ImageNet-10, ImageNet-5, and ImageNet-2, respectively. We refer to the resulting group of datasets (including the original ImageNet) as the ImageNet-X family. The training regime for ImageNet-100 is kept identical to the one of ImageNet, for the other three datasets we switch to top-1 error and train for $40$ epochs, to account for the smaller dataset size. (see section \ref{chp:ds_size_vs_class_nr}  in Appendix \ref{chp:robustness} for a control experiment that disentangles the effects of reduced dataset size and reduced number of classes)\par

\begin{figure}[t]
    \centering
    \includegraphics[width=0.22\textwidth]{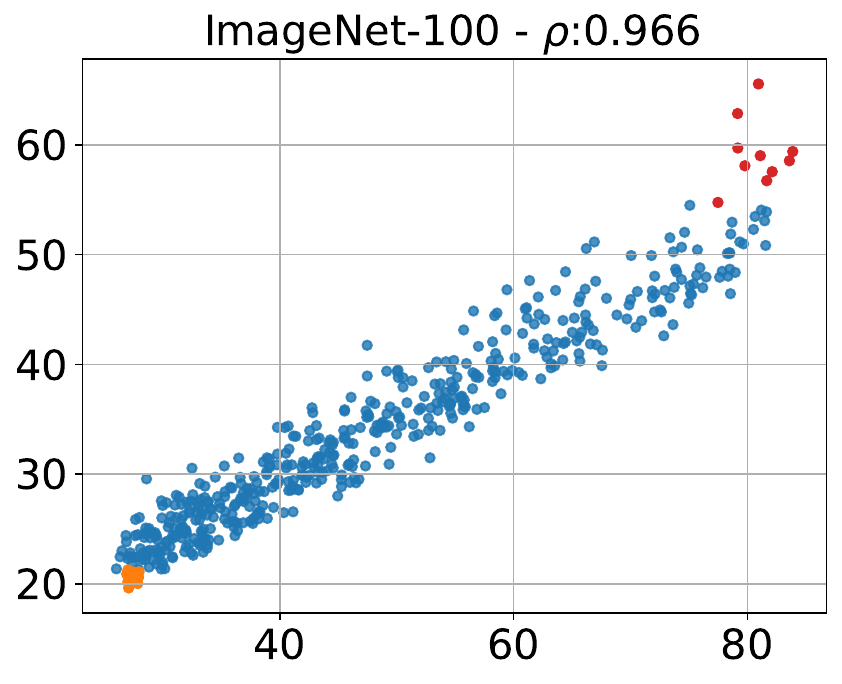} 
    \includegraphics[width=0.22\textwidth]{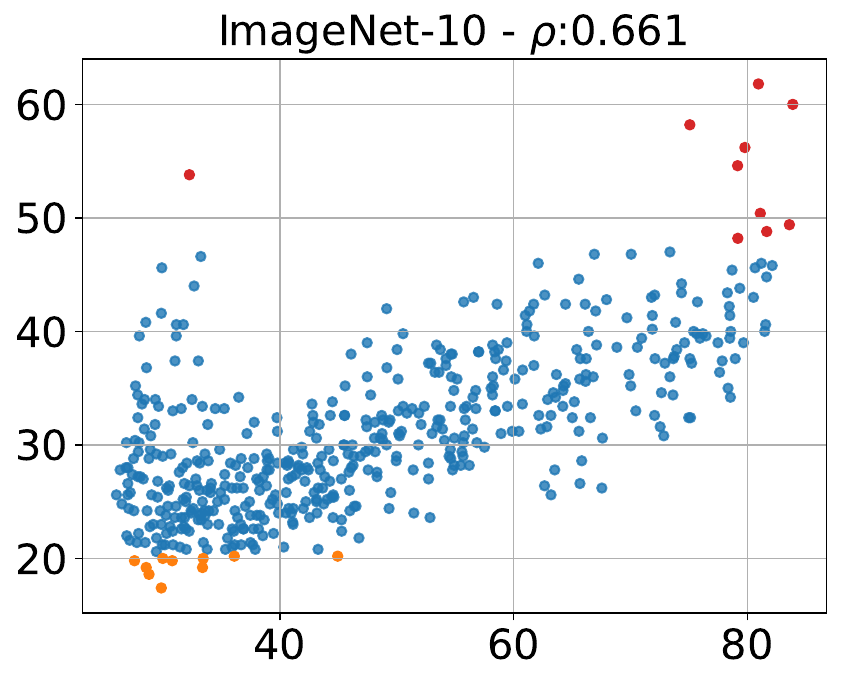} 
    \includegraphics[width=0.22\textwidth]{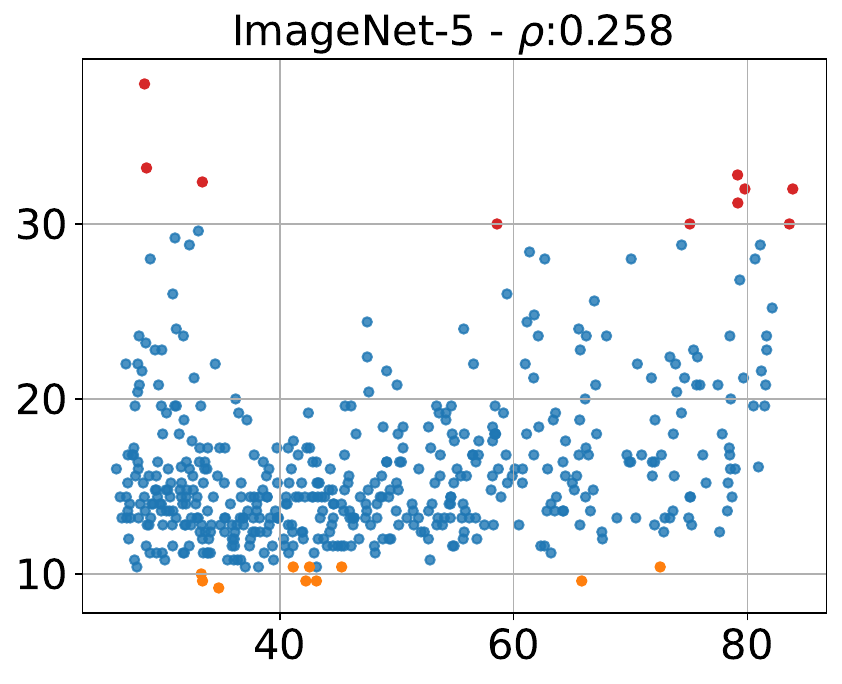} 
    \includegraphics[width=0.22\textwidth]{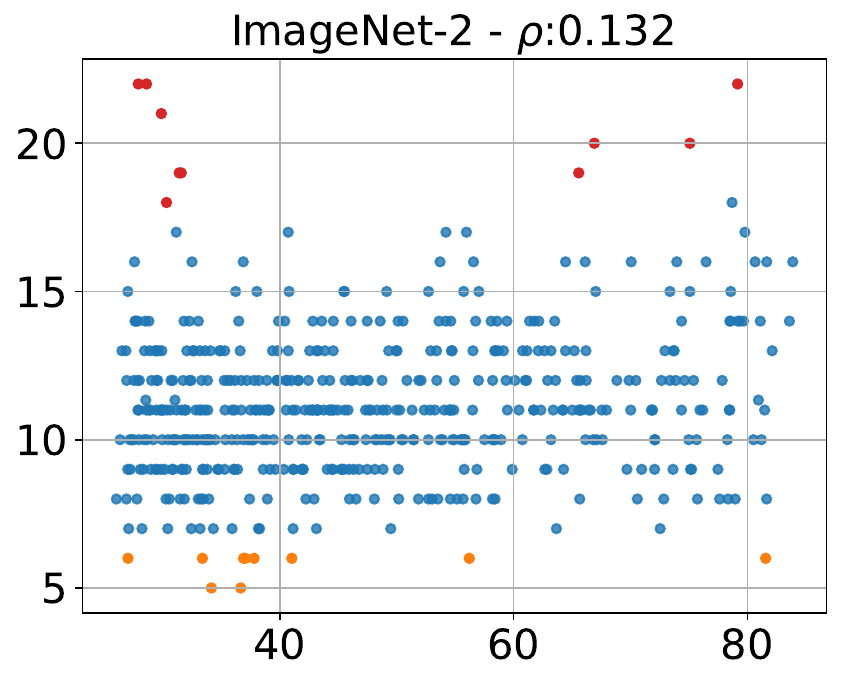} \\
    
    \caption{Error of all $500$ sampled architectures on subsampled (by number of classes) versions of ImageNet (y-axis) plotted against the error of the same architectures on regular ImageNet (x-axis). The top $10$ performances on the target dataset are plotted in orange and the worst $10$ performances in red.}
    \label{fig:scatter_performance_ImageNet}
\end{figure}

Figure \ref{fig:scatter_performance_ImageNet} shows the errors on the subsampled versions plotted against the errors on original ImageNet. APR on ImageNet-100 shows an extremely strong correlation with APR on ImageNet. This correlation significantly weakens as the class count gets smaller. ImageNet-2 is on the opposite end has errors which are practically independent from the ones on ImageNet. \emph{This confirms our hypothesis that the number of classes is a dataset property with significant effect on the architecture to performance relationship.}\par

\begin{figure}[th]
    \centering
    \includegraphics[width=0.22\textwidth]{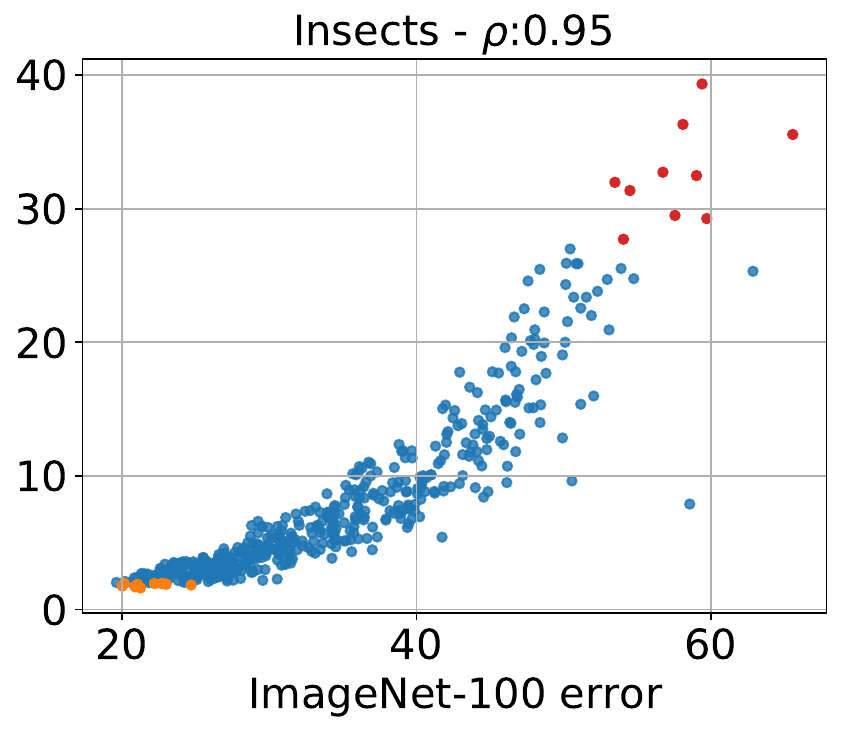}
      \includegraphics[width=0.22\textwidth]{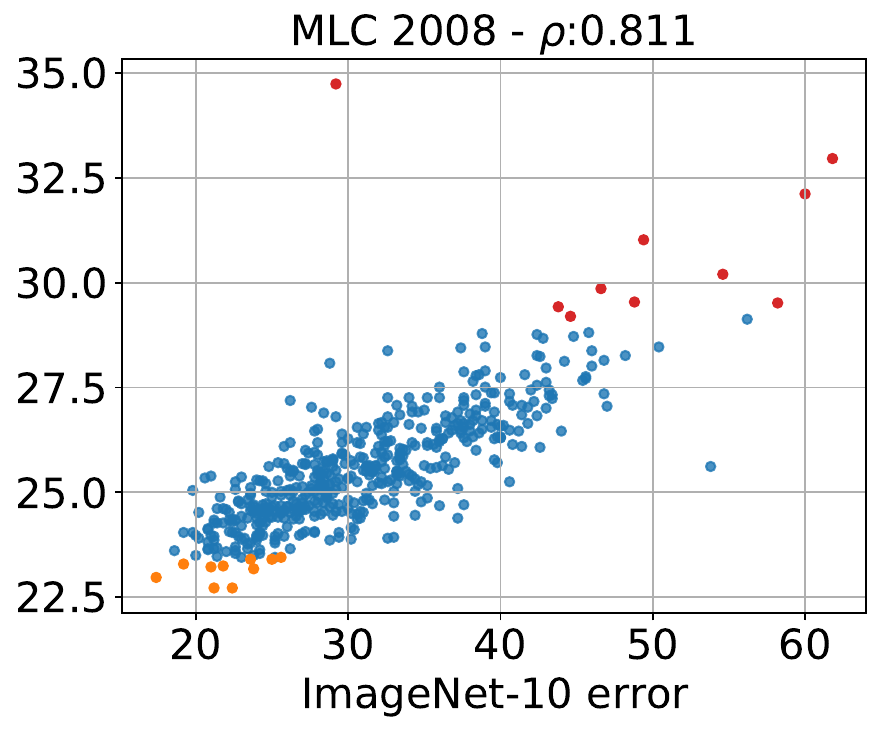} 
    \includegraphics[width=0.22\textwidth]{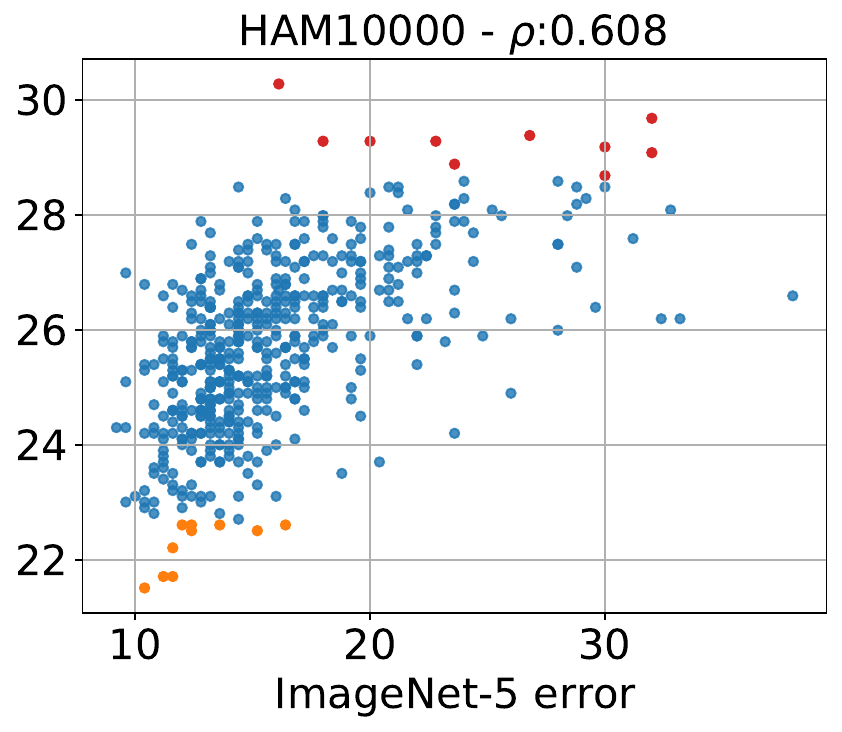} 
    \includegraphics[width=0.22\textwidth]{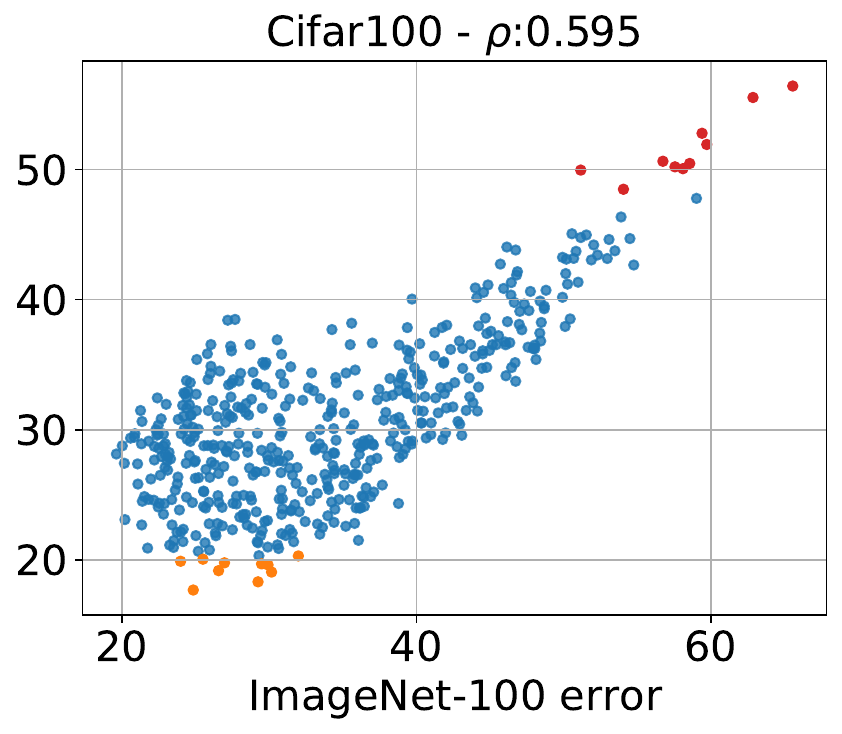}
    \includegraphics[width=0.22\textwidth]{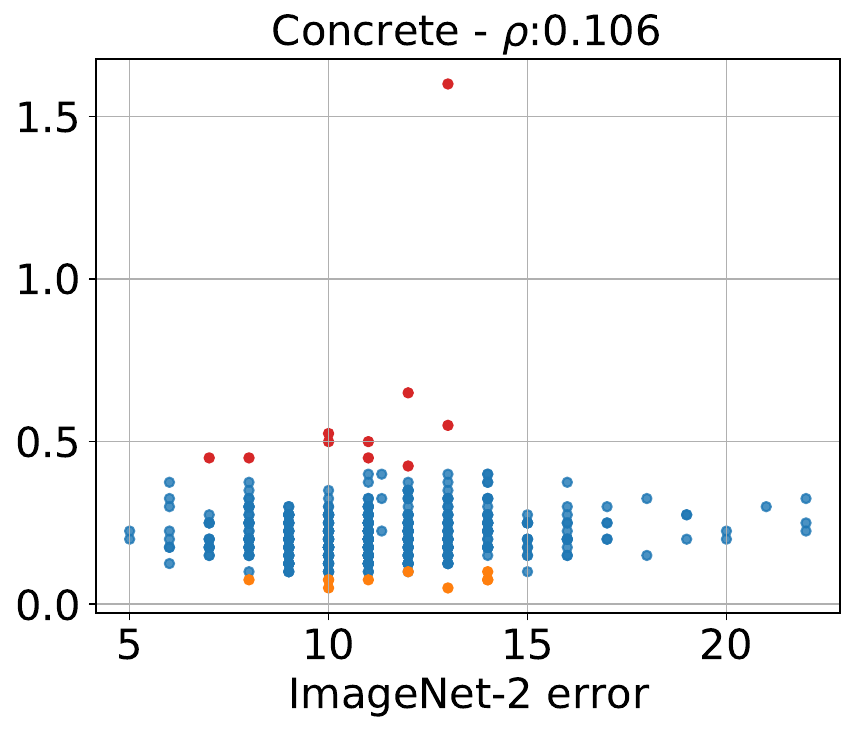} 
    \includegraphics[width=0.22\textwidth]{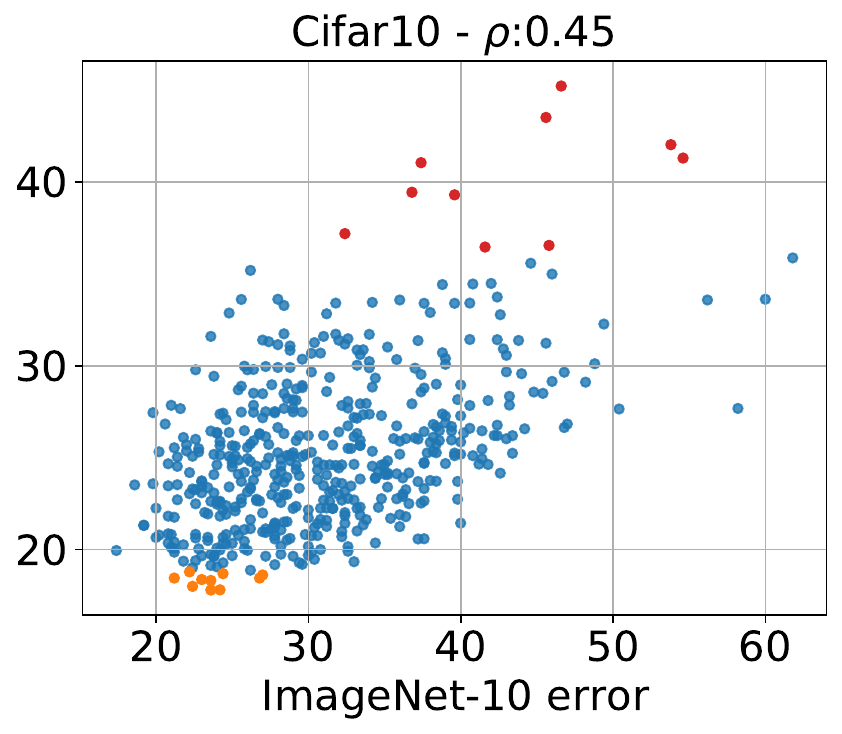} 
    \includegraphics[width=0.22\textwidth]{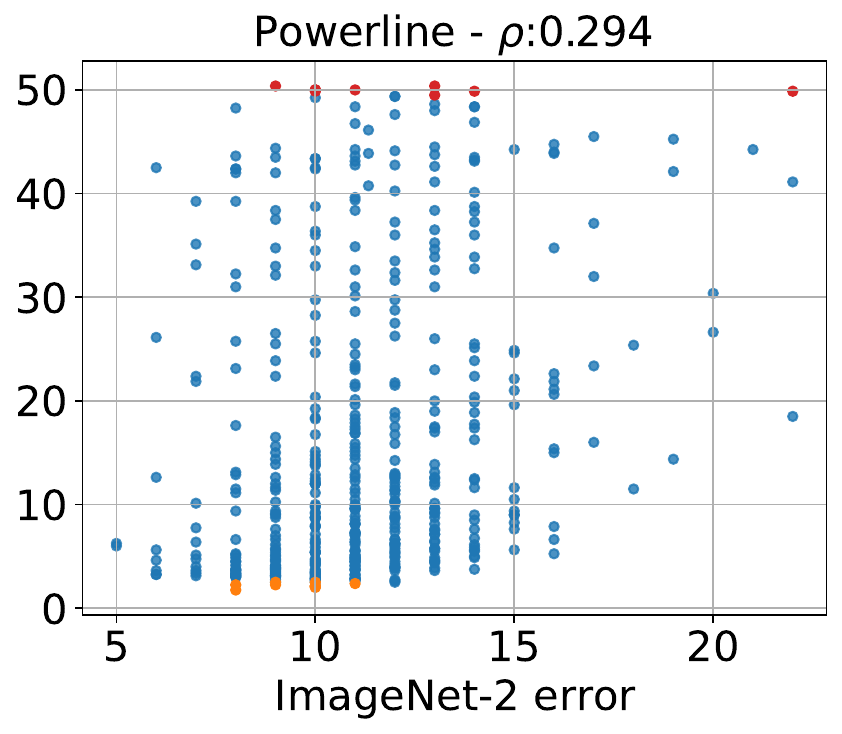} 
    \includegraphics[width=0.22\textwidth]{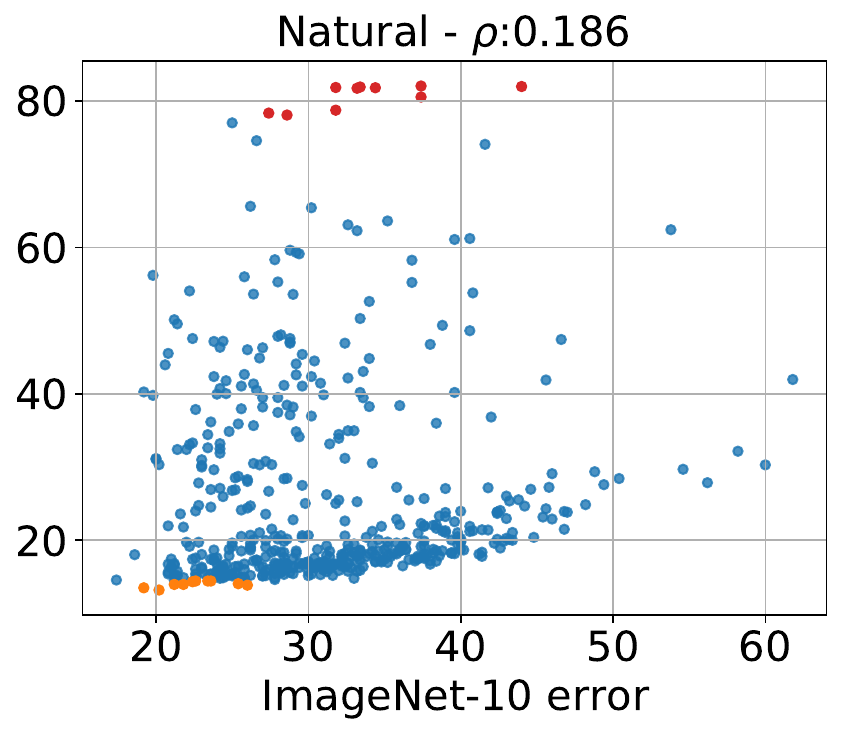}     
    \caption{Test errors of all $500$ sampled architectures on target datasets (y-axis) plotted against the test errors of the same architectures on the ImageNet-X (x-axis). The top $10$ performances on the target dataset are orange, the worst $10$ performances red.}
    \label{fig:scatter_performance_imgnet_x}
\end{figure}

We have observed that the number of classes has a profound effect on the APR associated with ImageNet-X members. It is unlikely that simply varying the number of classes in this dataset is able to replicate the diversity of APRs present in an array of different datasets. However, it is reasonable to assume that a dataset's APR is better represented by the ImageNet-X member closest in terms of class count, instead of ImageNet. We thus recreate Figure \ref{fig:scatter_performance} with the twist of not plotting the target dataset errors against ImageNet, but against the ImageNet-X variant closest in class count (see Figure \ref{fig:scatter_performance_imgnet_x}). We observe gain in correlation across all datasets, in the cases of MLC2008 or Cifar10 a quite extreme one. The datasets which have a strong negative correlation with ImageNet (Powerline, Natural) have slightly (Natural) or even moderately (Powerline) positive correlation to their ImageNet-X counterparts. A visual inspection shows that the best models on Imagenet-X also yield excellent results on Powerline and Natural, which was not the case for ImageNet. Table \ref{tbl:corr} shows the error correlations of all target datasets with ImageNet as well as with their ImageNet-X counterpart. \emph{The move from ImageNet to ImageNet-X more than doubles the average correlation (from $0.19$ to $0.507$), indicating that the ImageNet-X family of datasets is capable to represent a much wider variety of APRs than ImageNet alone.}

\begin{table}[t]
    \caption{Comparison of error correlations between target datasets and ImageNet as well as the closest  ImageNet-X member.}
    \label{tbl:corr}
    \vskip -0.1in
    \begin{center}
        \begin{small}
            \begin{sc}
                \begin{tabular}{lcccr}
                    \toprule
                    Dataset & $\rho$ -ImageNet  & $\rho$ -ImageNet-X & Difference \\
                    \midrule
                    concrete   & $0.001$ & $0.106$ & $0.105$ \\
                    MLC2008    & $0.476$  & $0.811$ & $0.335$\\
                    ham10000   & $0.517$  & $0.608$  & $0.091$\\
                    powerline  & $-0.436$ & $0.294$ & $0.73$\\
                    insects    & $0.967$  & $0.95$ & $-0.017$\\
                    natural    & $-0.38$ & $0.186$ & $0.566$\\
                    cifar10    & $-0.104$ & $0.45$ & $0.554$\\
                    cifar100   & $0.476$  & $0.595$ & $0.119$\\
                    \midrule
                    Average & $0.19$  & $0.507$  & $0.317$ \\
                    \bottomrule
                \end{tabular}
            \end{sc}
        \end{small}
    \end{center}
    \vskip -0.2in
\end{table}

\subsection{Identifying Drivers of Difference between Datasets}

\begin{figure}[t]
    \centering
    \includegraphics[width=0.23\textwidth]{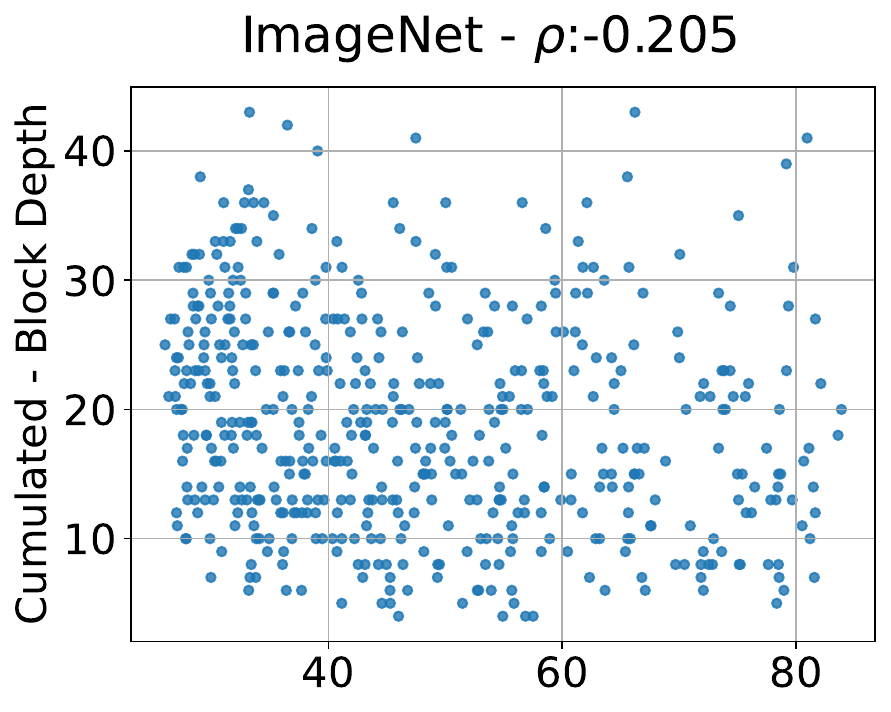} 
    \includegraphics[width=0.23\textwidth]{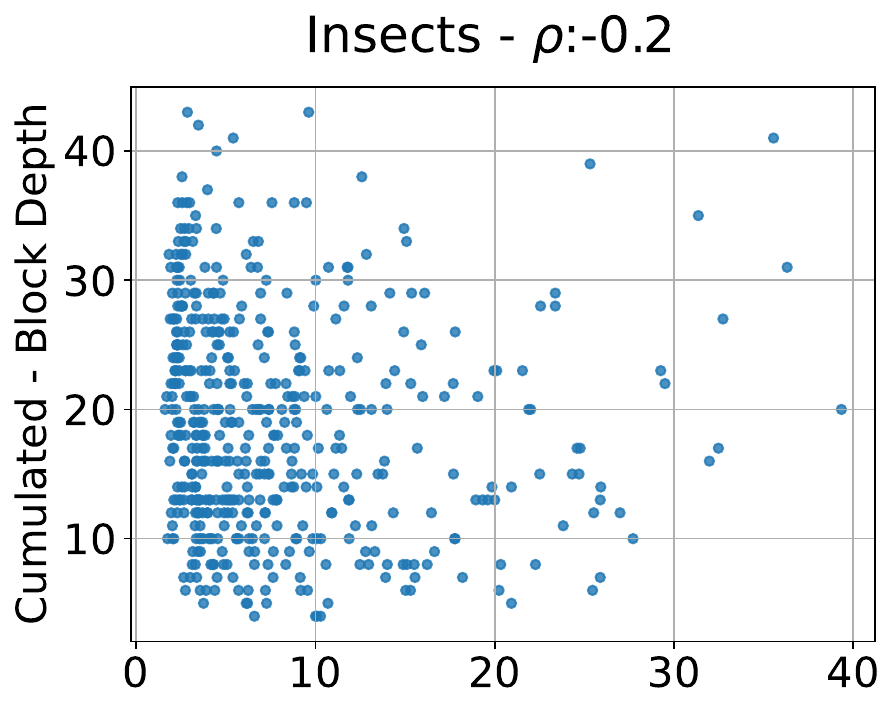} 
    \includegraphics[width=0.23\textwidth]{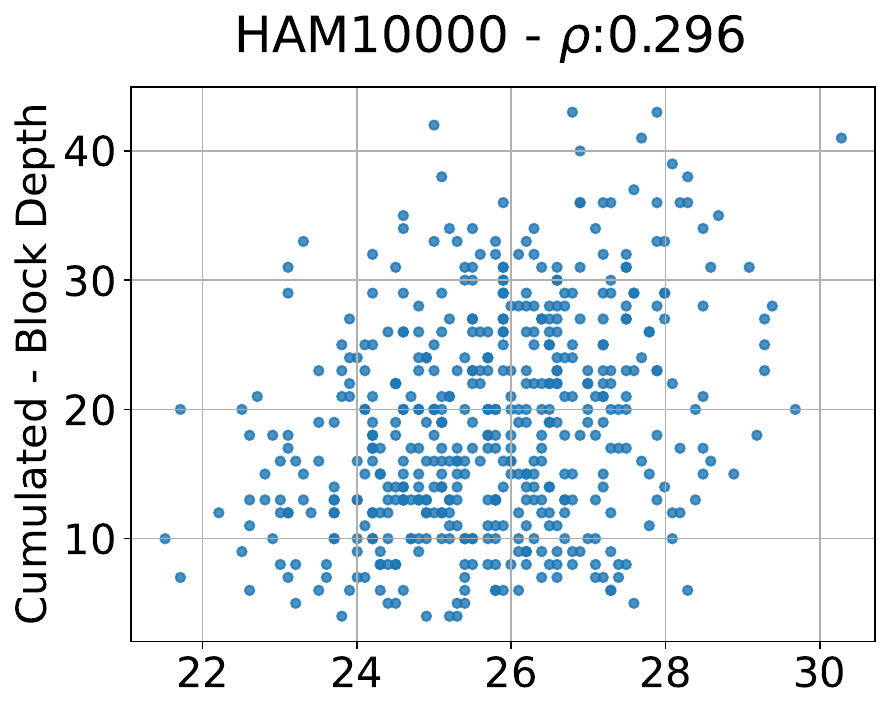} 
    \includegraphics[width=0.23\textwidth]{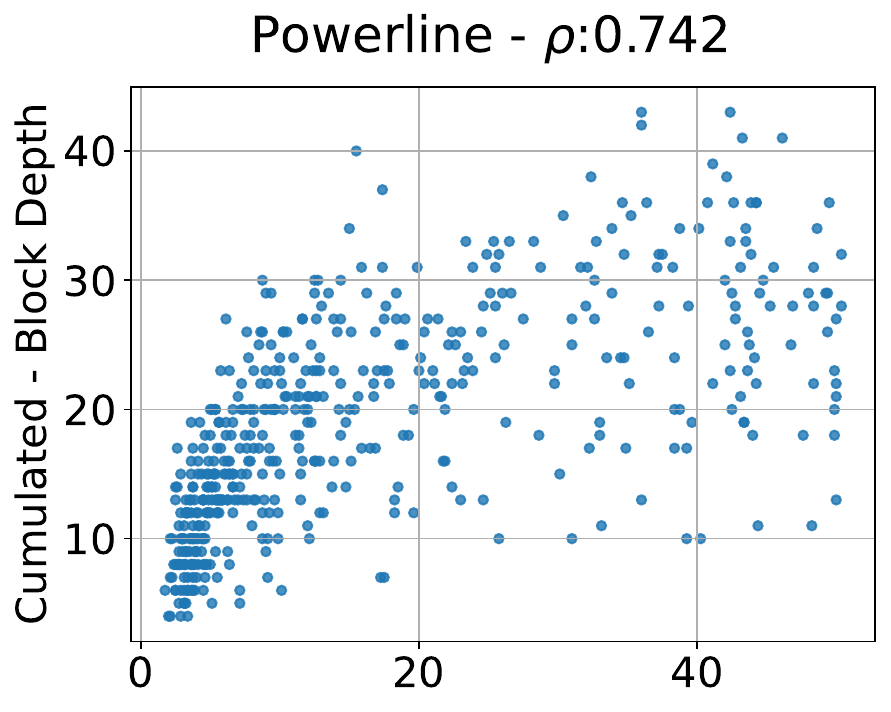} 
    \includegraphics[width=0.23\textwidth]{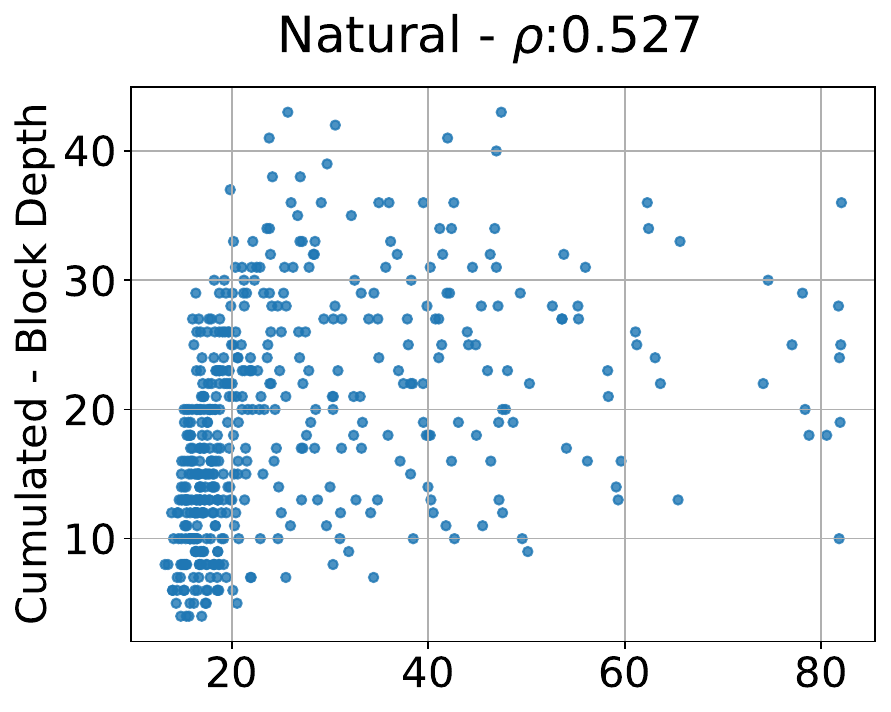}
    \includegraphics[width=0.23\textwidth]{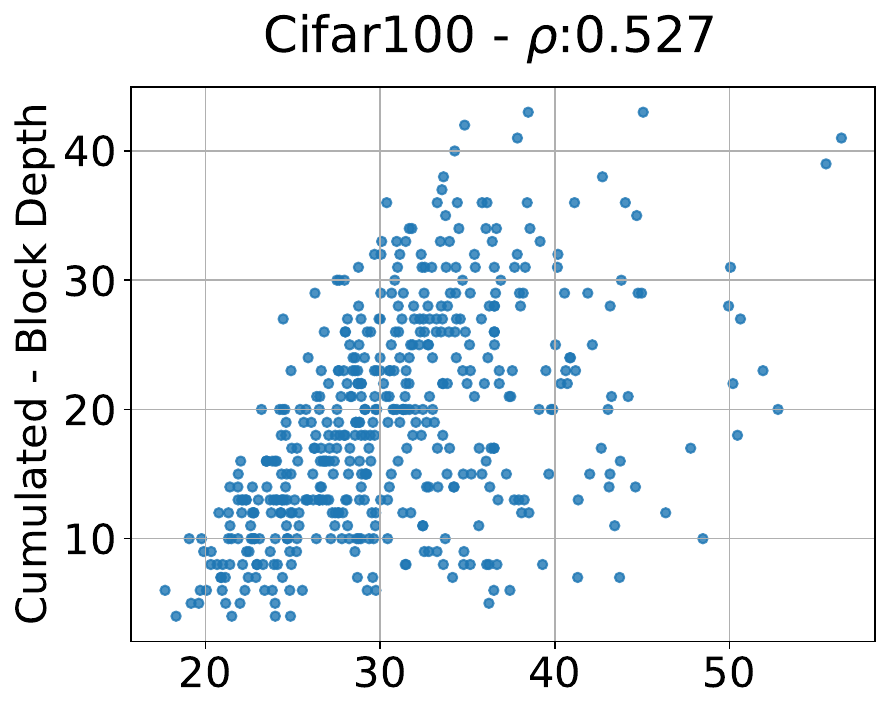} 
    \caption{Errors of all 500 sampled architectures on ImageNet, Insects, HAM10000, Powerline, Natural, and Cifar100 (x-axis) plotted against the cumulative block \emph{depths} (y-axis).}
    \label{fig:hexbin_block_depth}
\end{figure}

The block width and depth parameters of the top $15$ architectures for ImageNet (see Figure \ref{fig:top} in Appendix \ref{chp:ablation}) follow a clear structure: they consistently start with low values for both block depth and width in the first stage, then the values steadily increase across the stages for both parameters. The error relationships observed in Figure \ref{fig:scatter_performance} are consistent with how well these patterns are replicated by the other datasets. Insects shows a very similar pattern, MLC2008 and HAM10000 have the same trends but more noise. Powerline and Natural clearly break from this structure, having a flat or decreasing structure in the block width and showing a quite clear preference for a small block depth in the final stage. Cifar10 and Cifar100 are interesting cases, they have the same behaviour as ImageNet with respect to block width but a very different one when it comes to block depth.
\par

We thus investigate the effect of the cumulative block depth (summation of the depth parameter for all four stages, yielding the total depth of the architecture) across the whole population of architectures by plotting the cumulative block depth against the test error for the six above-mentioned datasets. Additionally, we compute the corresponding correlation coefficients. Figure \ref{fig:hexbin_block_depth} shows that the best models for ImageNet have a cumulative depth of at least $10$. Otherwise there is no apparent dependency between the ImageNet errors and cumulative block depth. The errors of Insects do not seem to be related to the cumulative block depth at all. HAM10000 has a slight right-leaning spread leading to a moderate correlation, but the visual inspection shows no strong pattern. The errors on Powerline, Natural, and Cifar100 on the other hand have a strong dependency with the cumulative block depth. The error increases with network depth for all three datasets. with the best models all having a cumulative depth smaller than $10$. 
\par

\begin{figure}[t]
    \centering
    \includegraphics[width=0.22\textwidth]{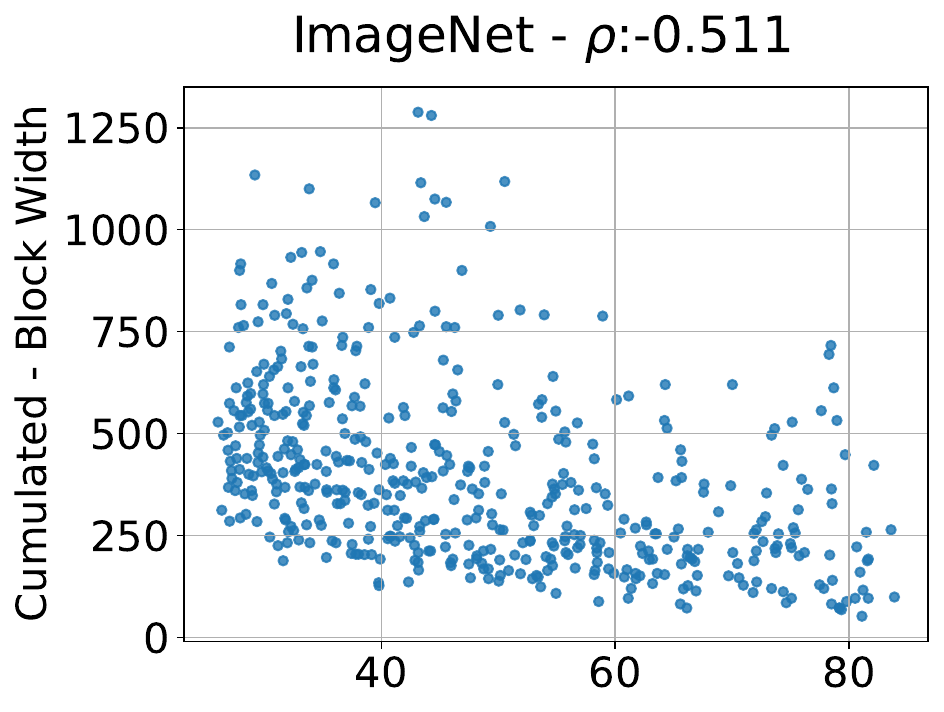} 
    \includegraphics[width=0.22\textwidth]{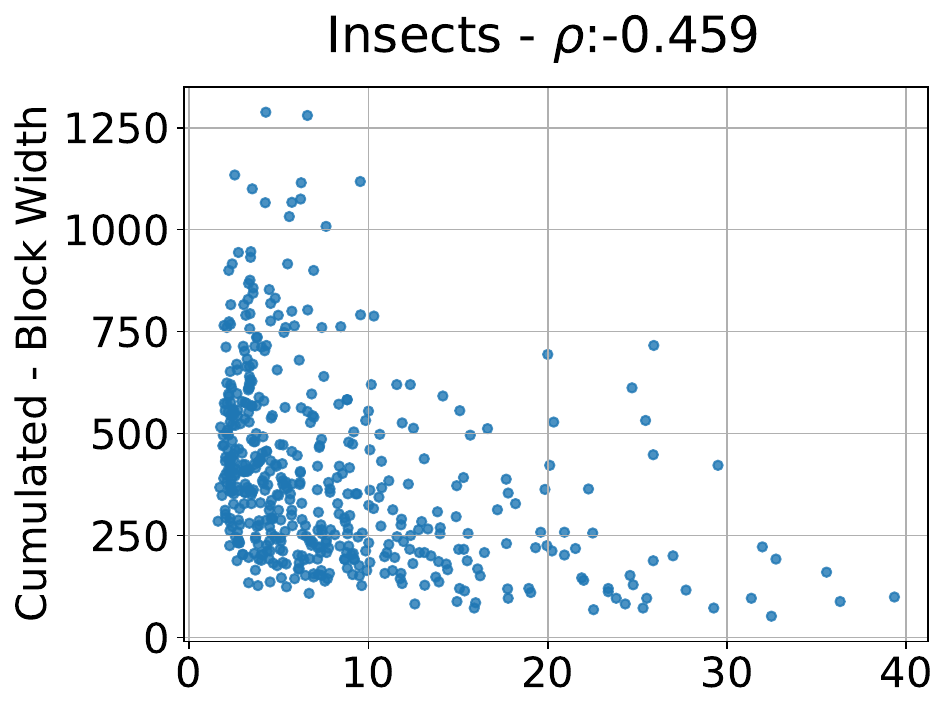} 
    \includegraphics[width=0.22\textwidth]{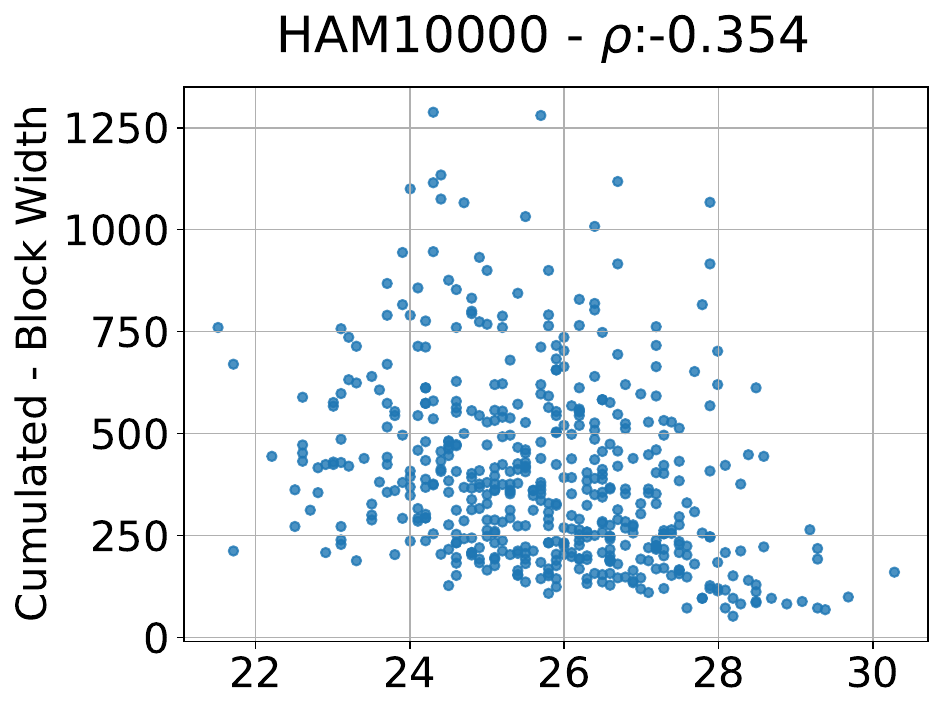}
    \includegraphics[width=0.22\textwidth]{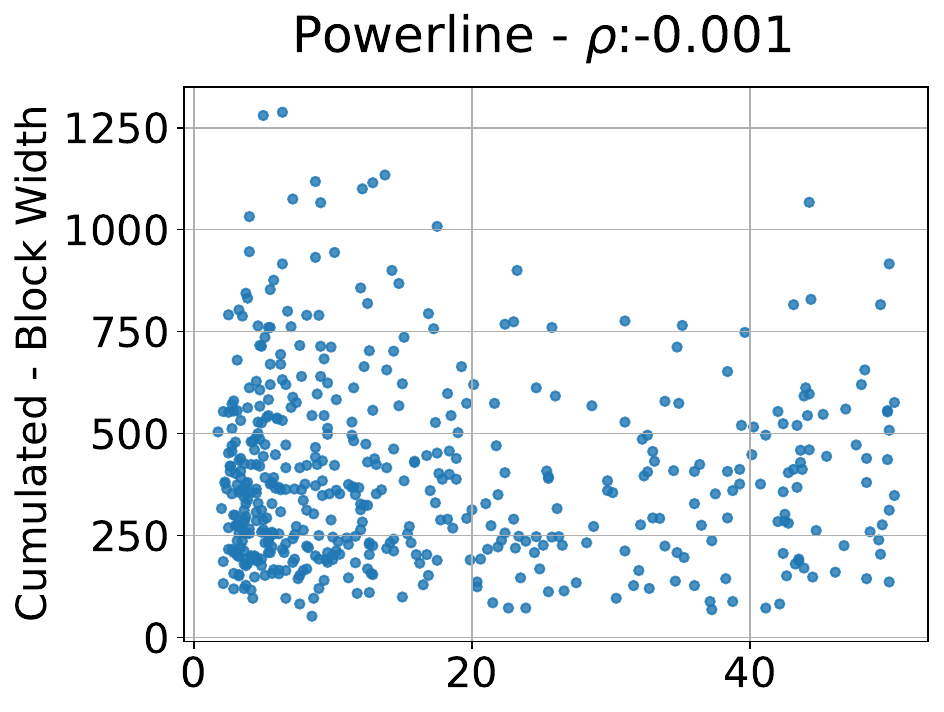}
    \includegraphics[width=0.22\textwidth]{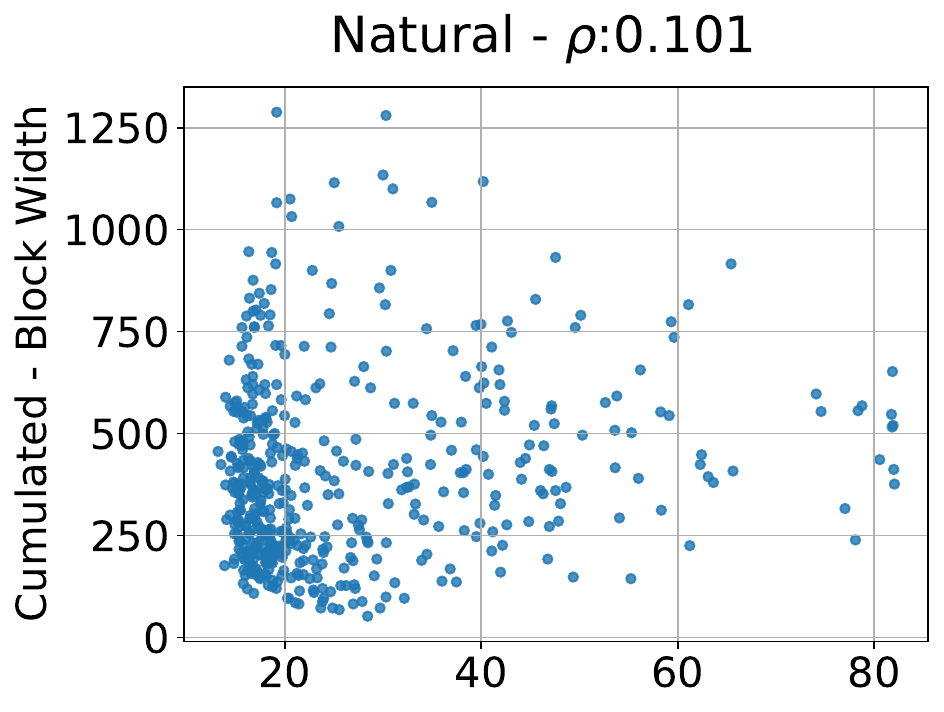}
    \includegraphics[width=0.22\textwidth]{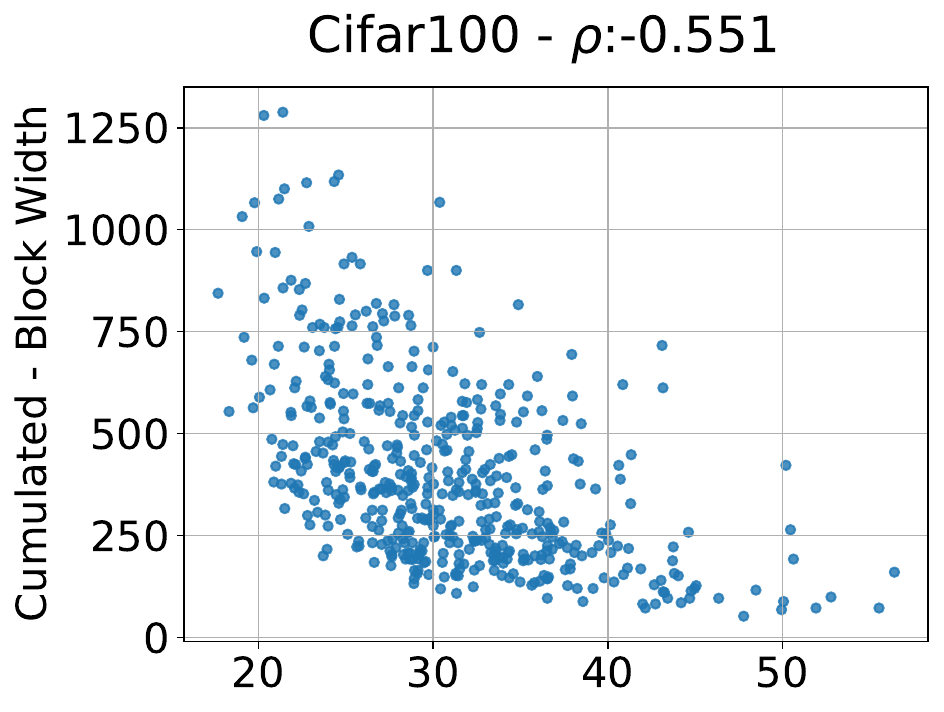} 
    \caption{Errors of all 500 sampled architectures on ImageNet, Insects, HAM10000 and Cifar100 (x-axis) plotted against the cumulative block \emph{widths} (y-axis).}
    \label{fig:hexbin_block_width}
    \vspace{-0.1cm}
\end{figure}

We also plot the cumulative block widths against the errors and compute the corresponding correlation coefficients for the same six datasets (see Figure \ref{fig:hexbin_block_width}). We observe that the ImageNet errors are negatively correlated with the cumulative block width, and visual inspection shows that a cumulative block width of at least $250$ is required to achieve a decent performance. The errors on Insects and HAM10000 replicate this pattern to a lesser extent, analogous to the top $15$ architectures. Powerline and Natural have no significant error dependency with the cumulative block width, but Cifar100 has an extremely strong negative error dependency with the cumulative block width, showing that it is possible for a dataset to replicate the behaviour on ImageNet in one parameter but not the other. In the case of Cifar100 and ImageNet, low similarity in block depth and high similarity in block width yield a medium overall similarity of ARPs on Cifar100 and Imagenet. This is consistent with the overall relationship of the two datasets displayed in Figure \ref{fig:scatter_performance}. \par

\begin{table}[t]
    \caption{Correlation of observed error rates with the cumulative block depth and width parameters for all ImageNet-X datasets.}
    \label{tbl:correlation_nr_classes}
    \vskip 0.0in
    \begin{center}
        \begin{small}
            \begin{sc}
                \begin{tabular}{lcccr}
                    \toprule
                    Dataset & C. Block Depth  & C. Block Width   \\
                    \midrule
                    ImageNet   & $-0.205$ & $-0.511$ \\
                    ImageNet-100  & $-0.022$  &$-0.558$  \\
                    ImageNet-10   & $0.249$ & $-0.457$ \\
                    ImageNet-5  & $0.51$ &  $-0.338$\\
                    ImageNet-2    & $0.425$ & $-0.179$ \\
                    \bottomrule
                \end{tabular}
            \end{sc}
        \end{small}
    \end{center}
    \vskip -0.1in
\end{table}

Combining this result with the outcome of the last section, we study the interaction between the number of classes, the cumulated block depth and the cumulative block width. Table \ref{tbl:correlation_nr_classes} contains the correlations between cumulative block depth/width and the errors on all members of ImageNet-X. With decreasing number of classes, the correlation coefficients increase for cumulative block depth and cumulative block width. Although the effect on cumulative block depth is stronger, there is a significant impact on both parameters. We therefore can conclude that \emph{both optimal cumulative block depth and cumulative block with can drastically change based on the dataset choice and that both are simultaneously influenced by the class count.}


\section{Discussion and Conclusions}
\label{chp:conc}

\textbf{ImageNet is not a perfect proxy.} We have set out to explore how well other visual classification datasets are represented by ImageNet. Unsurprisingly there are differences between the APRs induced by the datasets. More surprising and worrying, however, is that for some datasets ImageNet not only is an imperfect proxy but a very bad one. The negative error correlations with Natural, Powerline and Cifar10 indicates that architecture search based on ImageNet performance is worse than random search for these datasets. 

\par

\textbf{Varying the number of classes is a cheap and effective remedy.} It is striking how much more accurately the ImageNet-X family is able to represent the diversity in APRs present in our dataset collection, compared to just ImageNet by itself. It has become commonplace to test new architectures in multiple complexity regimes \cite{DBLP:conf/cvpr/HeZRS16,DBLP:journals/corr/HowardZCKWWAA17}, we argue for augmenting this testing regime with an additional dimension for class count. This simple and easy to implement extension would greatly extend the informative value of future studies on neural network architectures.
\par
  
\par

\textbf{Future directions.}
A future similar study should shed light on how well the breadth of other domains such as object detection or speech classification are represented by their essential datasets. In doing so it could be verified if the varying the number of classes also helps covering more dataset variability in these domains.\\
A labeled dataset will always be a biased description of the visual world, due to having a fixed number of classes and being built with some systematic image collection process. Self-supervised learning of visual representations \cite{DBLP:journals/corr/abs-1902-06162} could serve as remedy for this issue. Self-supervised architectures could be fed with a stream completely unrelated images, collected from an arbitrary number of sources in a randomized way. A comparison of visual features learned in this way could yield a more meaningful measure of the quality of CNN architectures.
\par
\textbf{Limitations}
As with any experimental analysis of a highly complex process such as training a CNN it is virtually impossible to consider every scenario. We list below three dimensions along which our experiments are limited together with measures we took to minimize the impact of these limitations.
\par
\textit{Data scope:} We criticize ImageNet for only representing a fraction of the ``visual world''. We are aware that our dataset collection does not span the entire ``visual world'' either but went to great lengths to maximise the scope of our dataset collection by purposefully choosing datasets from different domains, which are visually distinct.
\par
\textit{Architecture scope:} We sample our architectures from the large AnyNetX network space. It contains the CNN building blocks to span basic designs such as AlexNet or VGG as well as the whole ResNet, ResNeXt and RegNet families. We acknowledge that there are popular CNN components not covered, however, Radosavovic et al. \cite{DBLP:conf/cvpr/RadosavovicKGHD20} present ablation studies showing that network designs sourced from high performing regions in the AnyNetX space also perform highly when swapping in different originally missing components such as depthwise convolutions \cite{DBLP:conf/cvpr/Chollet17}, swish activation functions \cite{DBLP:conf/iclr/RamachandranZL18} or the squeeze-and-excitation \cite{DBLP:conf/cvpr/HuSS18} operations.
\par
\textit{Training scope:} When considering data augmentation and optimizer settings there are almost endless possibilities to tune the training process. We opted for a very basic setup with no bells an whistles in general. For certain such aspects of the training, which we assumed might skew the results of our study (such as training duration, dataset prepossessing etc.), we have conducted extensive ablation studies to ensure that this is not the case (see sec. \ref{chp:training_duration} and \ref{chp:ds_distrib} in Appendix \ref{chp:robustness}).

\label{chp:limit}


\subsubsection*{Acknowledgments}
This work has been financially supported by grants 25948.1 PFES-ES ``Ada'' (CTI), 34301.1 IP-ICT ``RealScore'' (Innosuisse) and ERC Advanced Grant AlgoRNN nr. 742870. We are grateful to Frank P. Schilling for his valuable inputs.

\bibliographystyle{IEEEtran}
\bibliography{iclr2022_conference}

\begin{appendices}
\section{Verifying the numerical robustness of our study}
\label{chp:robustness}
This Chapter we present additional studies designed to test for possible flaws or vulnerabilities in our experiments. We conduct these to further strengthen the empirical robustness of our results. 

\subsection{Stability of Empirical Results on Cifar10}
\label{chp:cifar_stability}

\begin{figure}[!htp]
    \centering
    \includegraphics[width=0.23\textwidth]{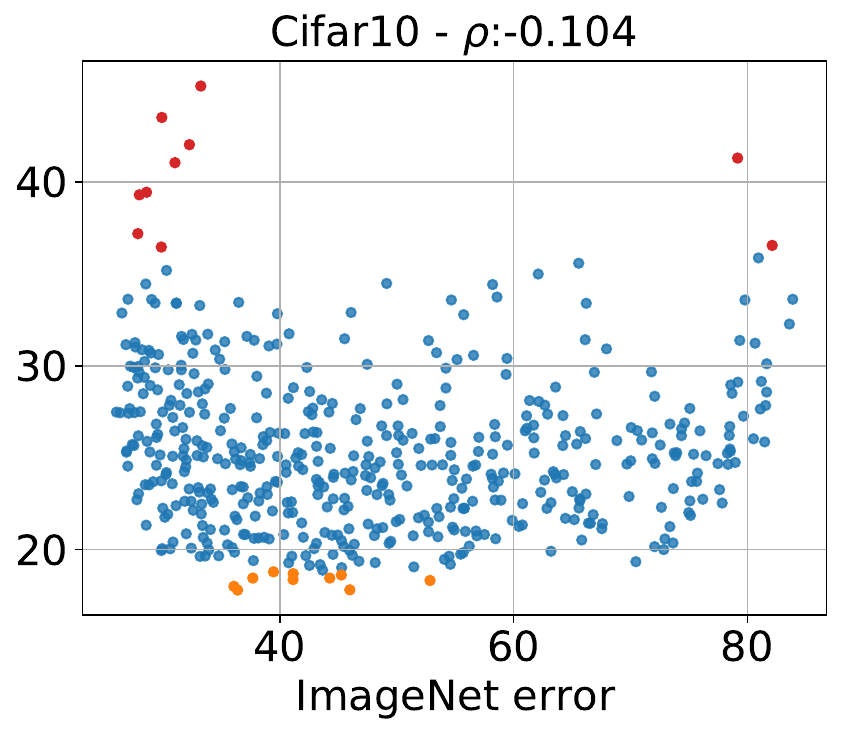} 
    \includegraphics[width=0.23\textwidth]{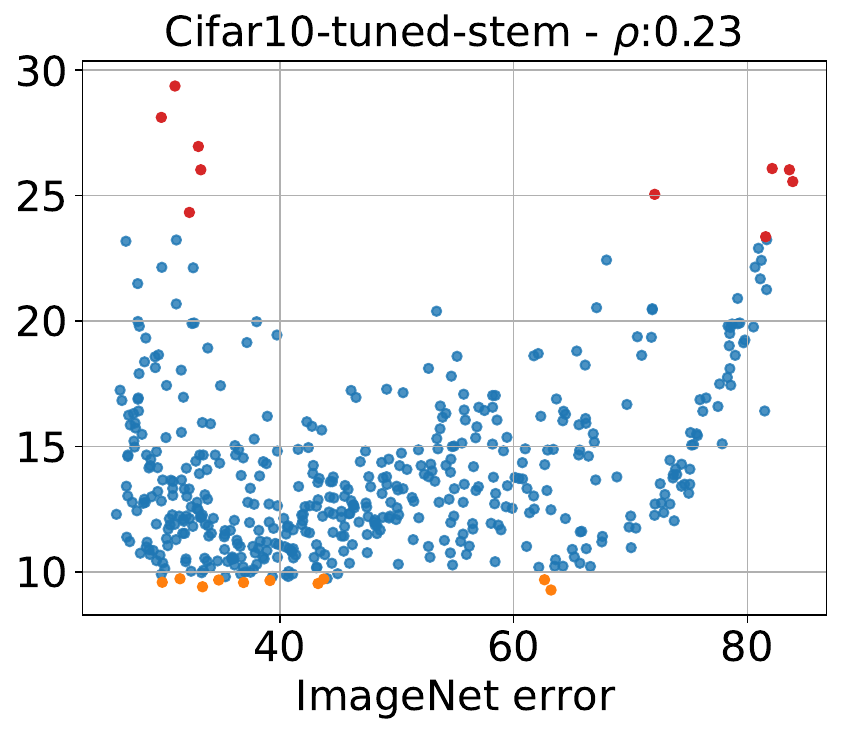} 
    \includegraphics[width=0.23\textwidth]{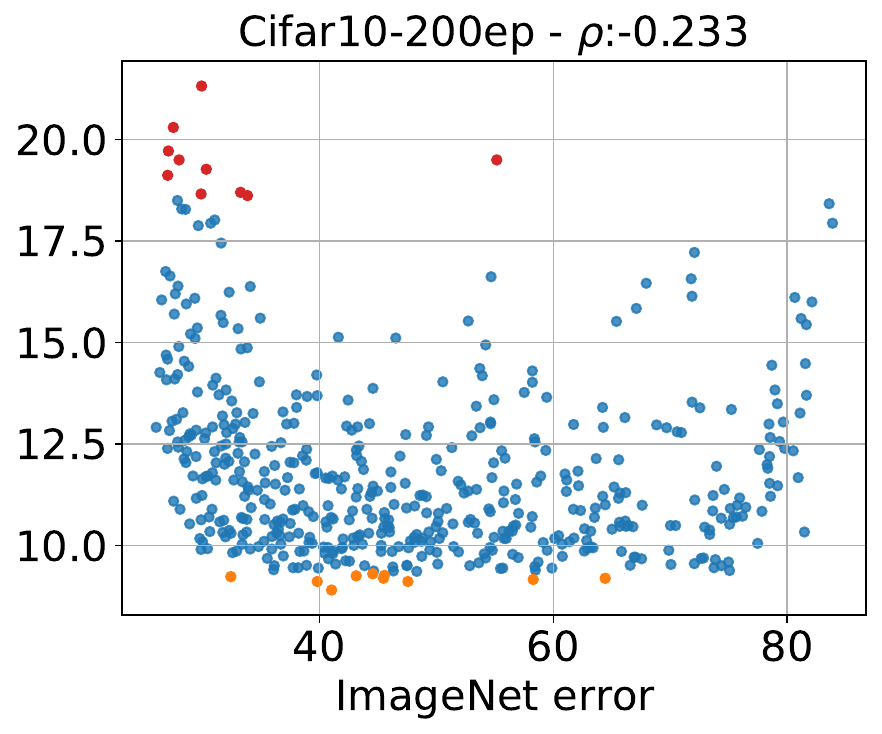} 
    \includegraphics[width=0.23\textwidth]{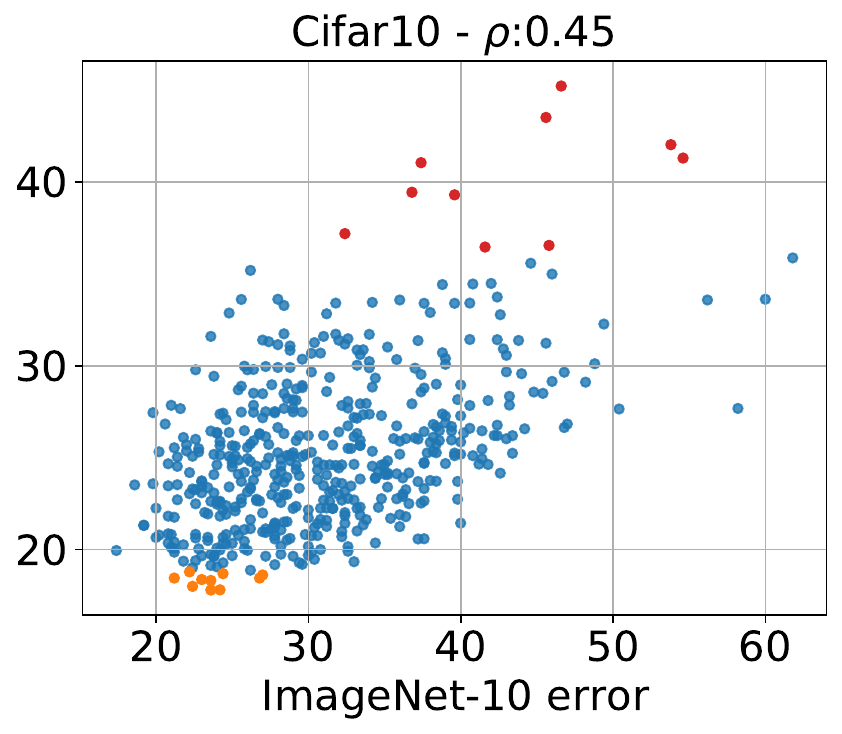}
    \includegraphics[width=0.23\textwidth]{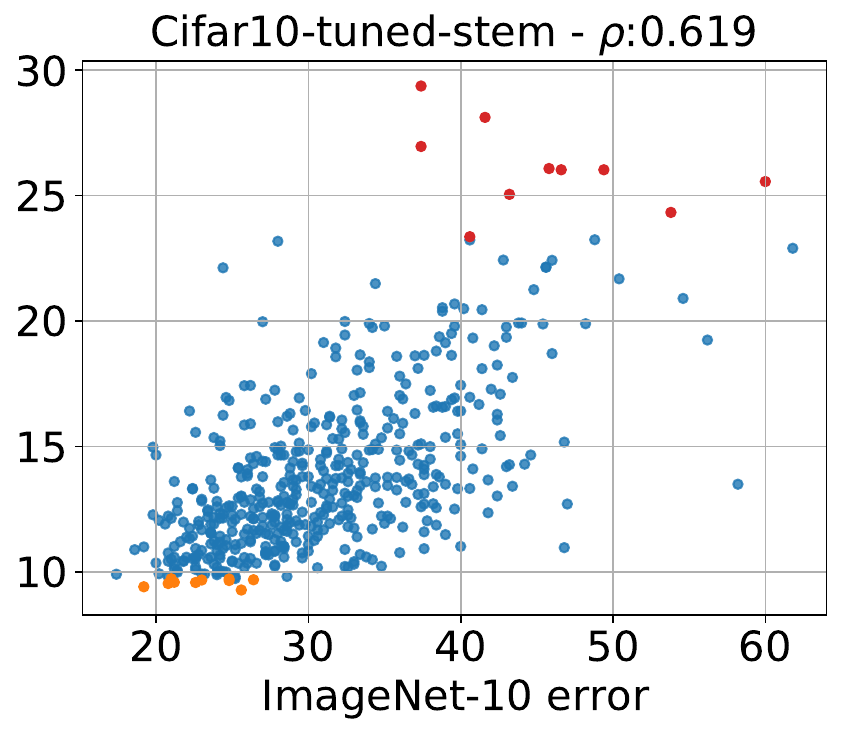} 
    \includegraphics[width=0.23\textwidth]{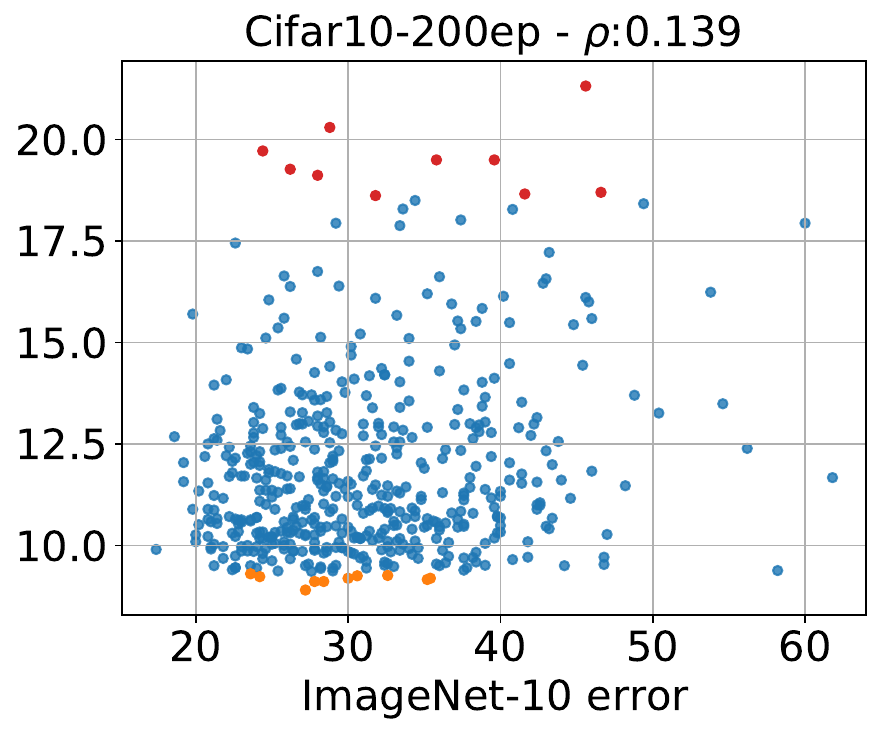}

    \caption{The Cifar10 test errors of all 500 architectures plotted against ImageNet (top row) and ImageNet-10 (bottom row), shown for our original Cifar10 training (left column), training with a Cifar10 specific stem in the architecture (middle column), and training for $200$ epochs, which is roughly $6$ times longer (right column). The plots show that the error correlation with ImageNet-10 is much larger in all three cases, confirming that optimizing for individual Cifar10 performance does not alter our core result.}
    \label{fig:cifar10_tuning}
\end{figure}

The top-1 errors of our sampled architectures on Cifar10 lie roughly between $18$ and $40$, which is fairly poor, not only compared to the state of the art but also compared to performance that can be achieved with fairly simple models. This calls into question if our Cifar10 results are flawed in a way that might have lead us to wrong conclusions. We address this by running additional tests on Cifar10 and evaluate their impact on our main results.
We get a goalpost for what performance would be considered good with our style of neural network and training setup by running the baseline code for Cifar10 published by Radosavovic et al. \cite{DBLP:conf/cvpr/RadosavovicKGHD20}. Table \ref{tbl:cifar10_baselines} shows that these baseline configurations achieve much lower error rates. 
We aim to improve the error results on Cifar10 in two ways: First we train our architecture population with standard settings for $200$ epochs instead of $30$, second we replaced the standard network stem with one that is specifically built for Cifar10, featuring less stride and no pooling. Figure \ref{fig:cifar10_tuning} shows scatterplots of the errors from all $500$ architectures on Cifar10 against the errors on ImageNet and ImageNet-10. We can see that both new training methods manage to significantly improve the performance with a minimum top-1 error below $10$ in both cases. More importantly can we observe that both new training methods have, despite lower overall error, a very similar error relationship to ImageNet. The error correlation is even slightly lower than with our original training (replicated in Figure \ref{fig:cifar10_tuning} left row). We can also see that in all three cases the error relationship can be significantly strengthened by replacing ImageNet with ImageNet-10, \textit{this shows that tuning for individual performance on a dataset does not significantly impact the error relationships between datasets which further strengthens our core claim.}

\begin{table}[]
    \caption{Top-1 error of reference network implementations \cite{DBLP:conf/cvpr/RadosavovicKGHD20} for Cifar10.}
    \label{tbl:cifar10_baselines}
    \begin{center}
    \begin{adjustbox}{width=\columnwidth}
            \begin{sc}
                \begin{tabular}{lcccc}
                    \toprule
                Model & ResNet-56 & ResNet-110 & AnyNet-56 & AnyNet-110 \\
                Error & 5.91      & 5.23       & 5.68      & 5.59      \\
                    \bottomrule
                \end{tabular}
            \end{sc}
        \end{adjustbox}
    \end{center}
\end{table}

\subsection{Verifying Training Duration}
\label{chp:training_duration}
\begin{figure}[!htp]
    \centering
    \includegraphics[width=0.35\textwidth]{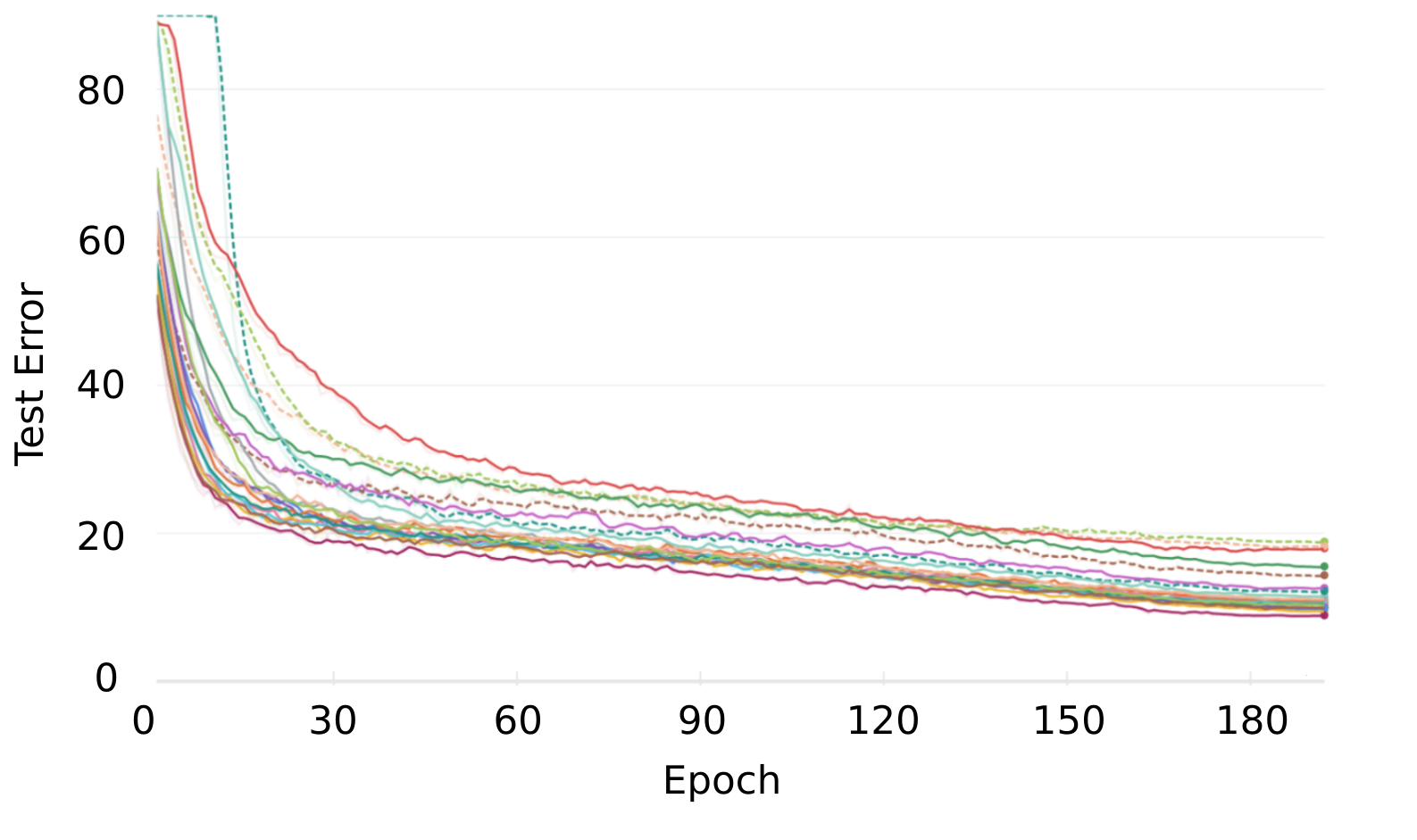} 
    \includegraphics[width=0.35\textwidth]{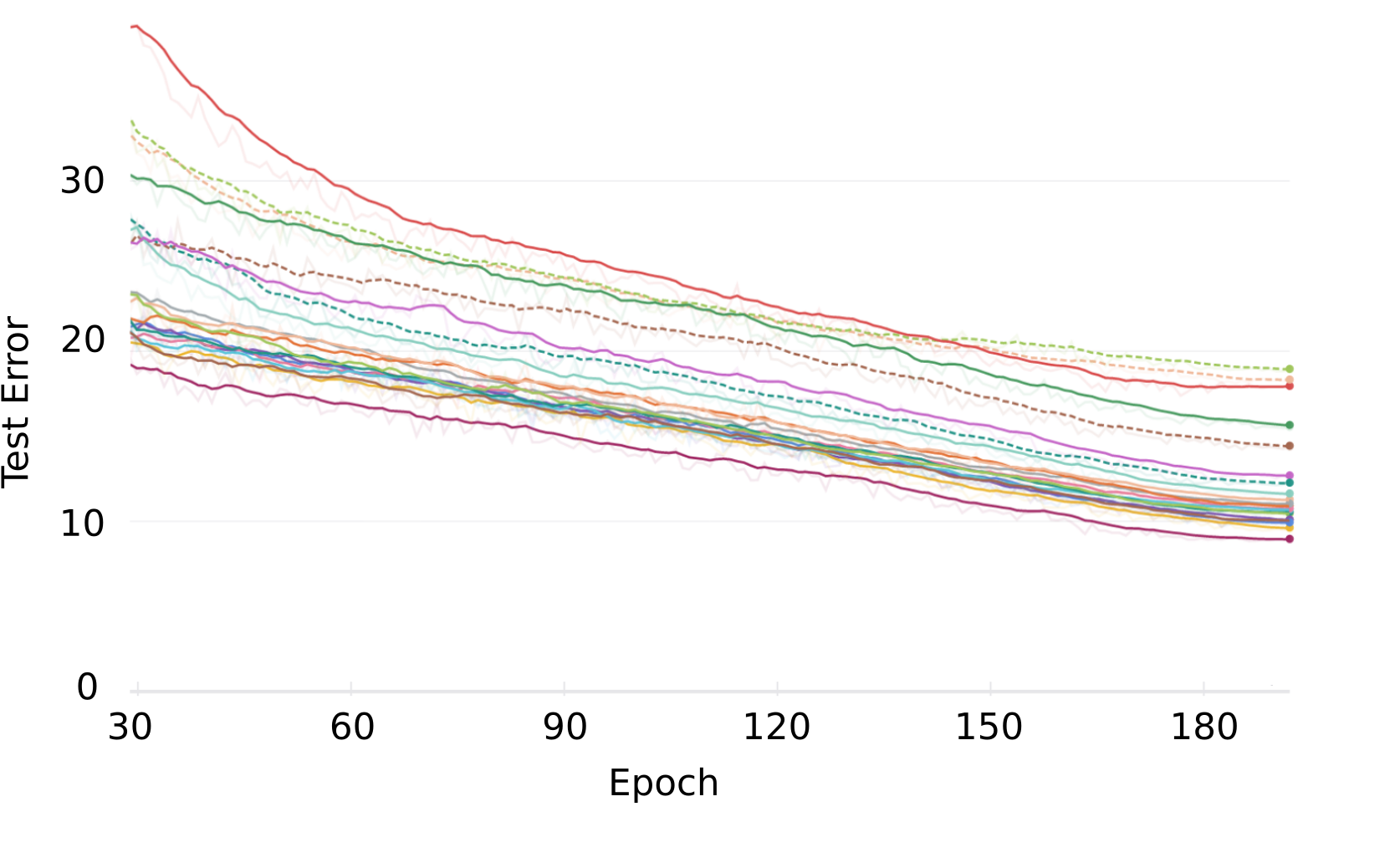} 
    \caption{Cifar10 test error curves of $20$ randomly sampled architectures trained over $200$ epochs (left). The same error curves but cut to epochs $30$ to $200$.}
    \label{fig:cifar10_long_train}
\end{figure}

Since we have a limited amount of computational resources and needed to train a vast number of networks we opted to train the networks up to the number of epochs where they started to saturate significantly in our pre-studies. As we have seen in section \ref{chp:cifar_stability} can the network performance still improve quite a bit if it is trained for much longer. Even though the improved performances on Cifar10 did not yield any results contradicting the findings of our study, we still deemed it necessary to closer inspect what happened in the later stages of training and thus performed a sanity check for Cifar10 as well as the other two datasets that show a negative error correlation with ImageNet---Powerline and Natural.
Figure \ref{fig:cifar10_long_train} shows the Cifar10 test error curves of $20$ randomly selected architectures over $200$ epochs. On the left side we see the same curves zoomed in to epochs $30$ to $200$. We see that the error decreases steadily for all architectures, the ranking among architectures barely changes past epoch $30$. The relative performance between architectures and not absolute error rates are relevant for our evaluations, we can therefore conclude that the errors at epoch $30$ are an accurate enough description of an architecture's power.

\begin{figure}[!htp]
    \centering
    \includegraphics[width=0.35\textwidth]{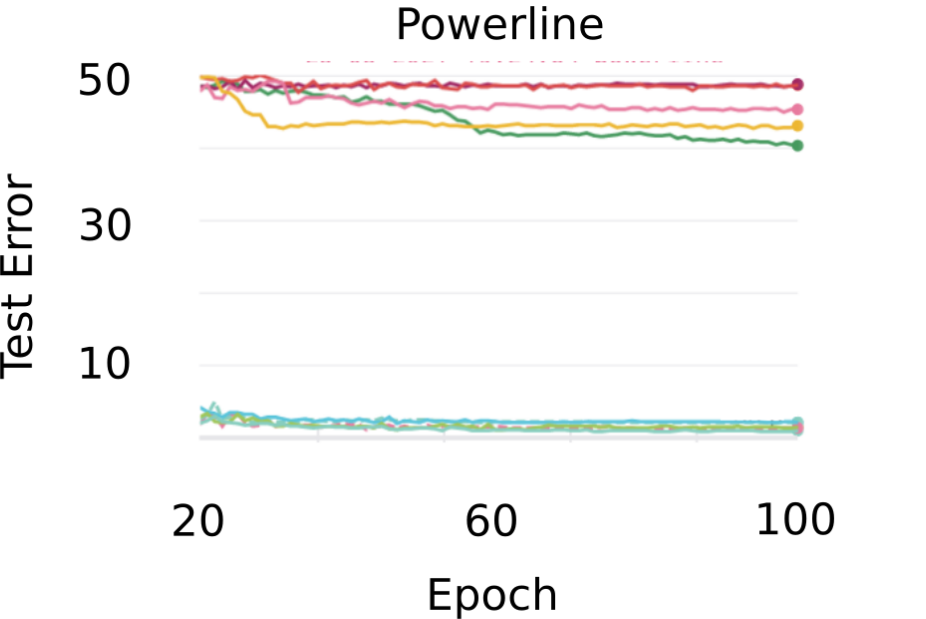} 
    \includegraphics[width=0.35\textwidth]{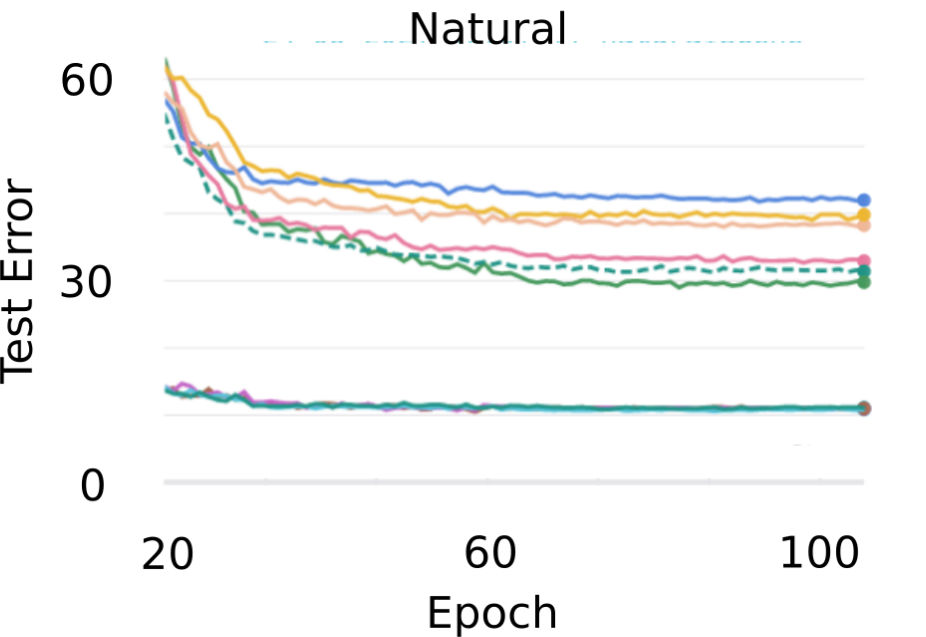} 
    \caption{Test error curves of the five best and five worst models on Powerline and Natural, respectively, when training is continued to epoch 100}
    \label{fig:rest_long_train}
\end{figure}

For Powerline and Natural, we select the five best and five worst architectures respectively and continue training them for a total of five times the regular duration. Figure \ref{fig:rest_long_train} shows the resulting error curves. Both datasets exhibit minimal changes in the errors of the top models. On Natural we observe clear improvements on the bottom five models but similar to Cifar10 there are very little changes in terms of relative performance. Powerline exhibits one clear cross-over but for the remainder of the bottom five models the ranking also stays intact. \textit{Overall we can conclude that longer training does not have a significant effect on the APR of our datasets.}

\subsection{Impact of Training Variability}
\label{chp:model_rand}
\begin{figure}[!htp]
    \centering
    \includegraphics[width=0.40\textwidth]{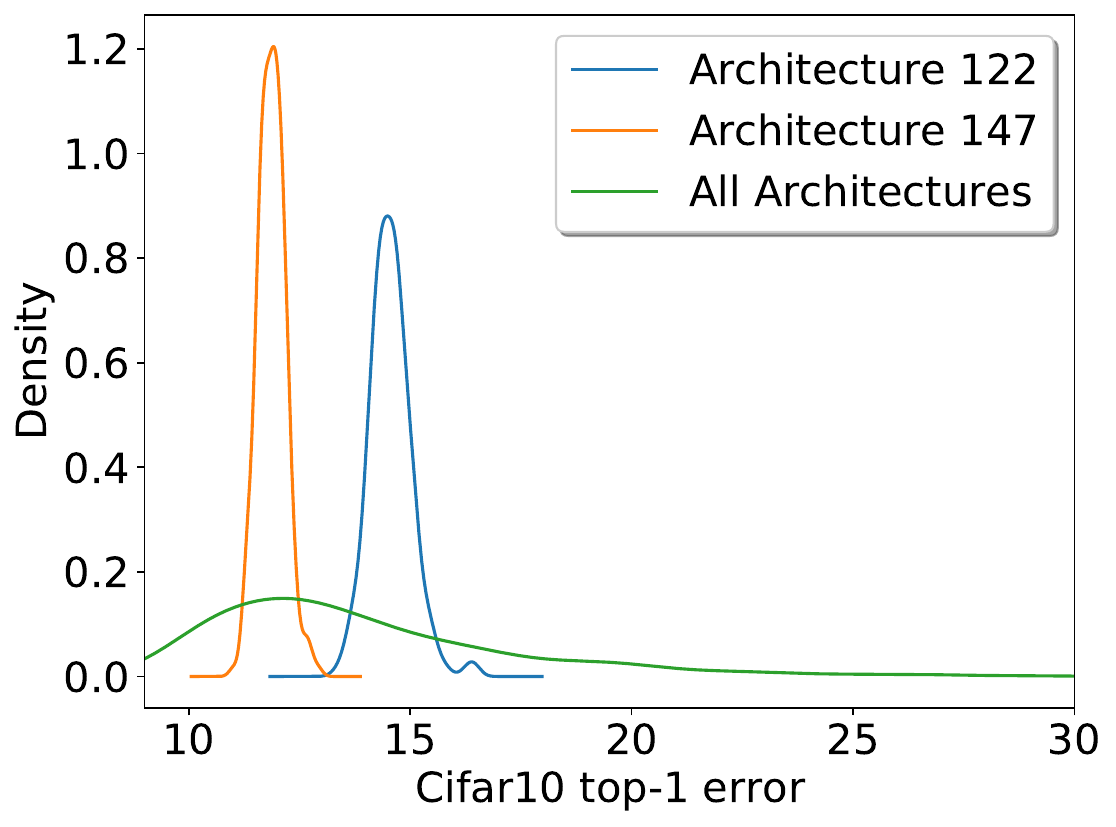} 
    \caption{Error distributions on Cifar10 of two architectures ($122$, $147$) both trained from scratch $250$ times as well as the Cifar10 error distribution of all $500$ architectures. The plot shows that the variability caused by changing architecture is much larger than the one caused by random training effects. }
    \label{fig:cifar10stab}
\end{figure}

The random initialization of the model weights has an effect on the performance of a CNN. In an empirical study it would therefore be preferable to train each model multiple times to minimize this variability. We opted to increase the size of our population as high as our computational resources allow, this way we get a large number of measurements to control random effects as well as an error estimate of a large set of architectures. However, we still wanted to determine how much of the total variability is caused by training noise and how much is due to changing the architectures. We estimate this by selecting two of the sampled CNN designs, number $147$ performing slightly above average with an error of $e_{147}=11.9$ and number $122$ performing slightly below average with $e_{122}=14.5$. The quantiles of the error distribution from all $500$ architectures are $q_{0.25}=11.53$, $q_{0.5}=13.02$ and $q_{0.75}=15.46$ with an overall mean of $\mu=13.9$. We then train the architectures $147$ and $122$ each $250$ times. Figure \ref{fig:cifar10stab} shows the error distributions of both selected architectures as well as the overall distribution obtained from training each of the $500$ architectures once. There is of course some variability within both architectures but both individual architectures produce very narrow densities and show essentially no overlap. \textit{We can therefore conclude that the effect of choosing an architecture is much greater than the variability caused by random training effects.}

\subsection{Relationship of Top-1 with Top-5 Error on ImageNet, Insects and Cifar100}
\begin{figure}[!htp]
    \centering
    \includegraphics[width=0.15\textwidth]{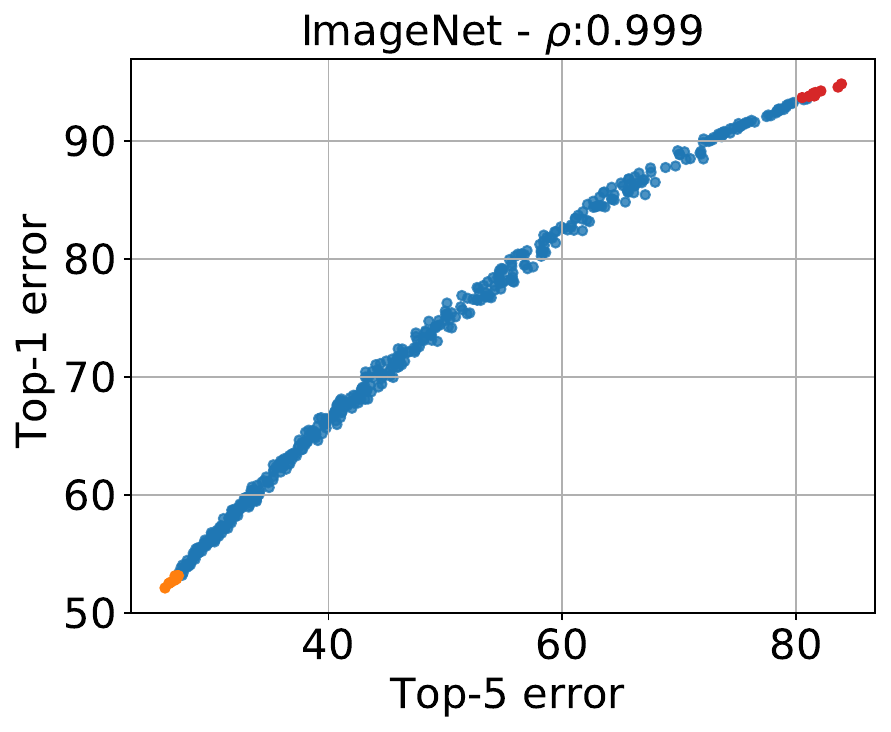} 
    \includegraphics[width=0.15\textwidth]{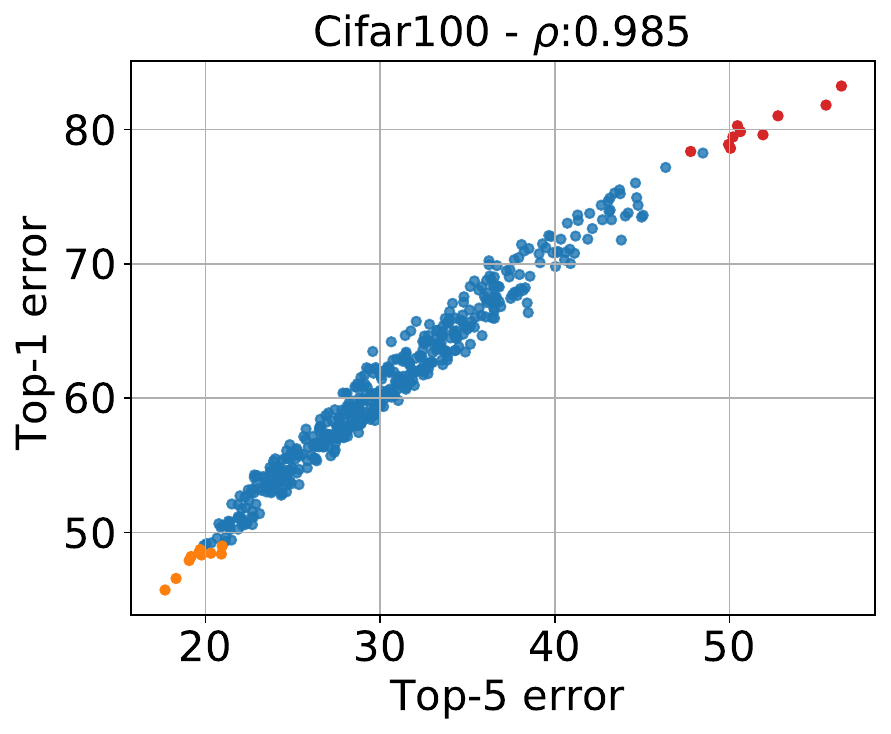} 
    \includegraphics[width=0.15\textwidth]{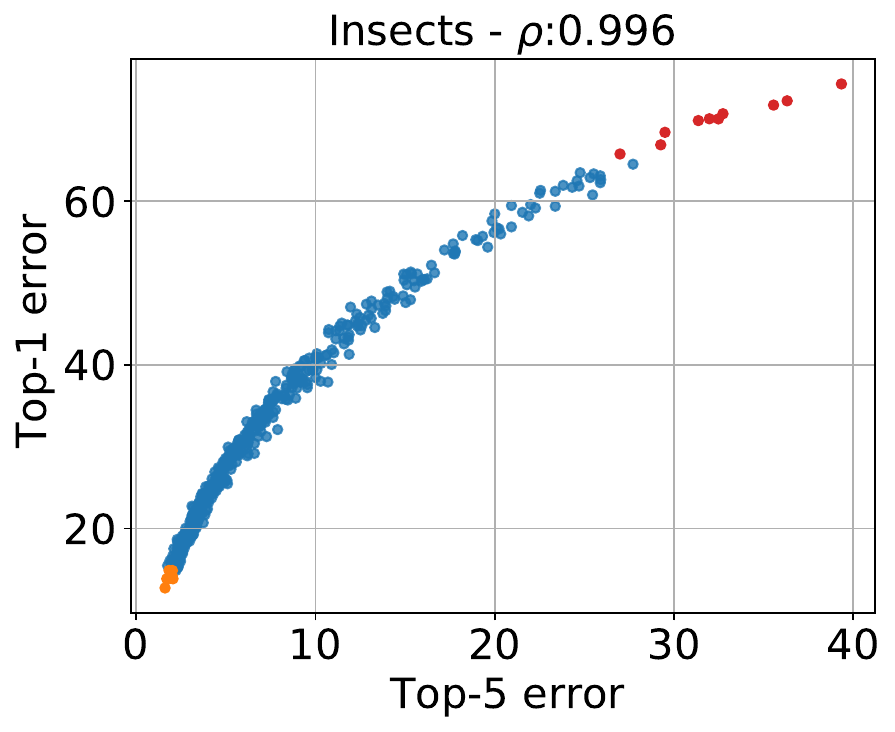} 
    \caption{Top-1 error plotted against top-5 error of all $500$ architectures on ImageNet, Cifar100, and Insects. The plots reveal that on all three datasets the errors have a very close relationship: it is not perfectly linear but is monotonically ascending.}
    \label{fig:top5vstop1}
\end{figure}

We opted to use top-5 error since it is the most widely reported metric for ImageNet and the top-5 numbers are therefore easy to interpret on that dataset. Many of our datasets have a significantly lower number of classes such that top-5 error makes little sense and we opted to use top-1 for those. This raises the question if comparing top-1 with top-5 errors introduces unwanted perturbations into our analysis. We therefore compare the top-1 and top-5 errors for the three datasets on which we use top-1 error (see Figure \ref{fig:top5vstop1}). We see that the two metrics have an almost linear relationship for the ImageNet and Cifar100 datasets. More importantly are the top-1 to top-5 error graphs monotonically ascending for all three datasets, such that the ordering of architectures does not change when swapping between the two metrics. \textit{Since we are interested in the relative performances of our sampled architectures changing between top-1 and top-5 error does not impact our analysis.}

\subsection{Overfitting of High-Capacity Architectures}
\begin{figure}[!htp]
    \centering
    \includegraphics[width=0.15\textwidth]{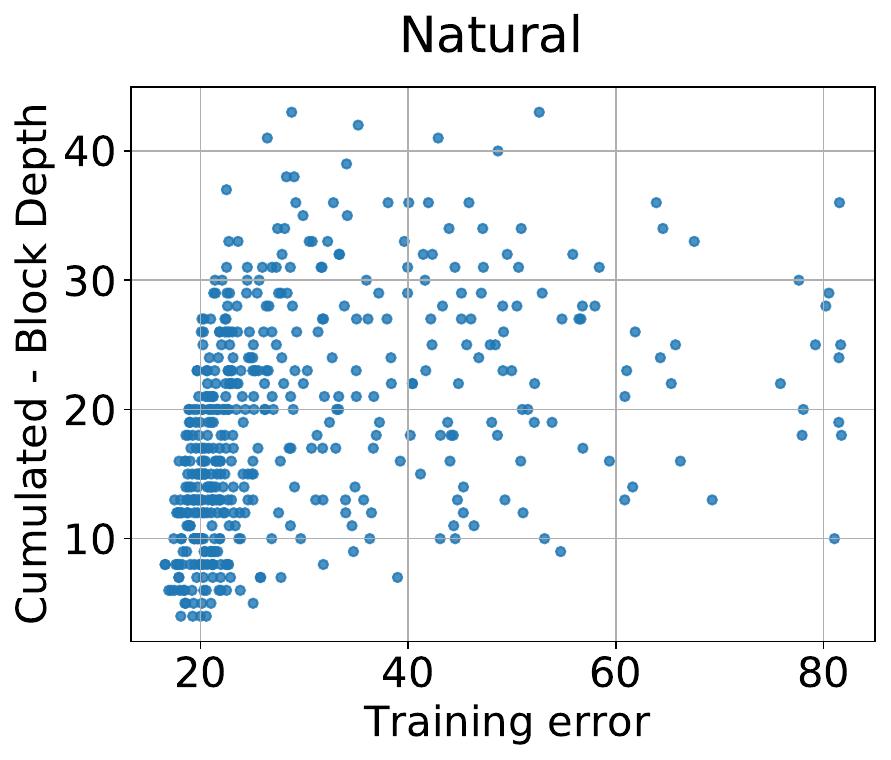} 
    \includegraphics[width=0.15\textwidth]{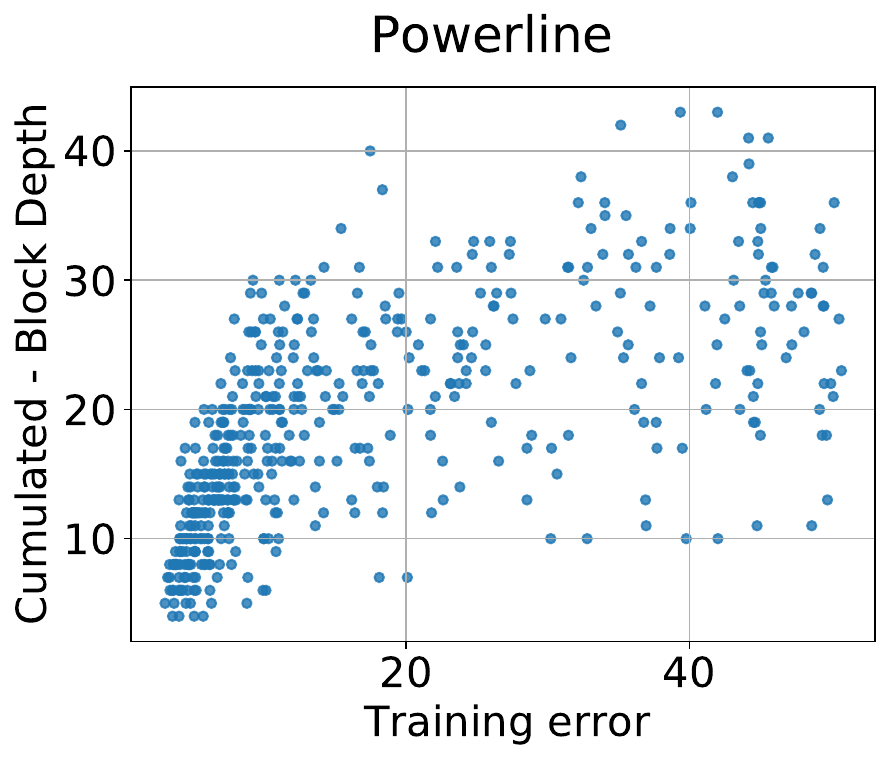} 
    \includegraphics[width=0.15\textwidth]{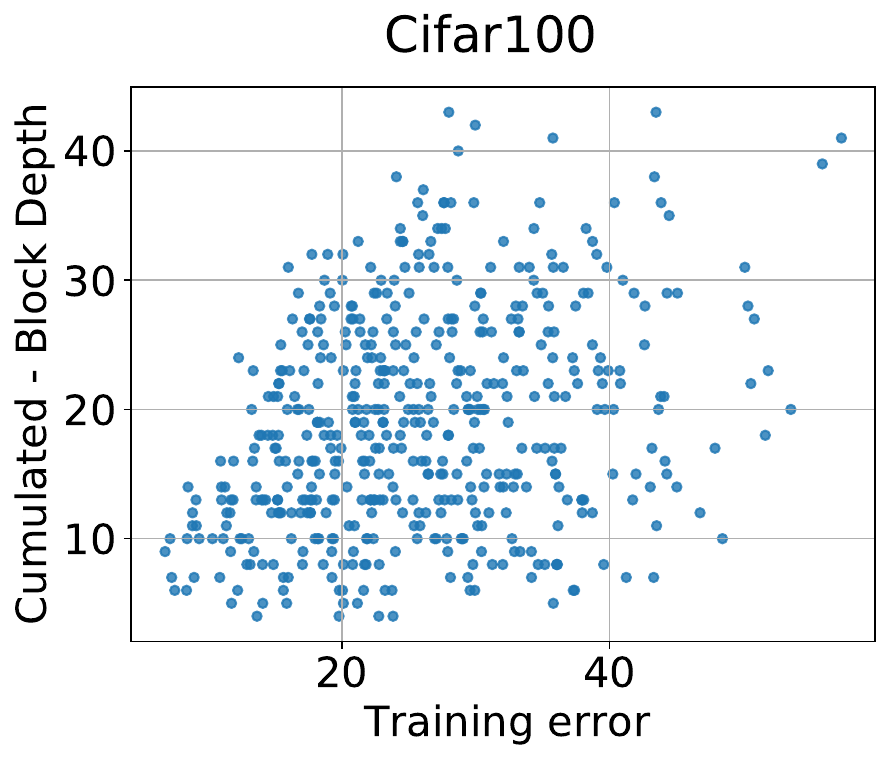} 
    \caption{Training errors of the sampled architectures (x-axis) plotted against the cumulated block \emph{depth} for the $3$ datasets that have the lowest test errors on shallow architectures. We observe that for all three datasets shallow architectures also have the lowest training errors. Therefore overfitting is not the cause of this behaviour.}
    \label{fig:overfitting}
\end{figure}

The best architectures on Powerline, Natural and Cifar100 have a very small cumulated depth, so it is only natural to ask if the deeper architectures perform poorly due to overfitting. We address this concern by plotting the \emph{training errors} of Powerline, Natural, and Cifar100 against the cumulative block depths (see Figure \ref{fig:overfitting}). The training errors are strongly correlated with the cumulative block depth, just like the test errors. \textit{Plots of the cumulated block depth show almost the same structure for training and test errors. We can therefore exclude overfitting as a reason why the shallower networks perform better on Powerline, Natural, and Cifar100.}

\subsection{Disentangling the Effects of Class Count and Dataset Size}
\label{chp:ds_size_vs_class_nr}
\begin{figure}[!htp]
    \centering
    \includegraphics[width=0.2\textwidth]{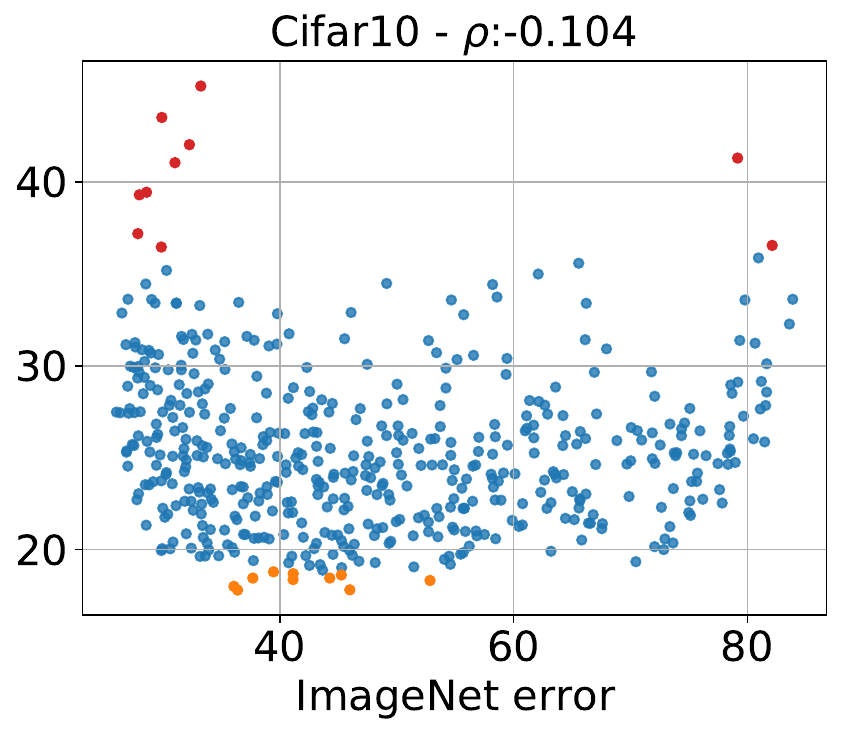} 
    \includegraphics[width=0.2\textwidth]{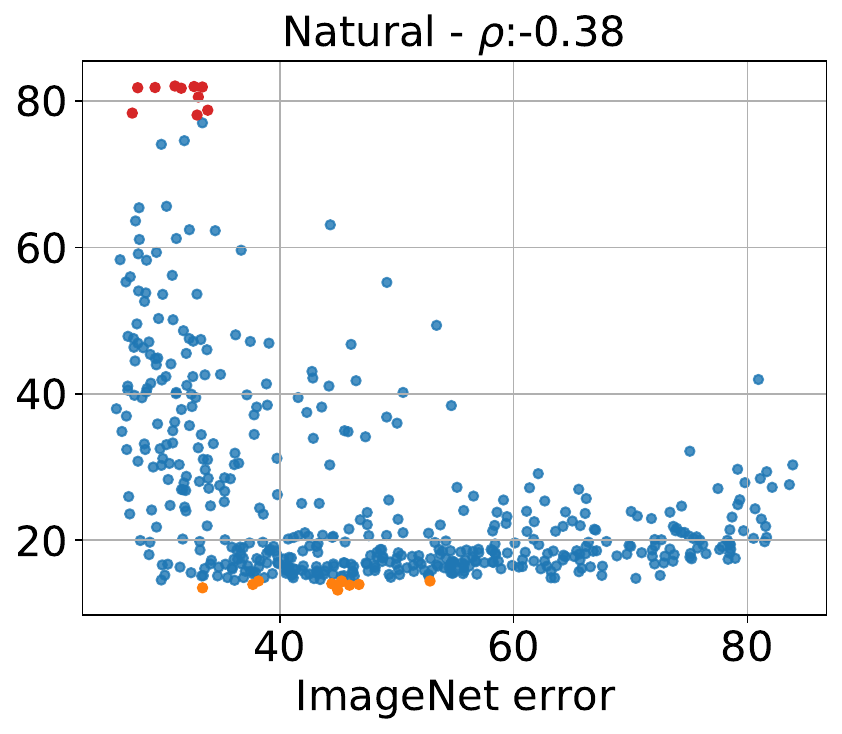} 
    \includegraphics[width=0.2\textwidth]{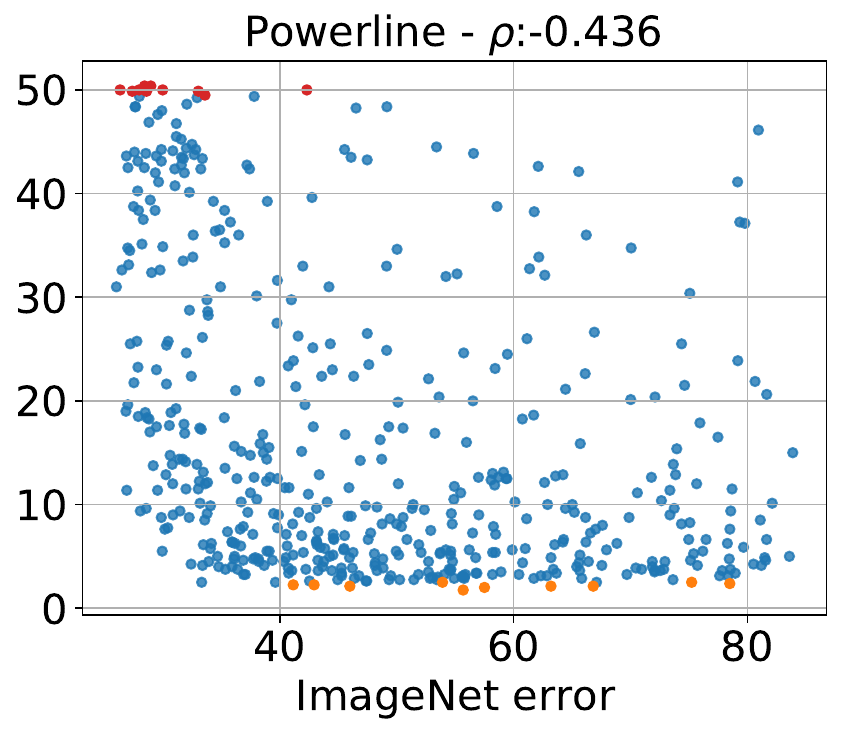}
    \includegraphics[width=0.2\textwidth]{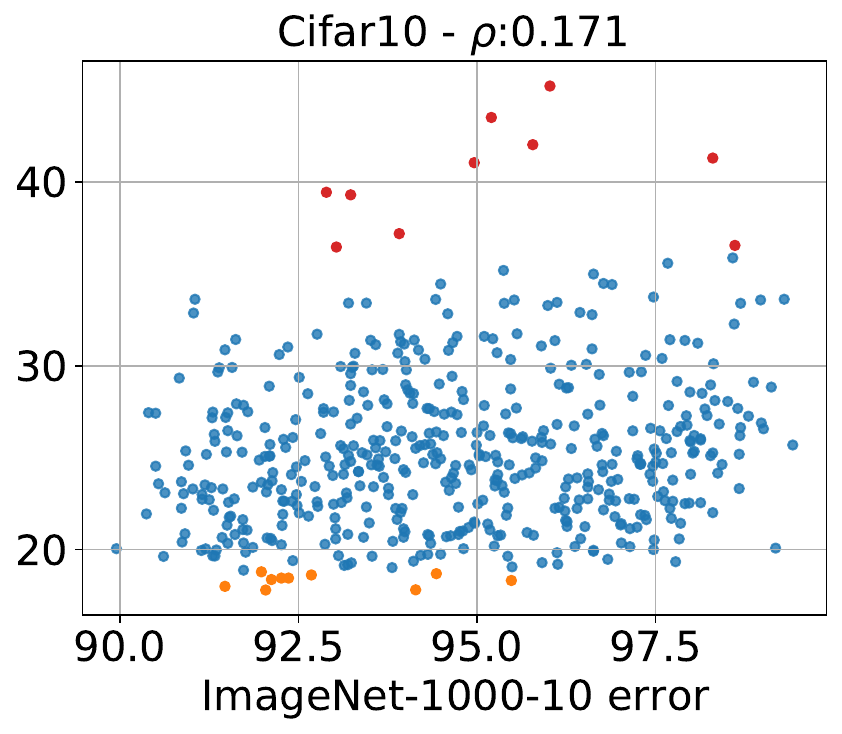} 
    \includegraphics[width=0.2\textwidth]{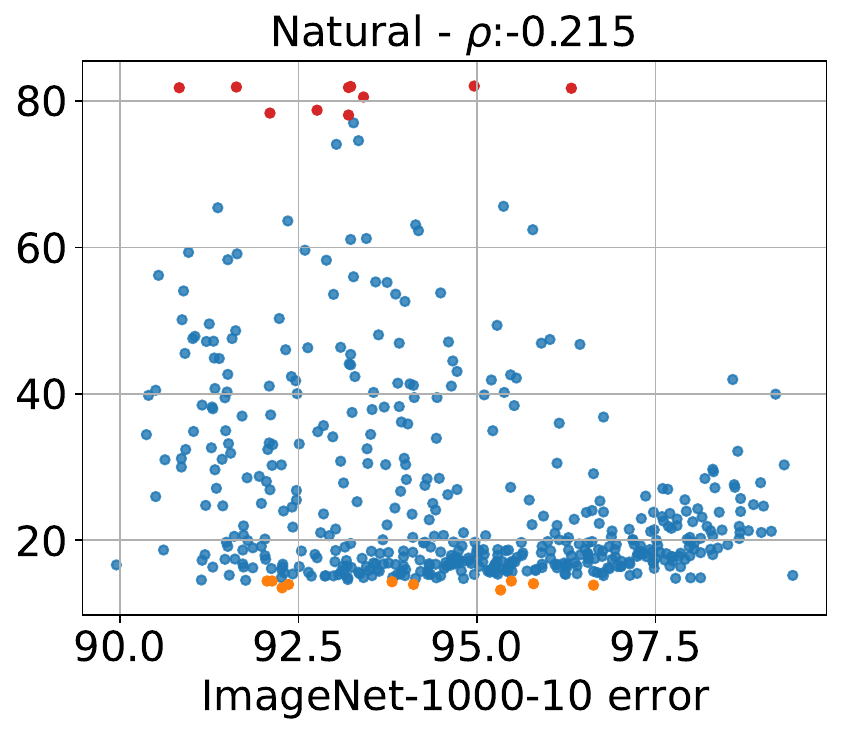} 
    \includegraphics[width=0.2\textwidth]{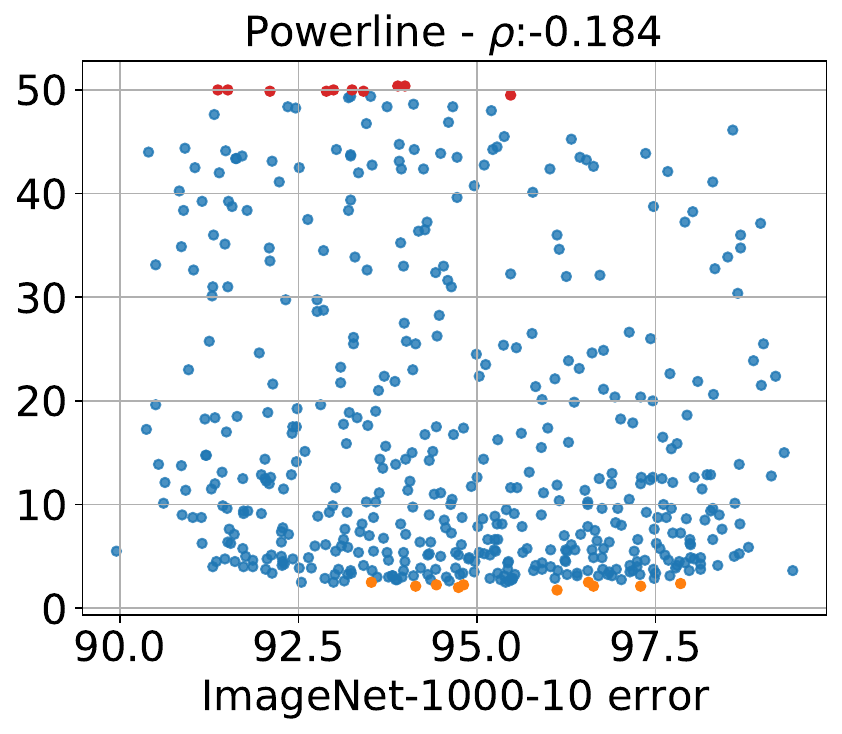}
    \includegraphics[width=0.2\textwidth]{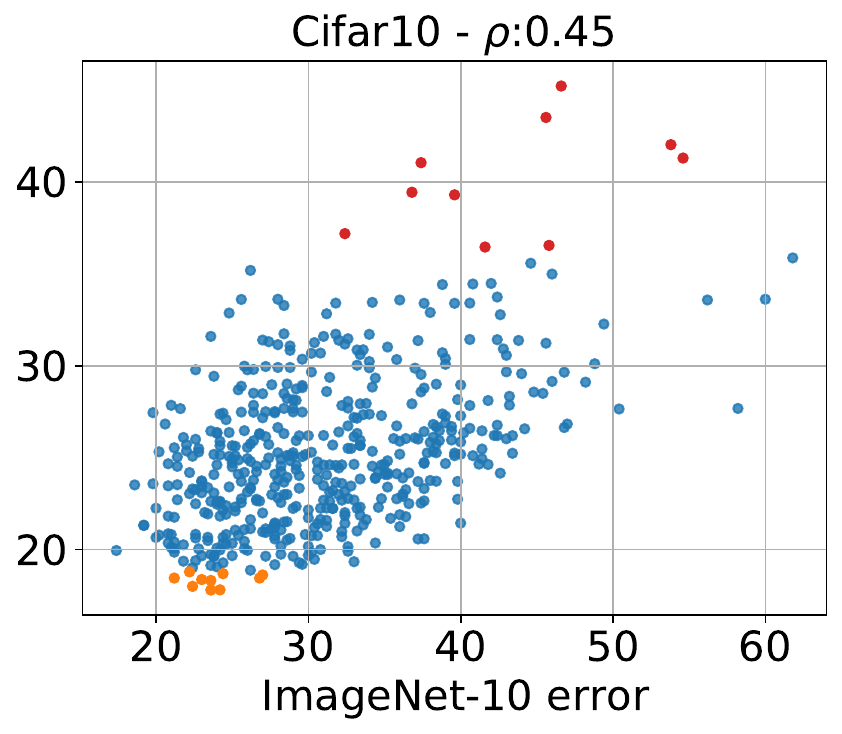} 
    \includegraphics[width=0.2\textwidth]{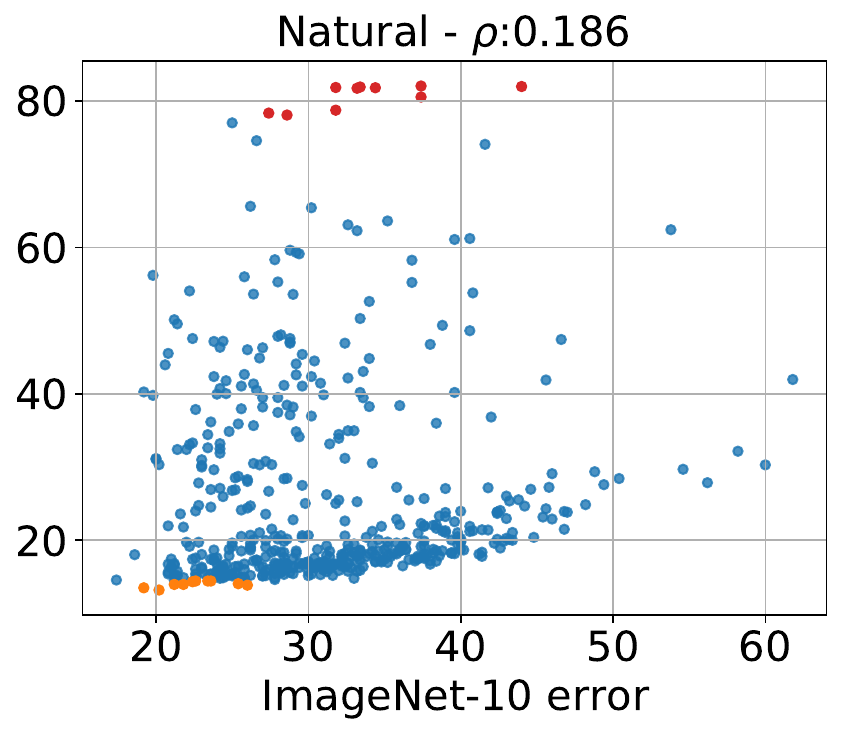} 
    \includegraphics[width=0.2\textwidth]{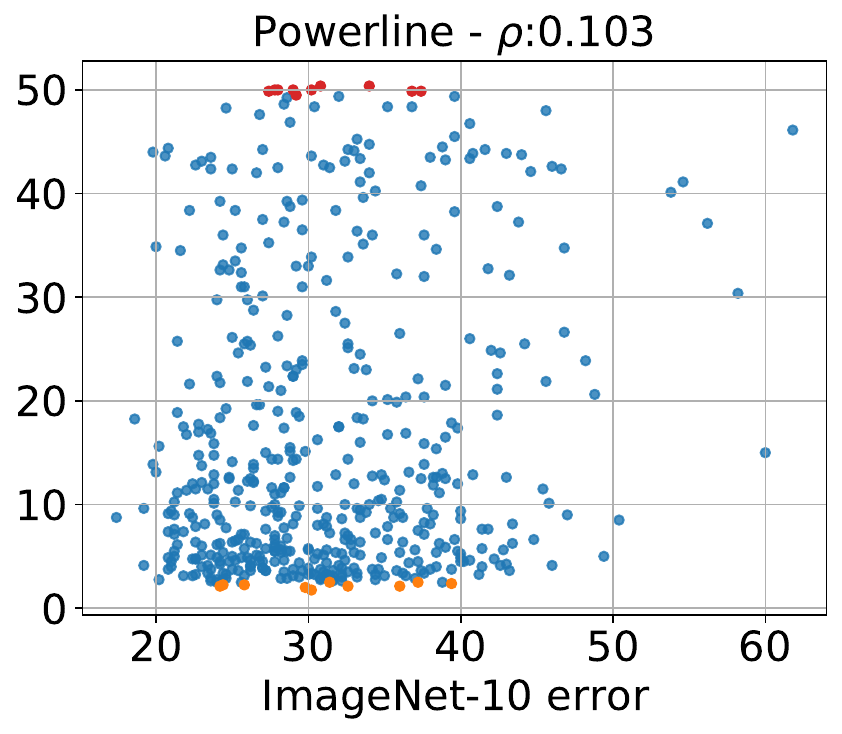} 
    \caption{The errors of all 500 architectures on Cifar10, Natural, and Powerline plotted against the errors on ImageNet (top row), ImageNet-1000-10 (middle row) and ImageNet-10 (bottom row). We observe that class-wise downsampling has the largest positive effect on error correlation.}
    \label{fig:ds_size_vs_class_count}
\end{figure}

A core contribution of our paper is that we show how sub-sampled versions of ImageNet matching the number of classes of the target dataset tend to represent the APR of said target dataset far better. A side effect of downsampling ImageNet to a specific number of classes is that the total number of images present in the dataset also shrinks. This raises the question if the increase in error correlation is actually due to the reduced dataset size rather than to the matching class count. We disentangle these effects by introducing another downsampled version of ImageNet, Imagenet-1000-10. It retains all $1000$ classes but only $10$ examples per class resulting in a datastet with the same number of classes as ImageNet but with the total number of images of ImageNet-10. We train our population of architectures on ImageNet-1000-10 and show the error relationship of Cifar10, Natural, and Powerline with ImageNet-1000-10 (as well as with ImageNet and ImageNet-10 as a reminder) in Figure \ref{fig:ds_size_vs_class_count}. The plots show that there are some correlation gains by using ImageNet-1000-10 over ImageNet, but the effect is far lower compared to ImageNet-10. \textit{This shows that downsampling size has a minor positive effect but the majority of the gain in APR similarity achieved trough class downsampling actually stems from the reduced the class number.}

\subsection{Impact of Class Distribution}
\label{chp:ds_distrib}
\begin{figure}[t]
    \begin{center}
        \includegraphics[height=0.17\textwidth]{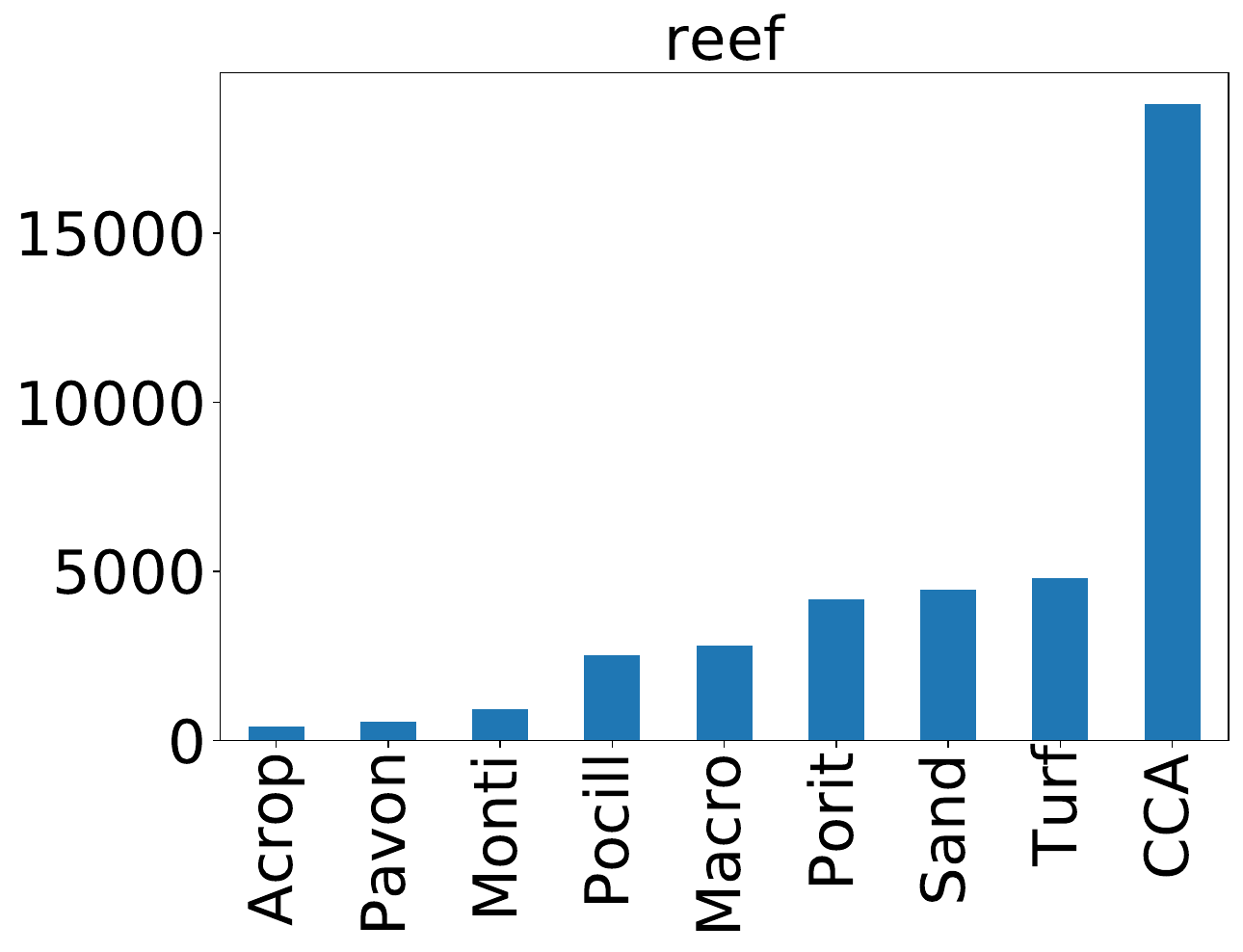}
        \includegraphics[height=0.17\textwidth]{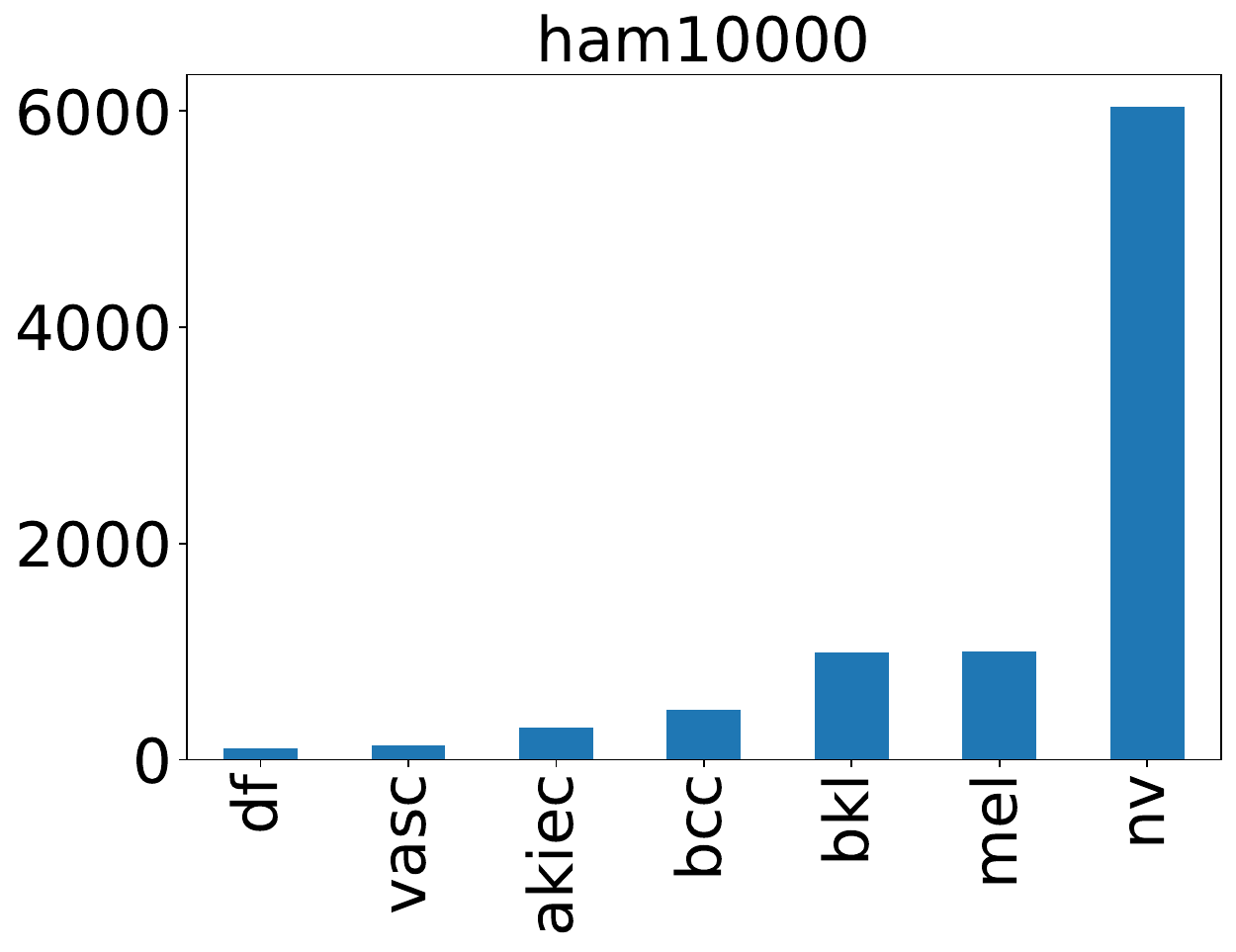}
        \includegraphics[height=0.17\textwidth]{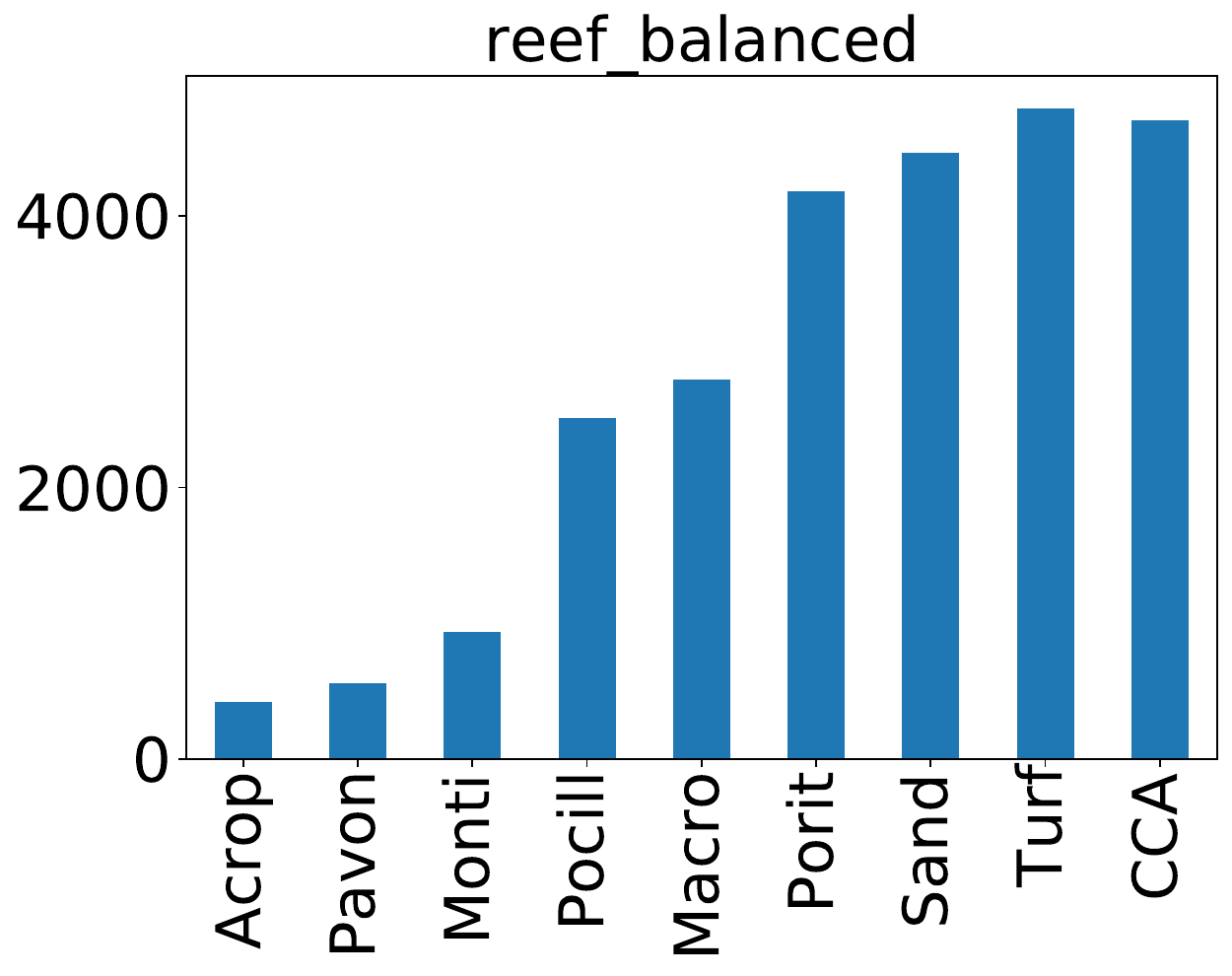}
        \includegraphics[height=0.17\textwidth]{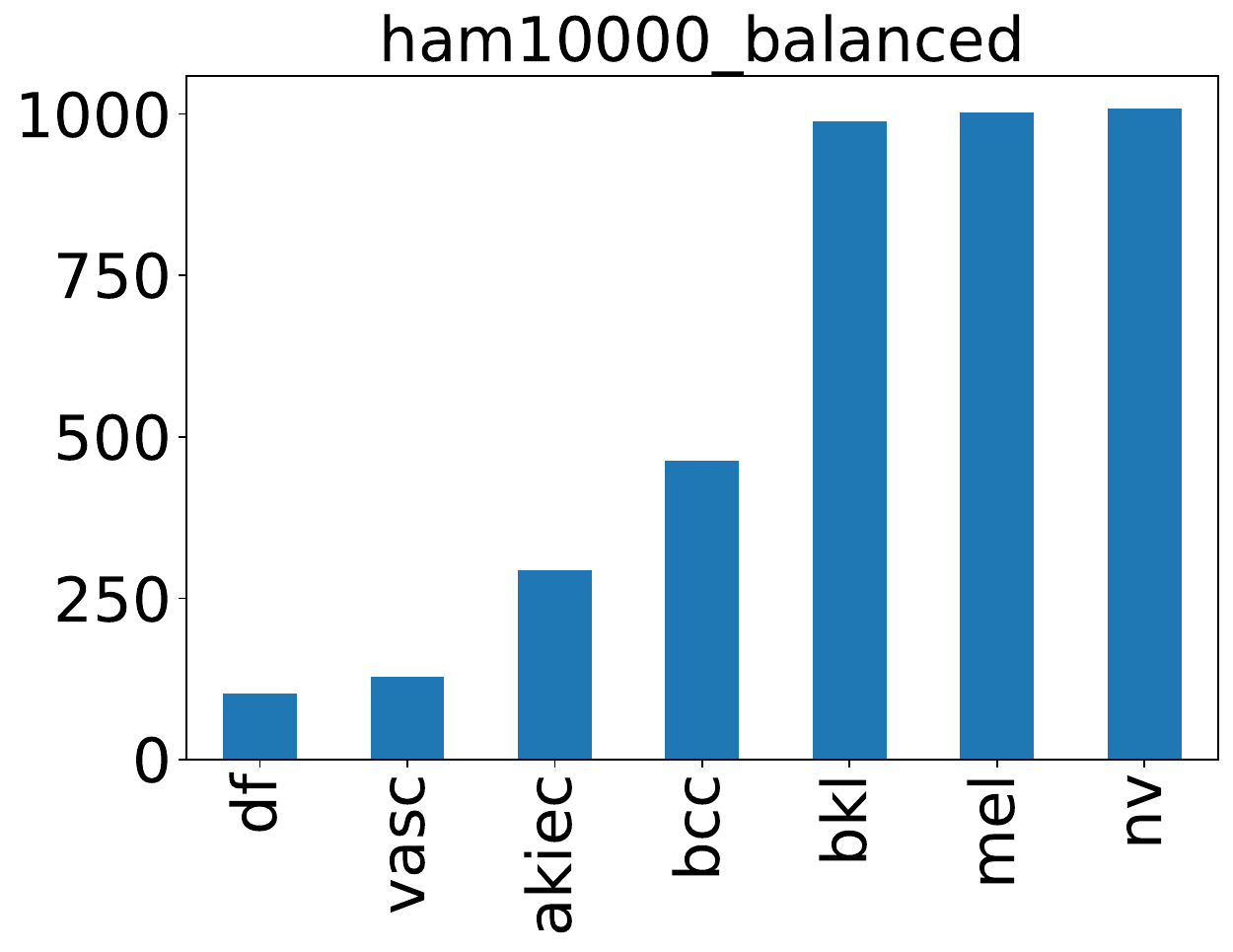}
        \caption{Class distributions of MLC2008, HAM10000, and their balanced versions.}
        \label{fig:class_distributions}
    \end{center}
\end{figure}

\begin{figure}[t]
    \centering
    \includegraphics[width=0.23\textwidth]{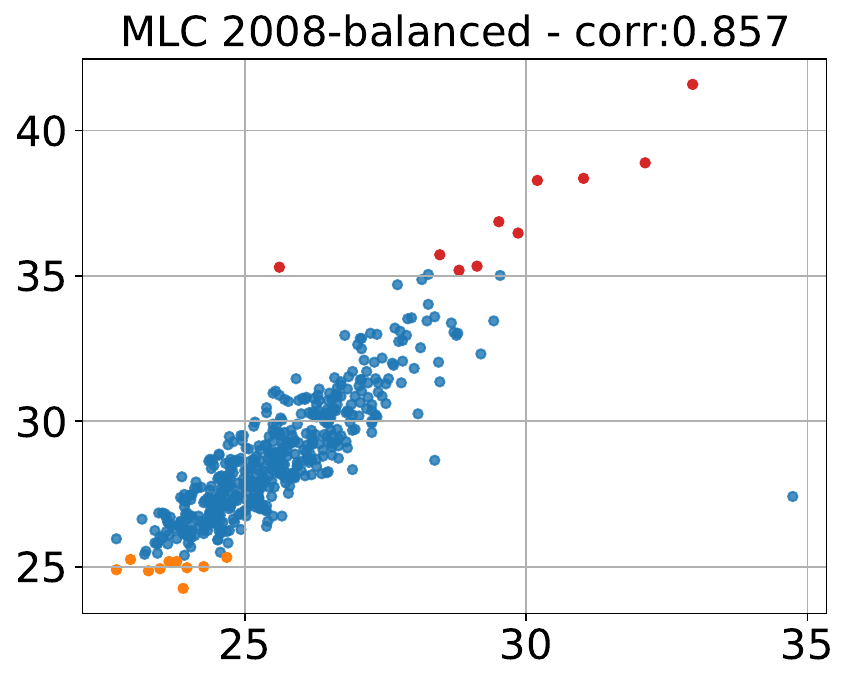} 
    \includegraphics[width=0.23\textwidth]{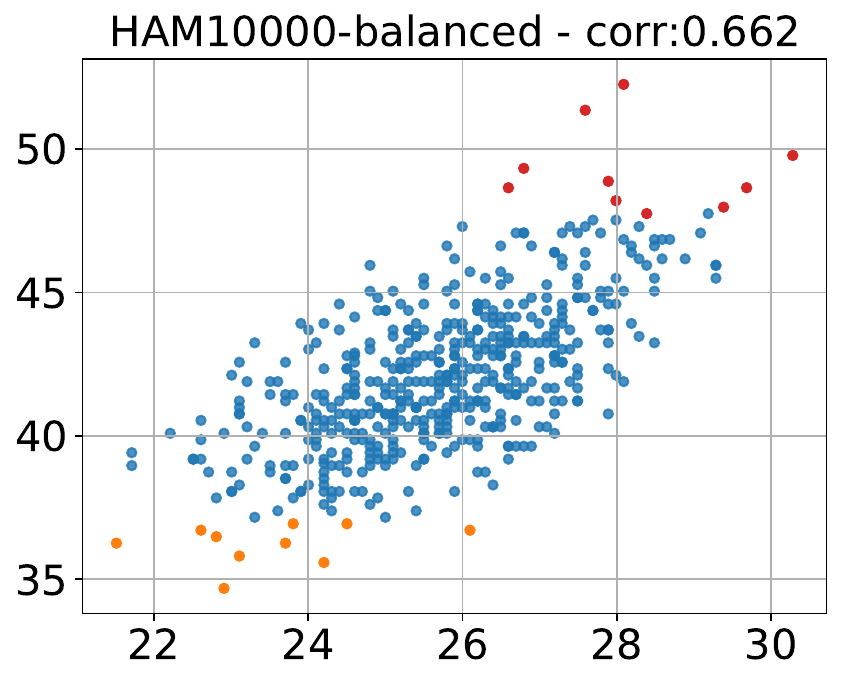} 
    \caption{Errors of all $500$ sampled architectures on MLC2008-balanced and HAM1000-balanced (y-axis) plotted against the errors of their unbalanced counterparts (x-axis). The top $10$ performances on the target dataset are plotted in orange, the worst $10$ performances in red. We observe a clear positive correlation for both datasets, hence we conclude that the dataset imbalance has a limited impact on the APRs.}
    \label{fig:ds_balanced}
\end{figure}

MLC2008 and HAM1000 have a strong class imbalance. They both have one class which makes up a large amount of the dataset. In order to study the impact of an imbalanced class distribution, we created two new more balanced datasets out of the existing data the following way: we reduced the number of samples in the overrepresented class such that it has the same amount of samples as the second most common class. We call these datasets MLC2008-balanced and HAM10000-balanced. Their new class distributions can be seen in Figure \ref{fig:class_distributions}. We train our architecture population on MLC2008-balanced and HAM10000-balanced leaving the training configuration otherwise unaltered. Figure \ref{fig:ds_balanced} shows the errors on the balanced datasets versus the errors on the unbalanced counterparts. \par

For both HAM10000 and and MLC2008, there is a strong correlation between the errors on the balanced and unbalanced datasets. \textit{We can therefore conclude that class imbalance is not a determining factor for the APRs of HAM10000 or MLC2008.}

\balance
\section{Additional ablation studies}
\label{chp:ablation}
\subsection{Impact of Pretraining}
\label{chp:pretrain}
\begin{figure}[t]
    \begin{center}
        \includegraphics[height=0.17\textwidth]{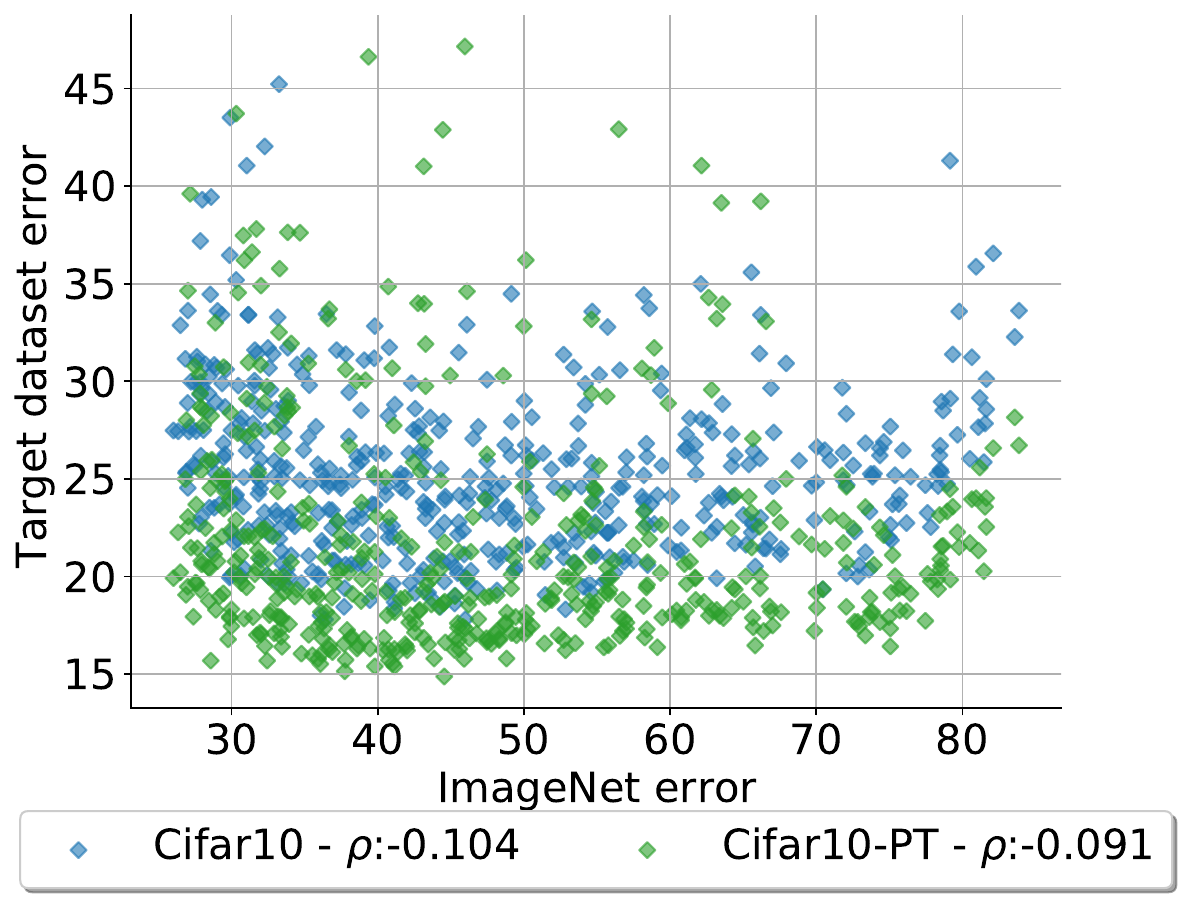}
        \includegraphics[height=0.17\textwidth]{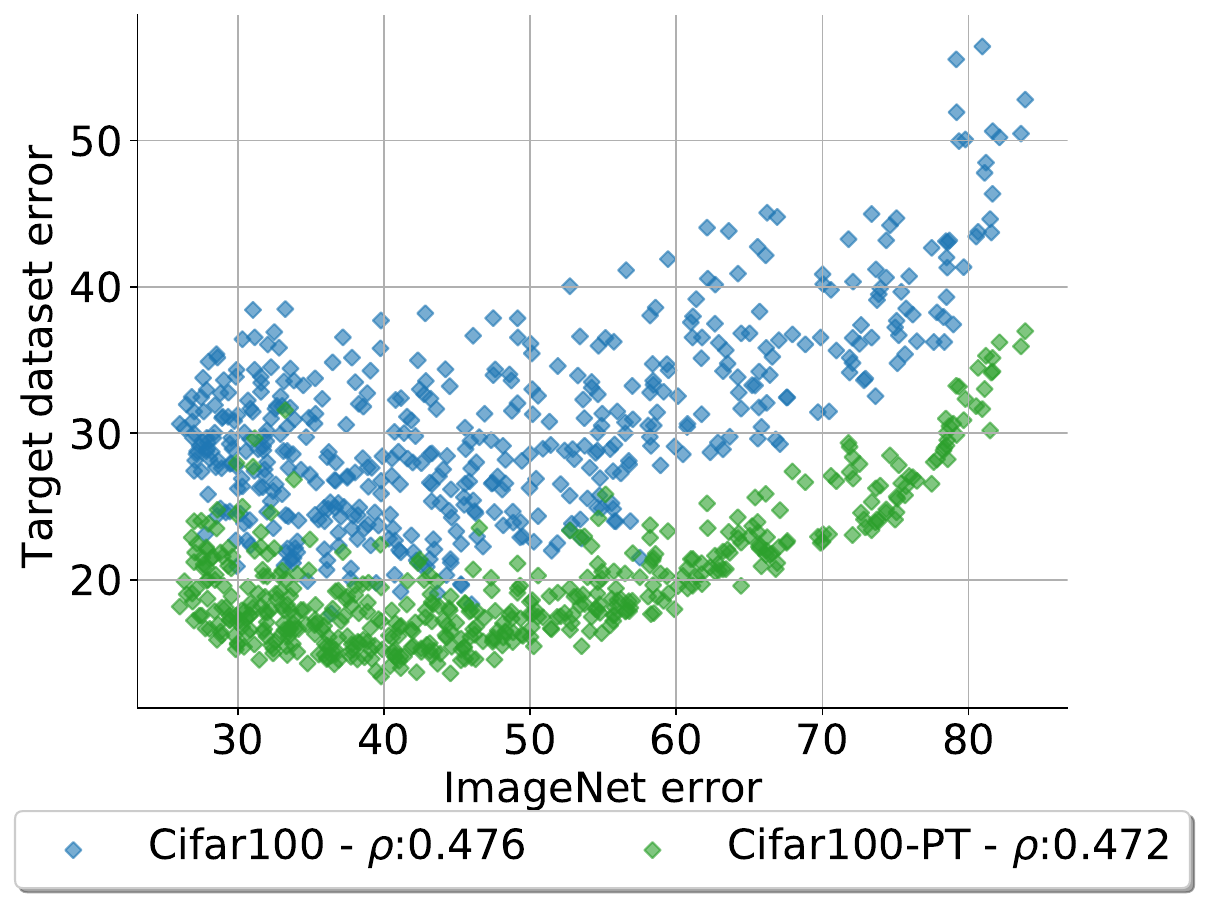}
        \includegraphics[height=0.17\textwidth]{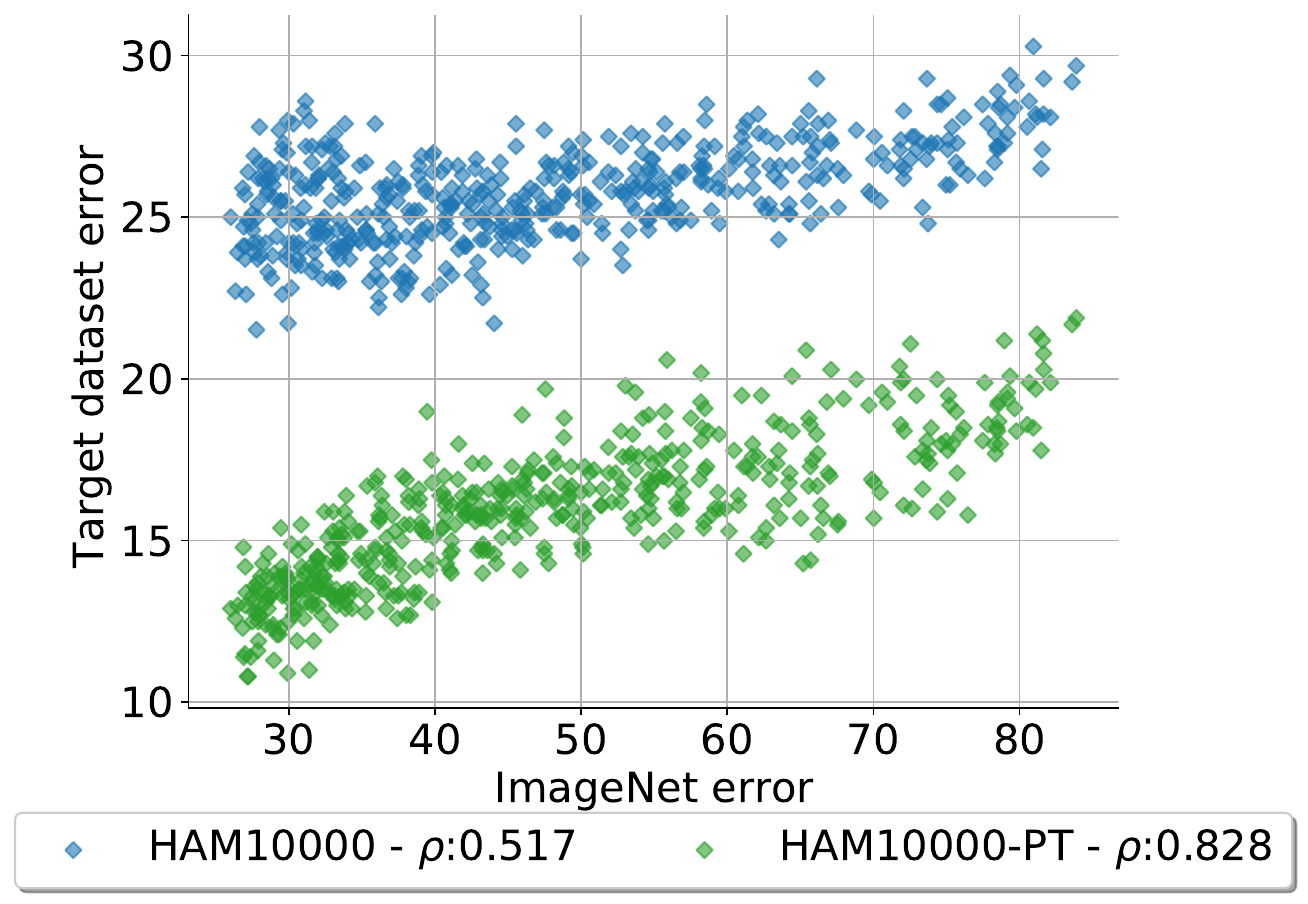}
        \includegraphics[height=0.17\textwidth]{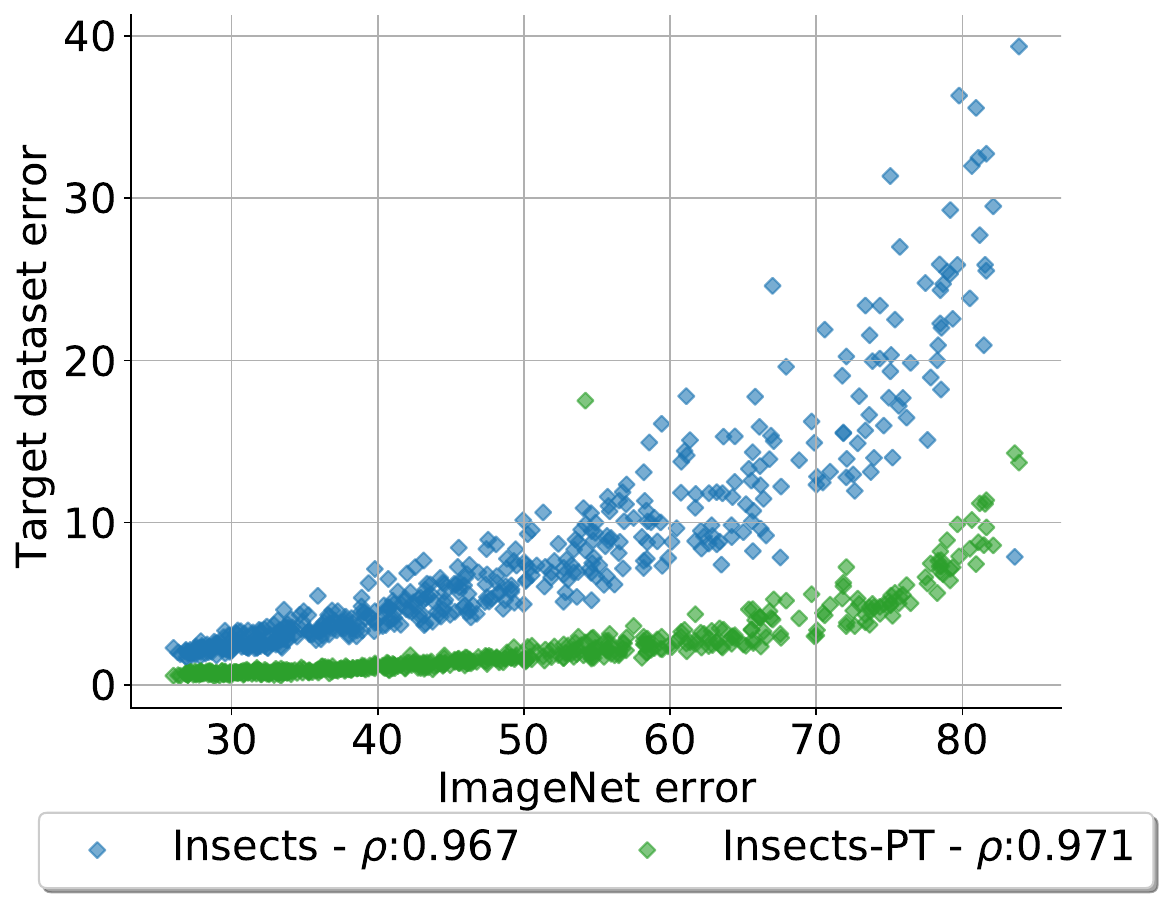}
        \includegraphics[height=0.17\textwidth]{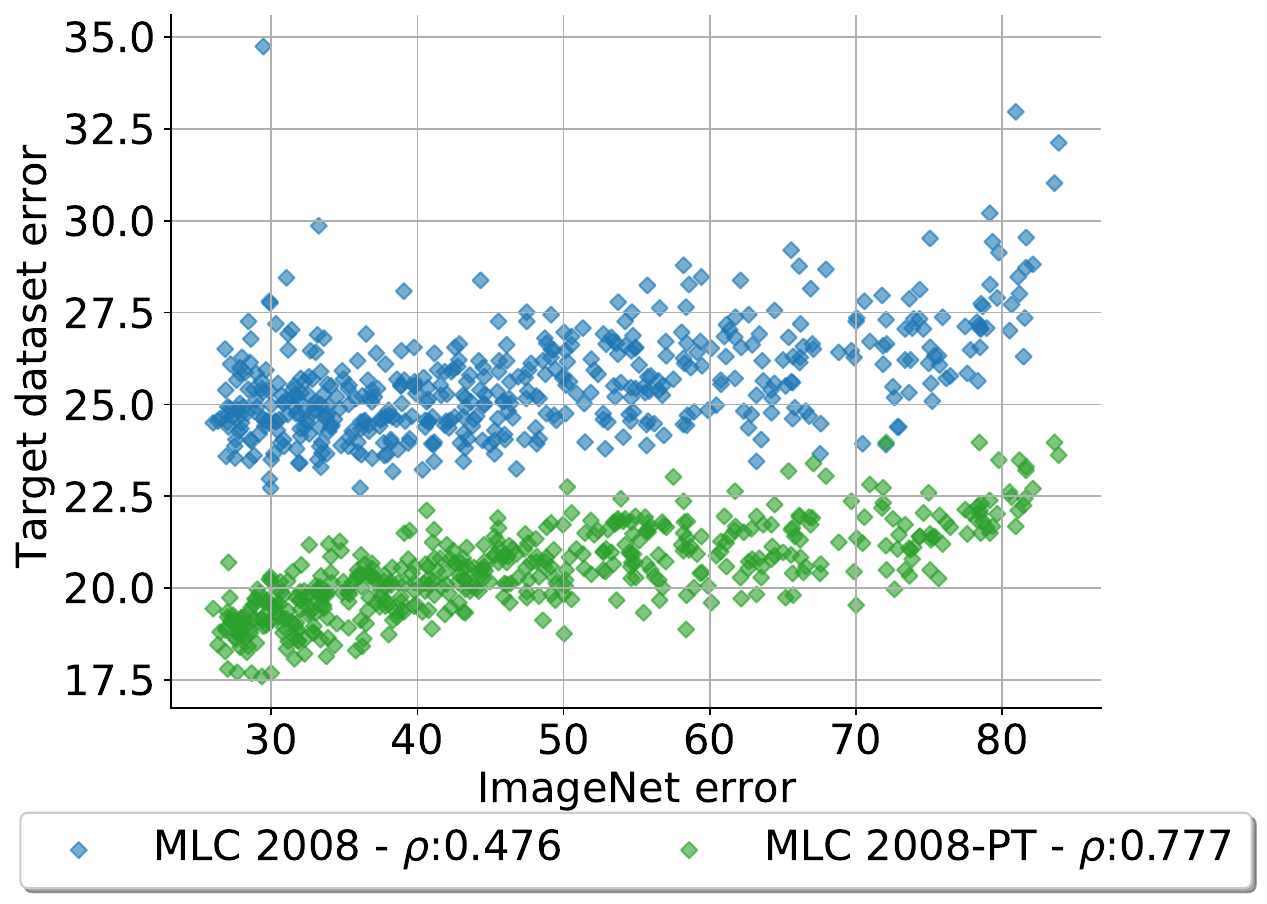}
        \includegraphics[height=0.17\textwidth]{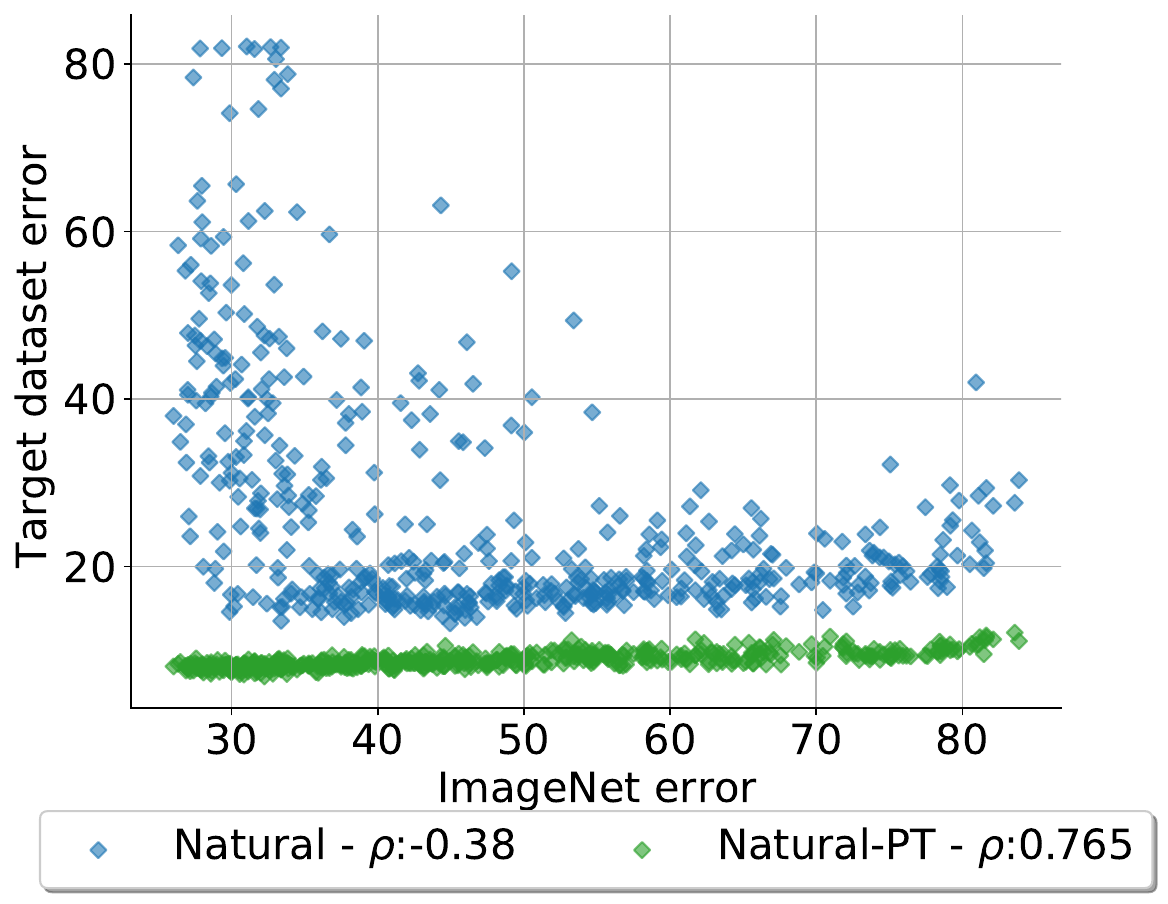}
        \includegraphics[height=0.17\textwidth]{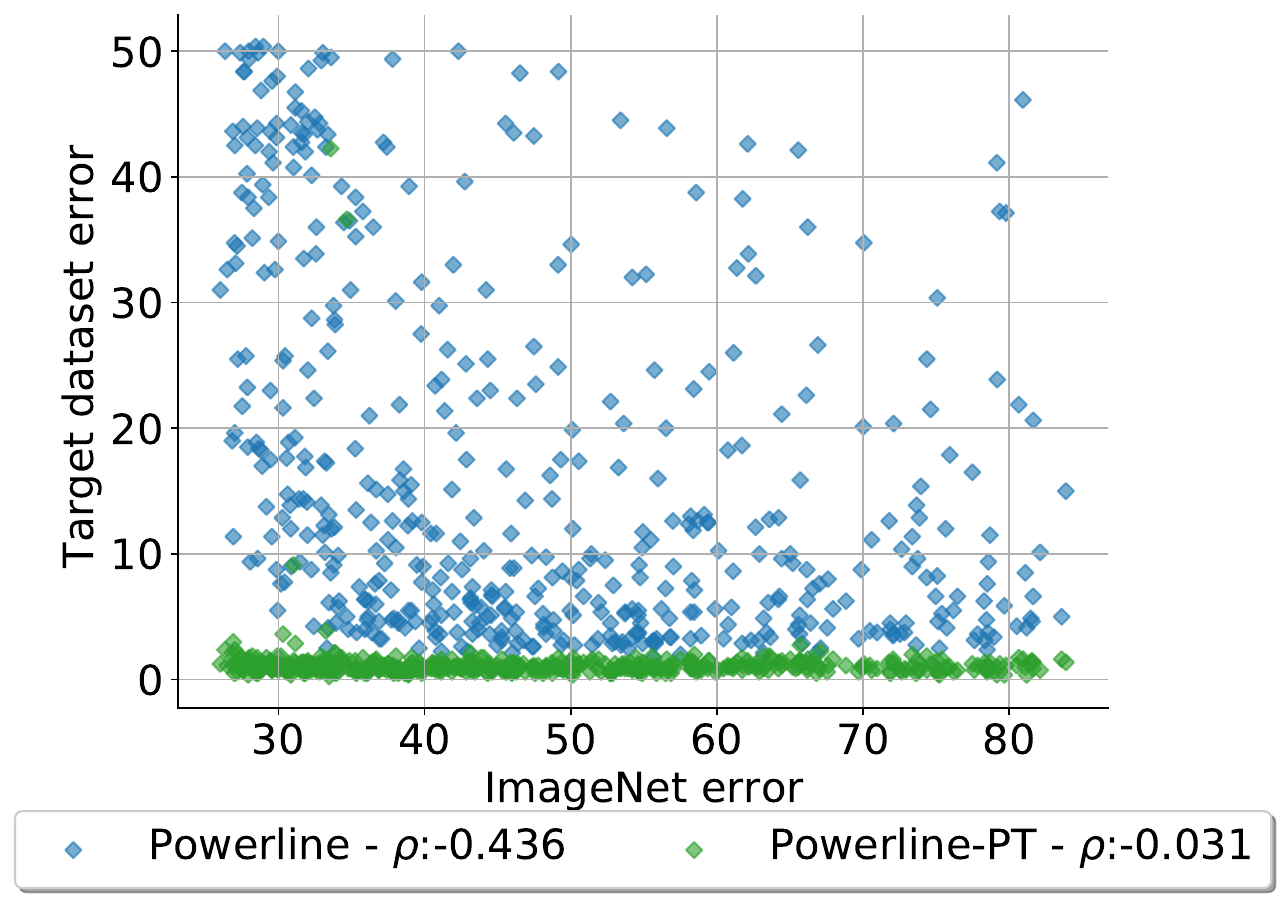}

        \caption{Errors form all 500 architectures trained from scratch (blue) as well as the same architectures pretrained on ImageNet (green), plotted against the respective ImageNet errors. We observe that the error correlation with ImageNet increases relative to the performance gain due to pretraining.}
        \label{fig:pretraining}
    \end{center}
\end{figure}

The main objective of our study is to identify how well different CNN designs perform on varying datasets and if the best architectures are consistent across the datasets. For this reason we train all of our networks from scratch on each dataset. However, we cannot ignore that pretraining on ImageNet is a huge factor in practice and we therefore study its impact on our evaluations. To this end have we train all of our sampled architectures again on each dataset but this time we initialize their weights with ImageNet pretraining (we omit Concrete, which has very low errors even without pretraining). Figure \ref{fig:pretraining} shows the errors of each dataset without (blue) and with (green) pretraining plotted against the ImageNet errors. The data shows a distinct trend: 
 the overall performance improvement due to pretraining dictates how much stronger the ImageNet-correlation of the pretrained errors is compared to the errors without pretraining.
For Cifar10 and Cifar100 where the performance gain with pretraining is low to moderate the error correlations do not drastically change. On the other end of the spectrum are Natural and Powerline, where pretraining leads to drastically lower errors. This in turn leads to much higher error correlation with ImageNet(the Powerline correlation can not grow significantly above 0 because the overall errors are so small across all architectures). \textit{We can conclude that our findings are still valid when pretraining is used, but their effects can be masked when pretraining is the most important factor contributing to the overall final performance.}

\subsection{Structure of Top Performing Architectures}
\label{chp:top_arch}
\begin{figure*}[h]
    \centering
    \includegraphics[width=0.49\textwidth]{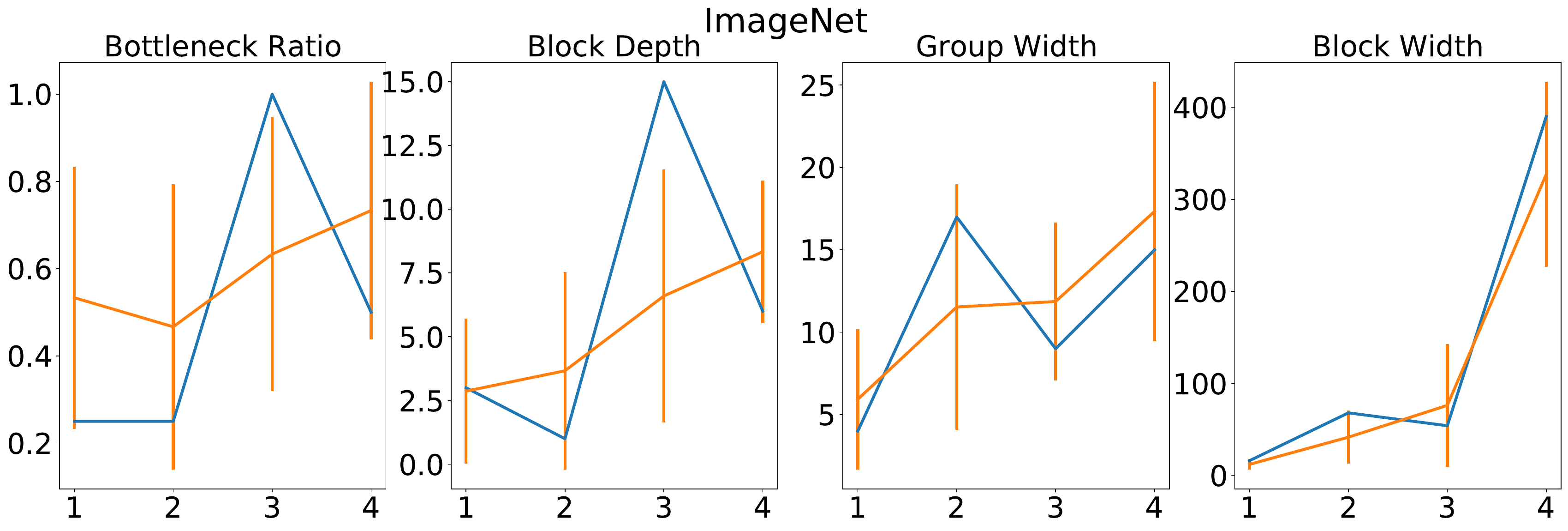} \\
    \includegraphics[width=0.49\textwidth]{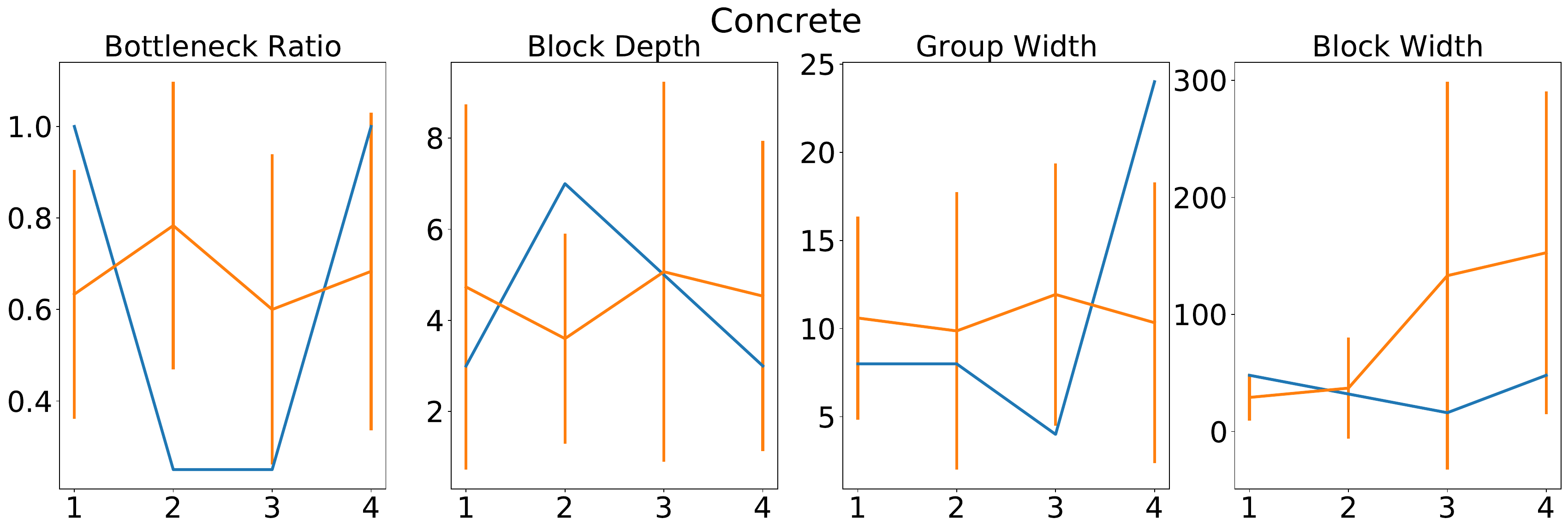} 
    \includegraphics[width=0.49\textwidth]{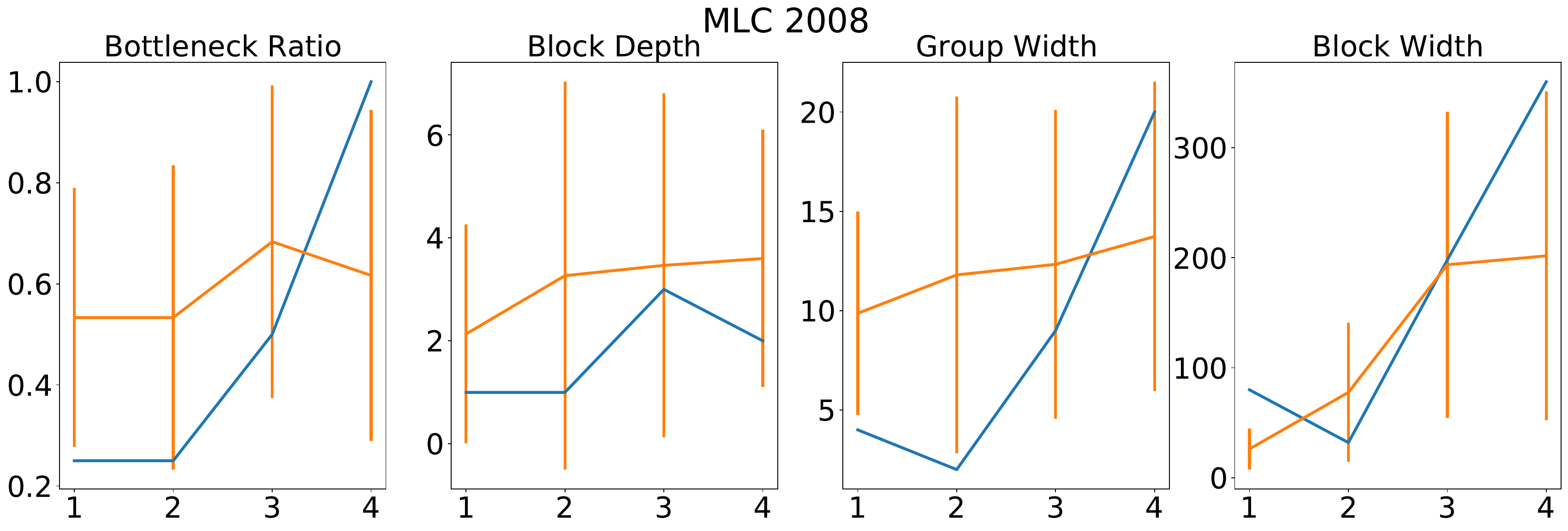} 
    \includegraphics[width=0.49\textwidth]{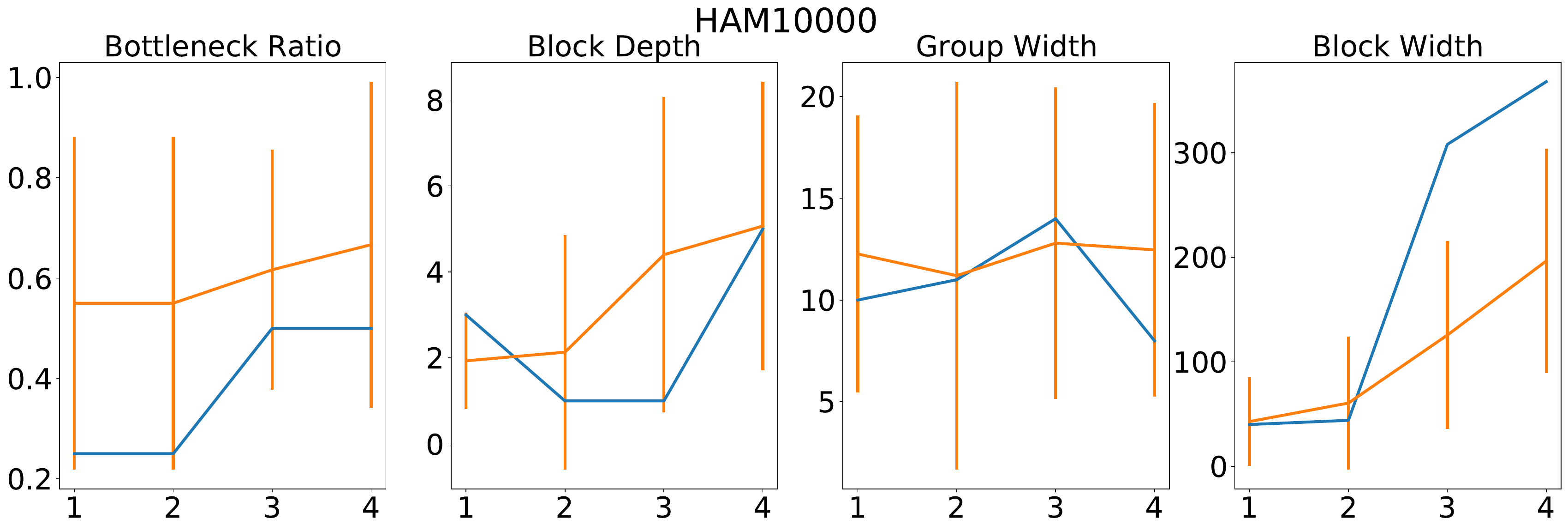} 
    \includegraphics[width=0.49\textwidth]{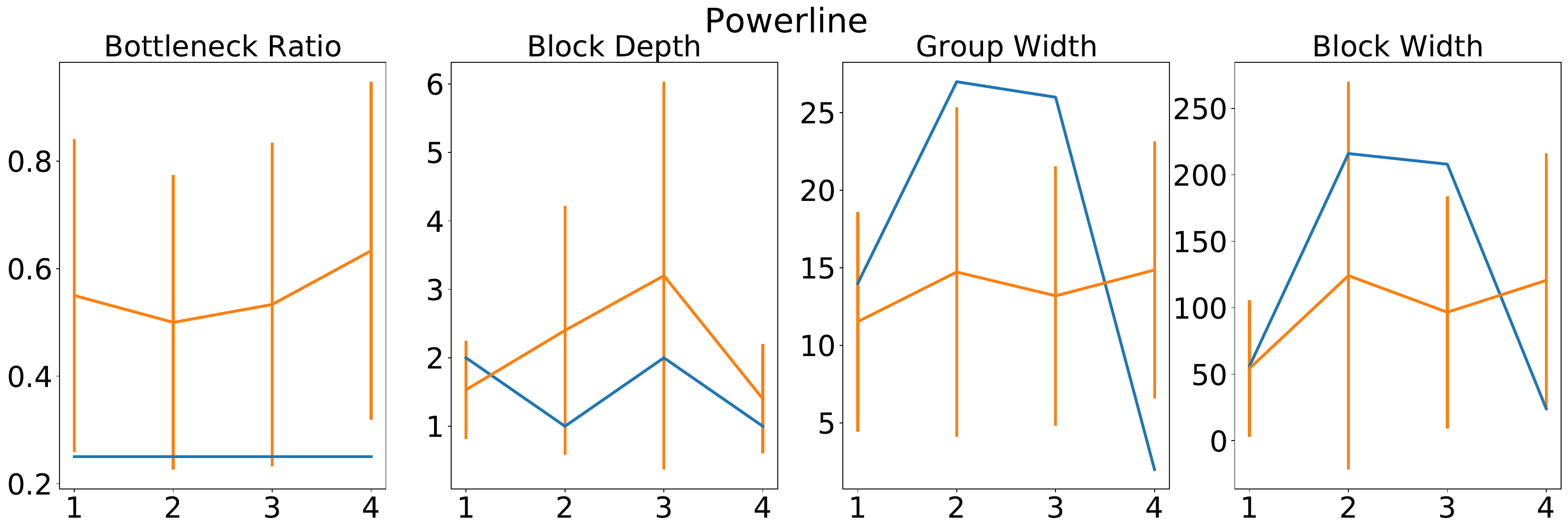}

    \includegraphics[width=0.49\textwidth]{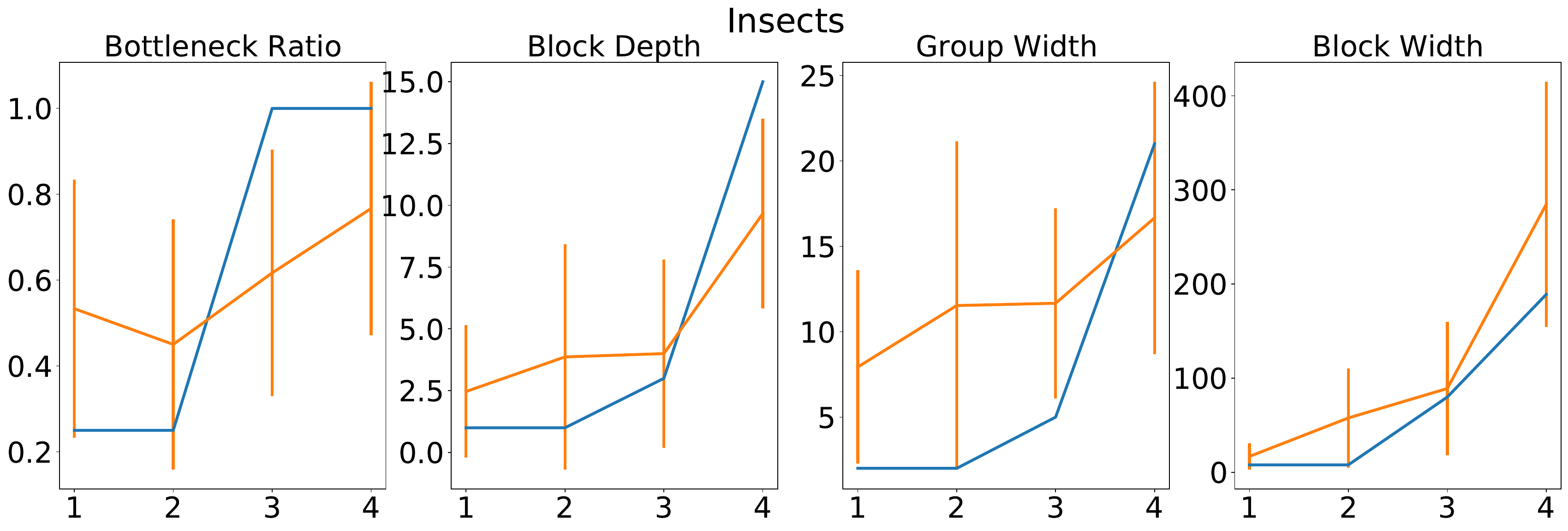}
    \includegraphics[width=0.49\textwidth]{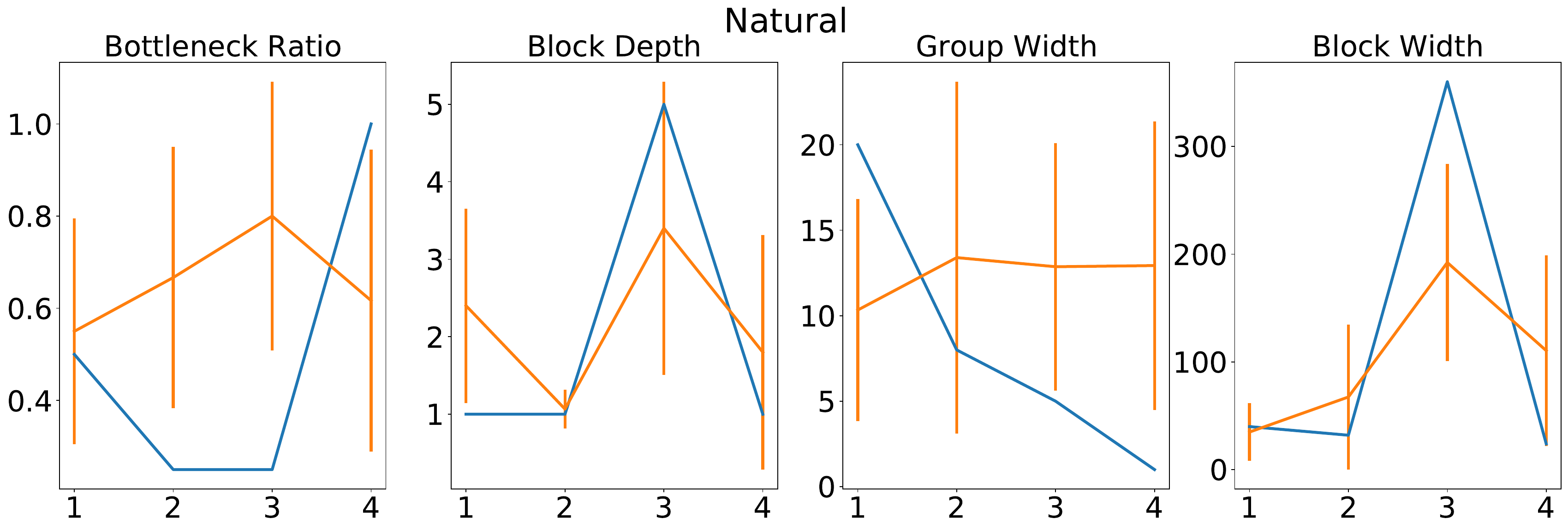} 
    \includegraphics[width=0.49\textwidth]{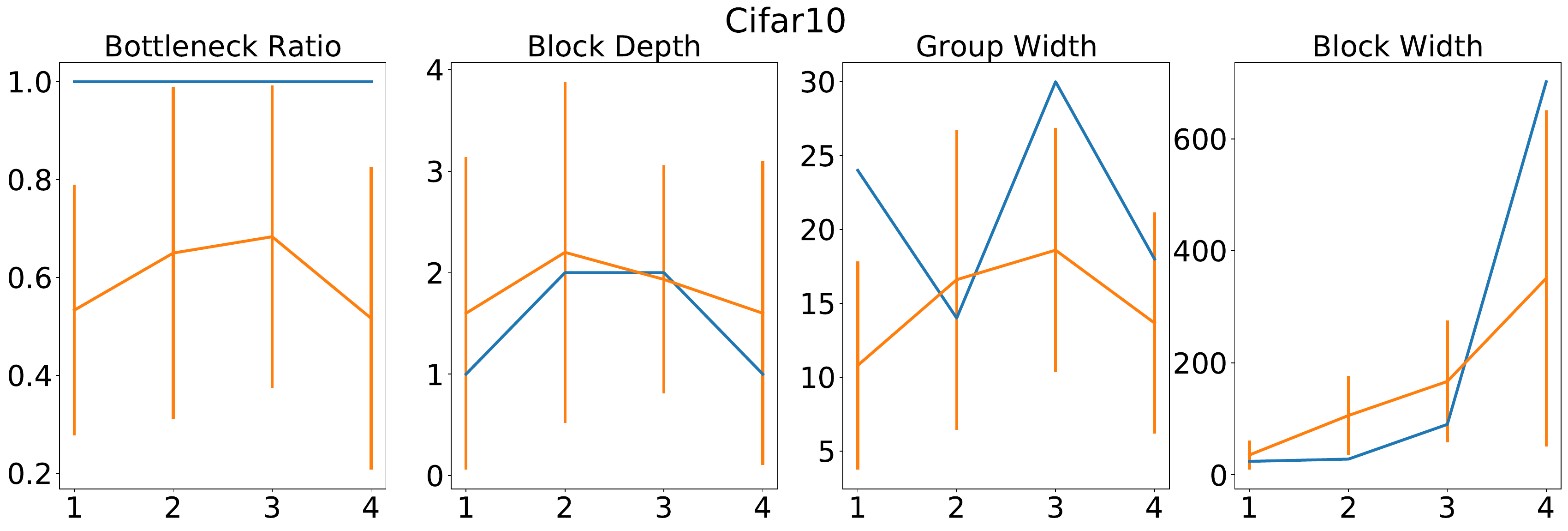} 
    \includegraphics[width=0.49\textwidth]{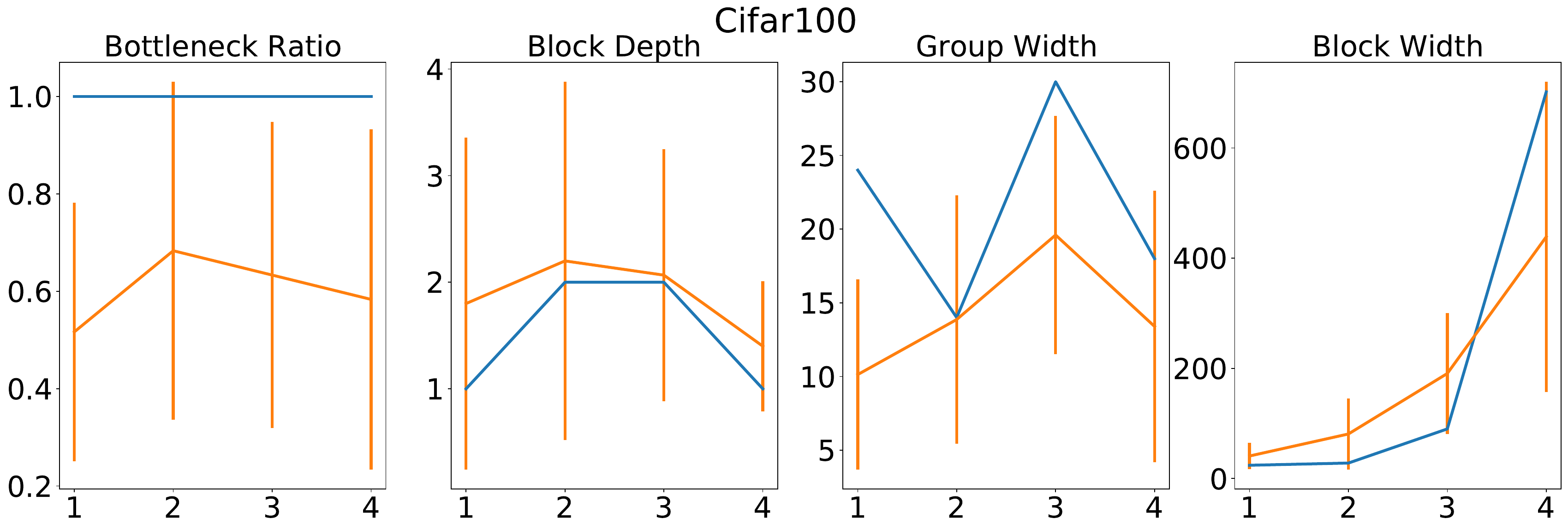} 
    \caption{Configurations of the top-performing architectures, with the four stages depicted on the x-axis and the parameter values on the y-axis. The best architectures are shown in blue, the mean of the top $15$ architectures is depicted in orange with with a vertical indication of one standard deviation.}
    \label{fig:top}
\end{figure*}

Figure \ref{fig:top} shows the configuration of the top performing architecture in blue, as well as the mean and standard deviation of the top $15$ configurations for every dataset. We observe that the top $15$ architectures have very high variance in both bottleneck ratio and group width. 

Block width on the other hand shows a clear pattern: almost all high-performing architectures start with a very small block width that increases across the stages. Only Powerline and Natural do not show this pattern. In block depth, we observe a similar pattern with a bit more noise. For block depth, Powerline, Natural, Cifar10 and Cifar100, no such trend of increased parameter values towards the later stages is observed. \emph{This reinforces the idea that block width and block depth greatly impact an architectures performance and their optimal choices are dataset dependent.}

\begin{figure*}[h!]
    \centering
    \includegraphics[width=1\textwidth]{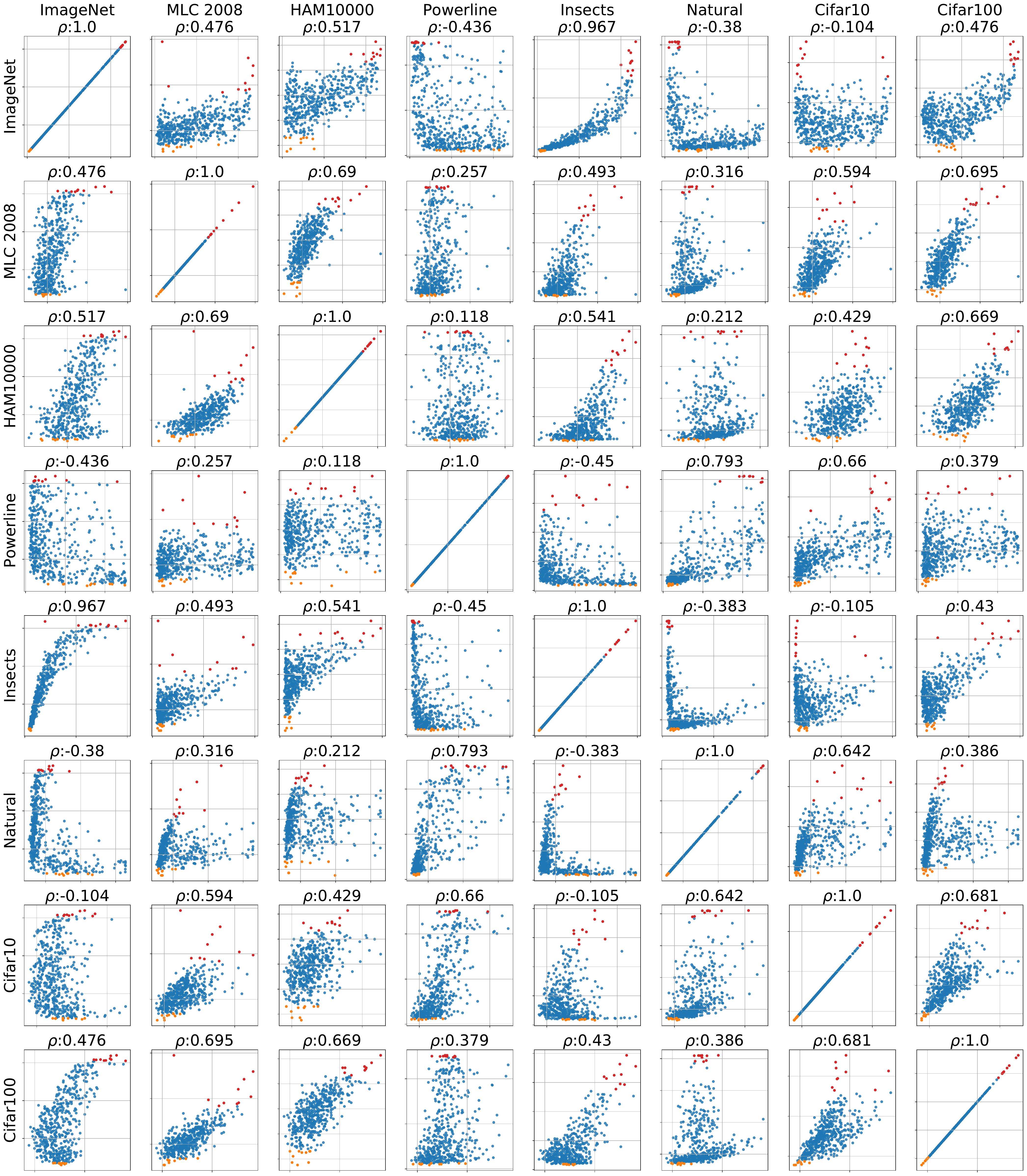}
    \caption{Matrix of error scatterplots of all datasets except Concrete (The first row replicates plots shown in Figure \ref{fig:scatter_performance}).}
    \label{fig:scatter_all}
\end{figure*}

\clearpage
\onecolumn
\section{Additional Material}
\label{chp:plot_matrices}

\subsection{Restricting Parameters to be Monotonically Increasing Along Stages}
\label{chp:design_param}

We have observed that architectures with increasing parameter values across successive stages have tend to perform better on ImageNet. We thus study the impact of restricting the architectures to having strictly increasing parameter values across the stages on the other datasets. We create four subsets of our architecture population by filtering it for ascending bottleneck ratio, block depth, group width and block width each individually.
\par
\begin{figure*}[]
    \centering
    \includegraphics[width=0.29\textwidth]{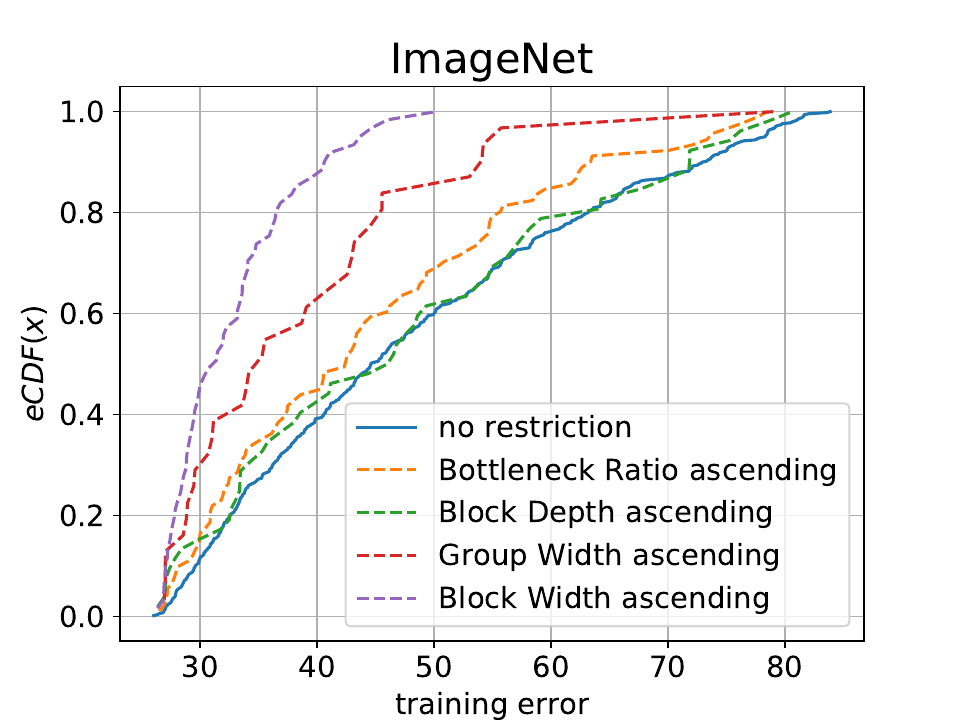} 
    \includegraphics[width=0.29\textwidth]{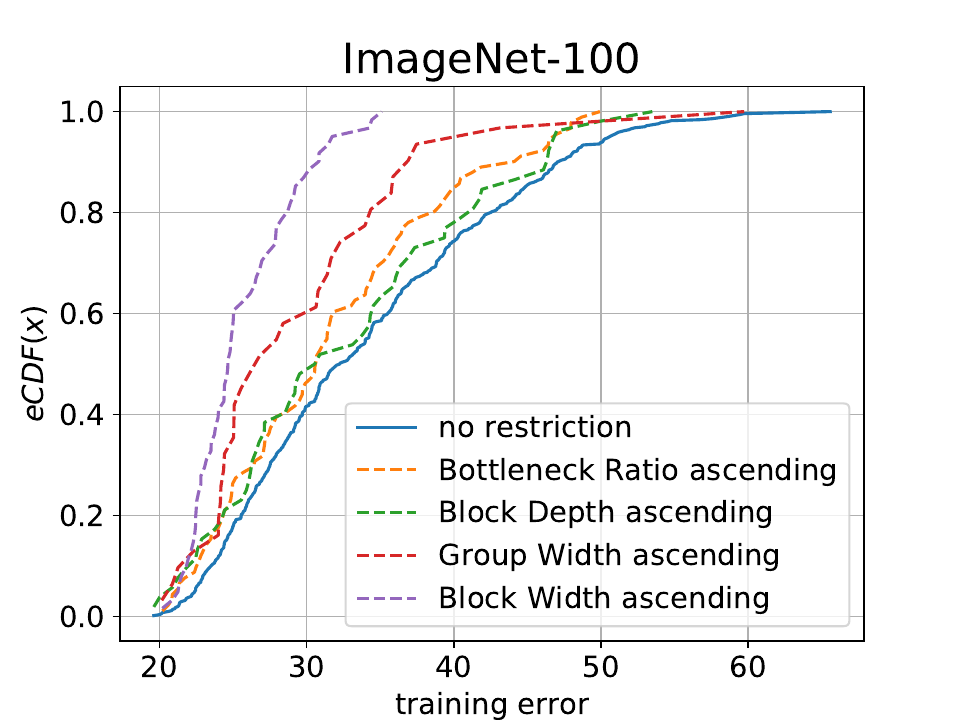} 
    \includegraphics[width=0.29\textwidth]{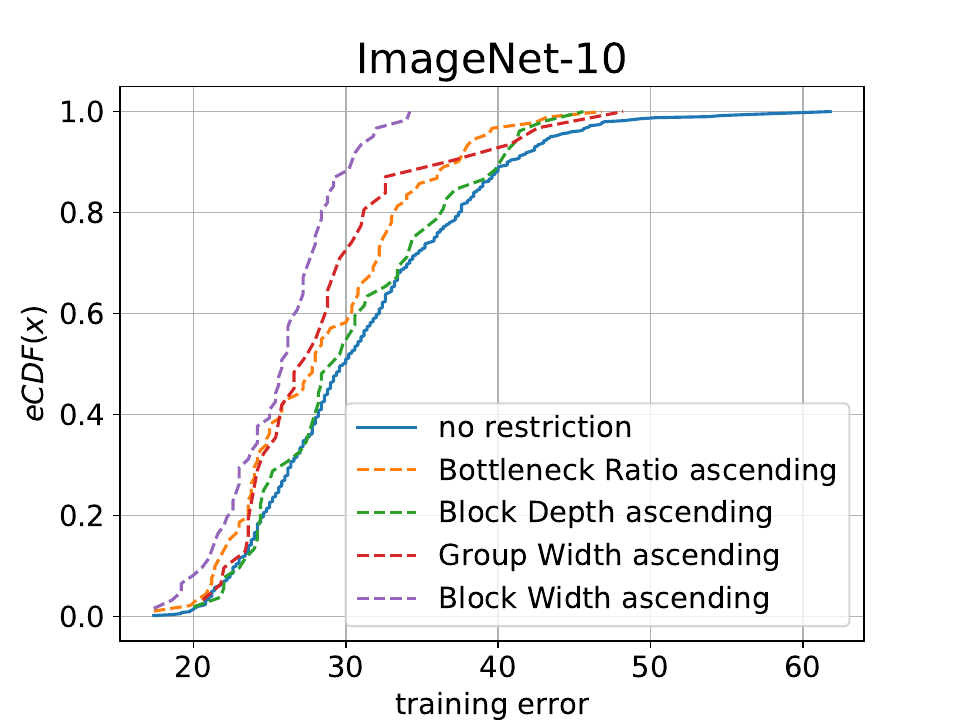} 
    \includegraphics[width=0.29\textwidth]{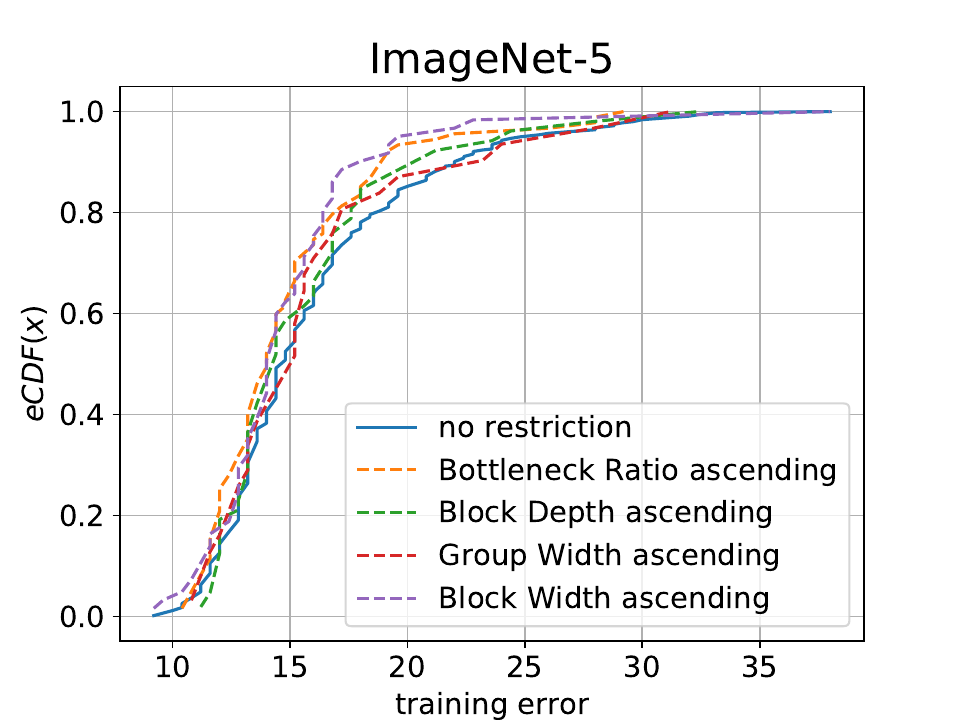} 
    \includegraphics[width=0.29\textwidth]{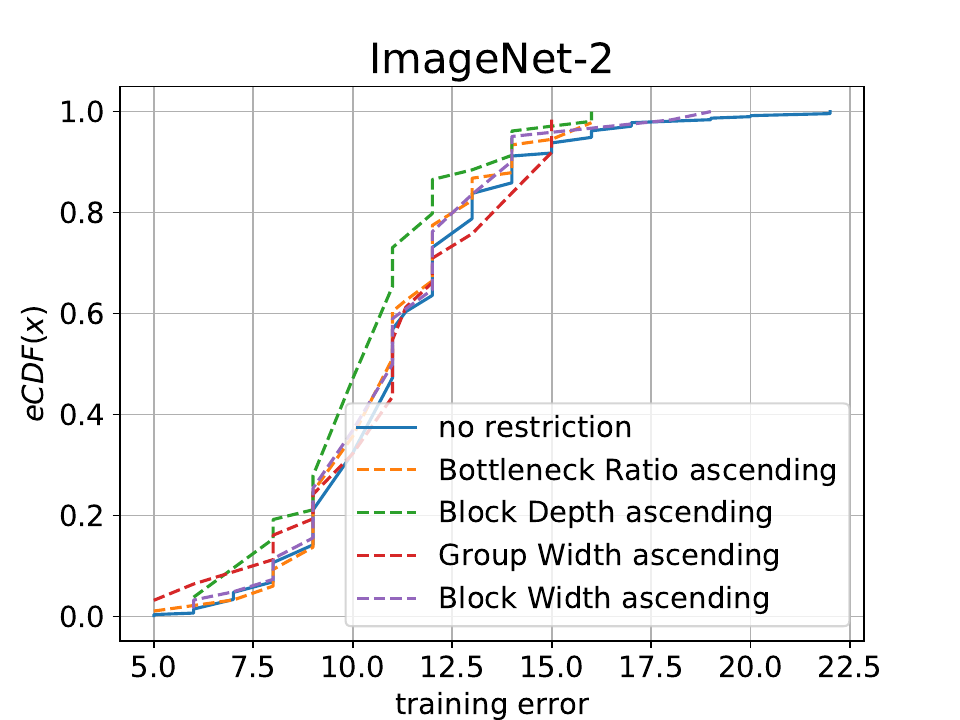} 
    \includegraphics[width=0.29\textwidth]{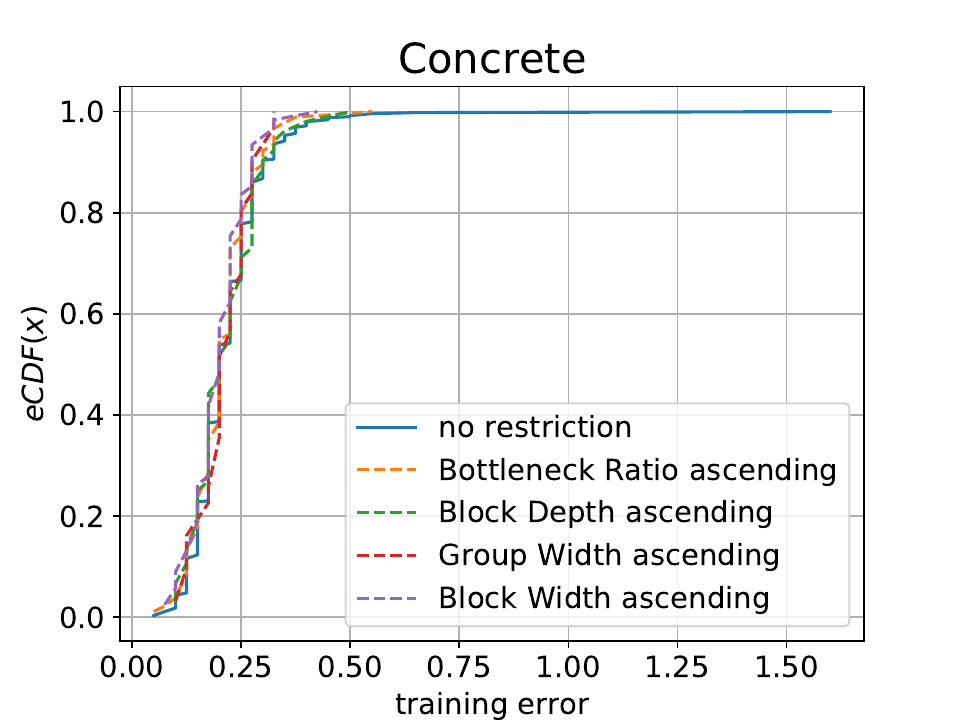} 
    \includegraphics[width=0.29\textwidth]{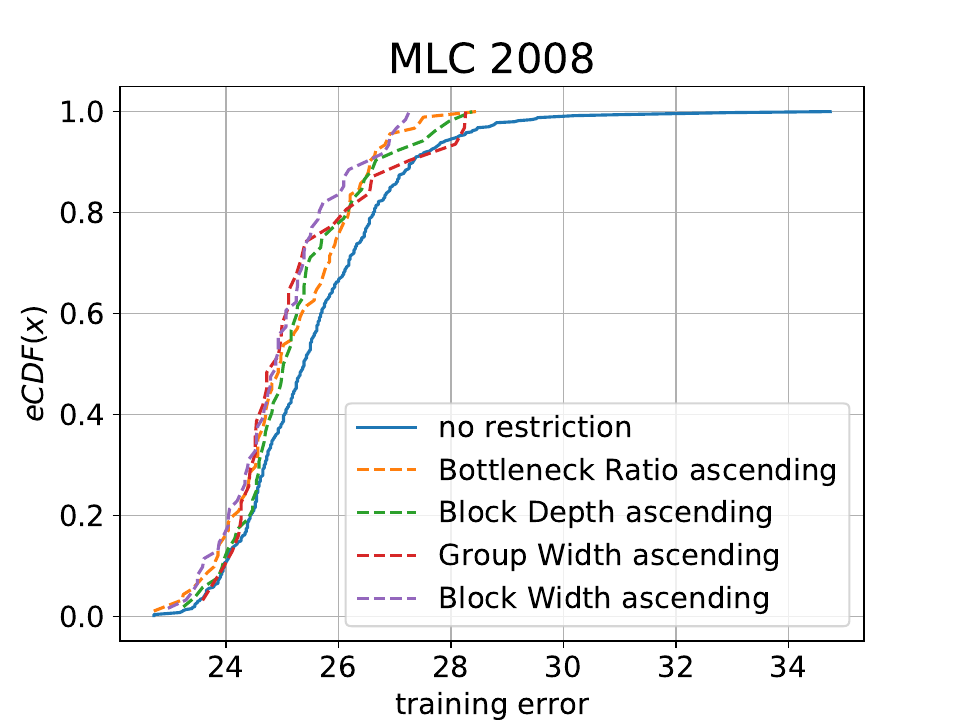} 
    \includegraphics[width=0.29\textwidth]{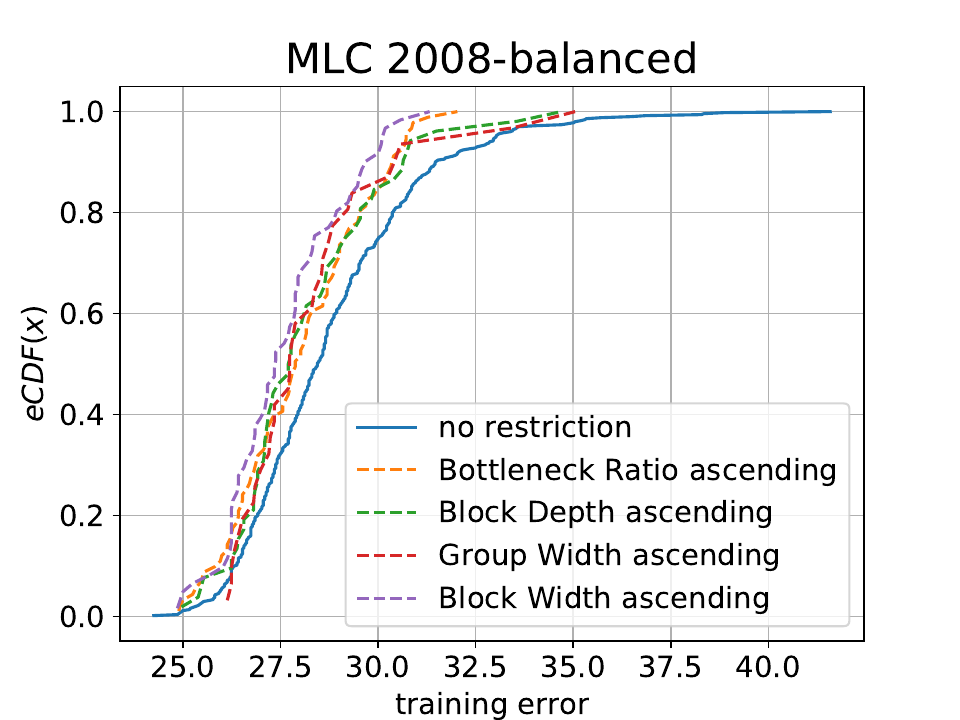} 
    \includegraphics[width=0.29\textwidth]{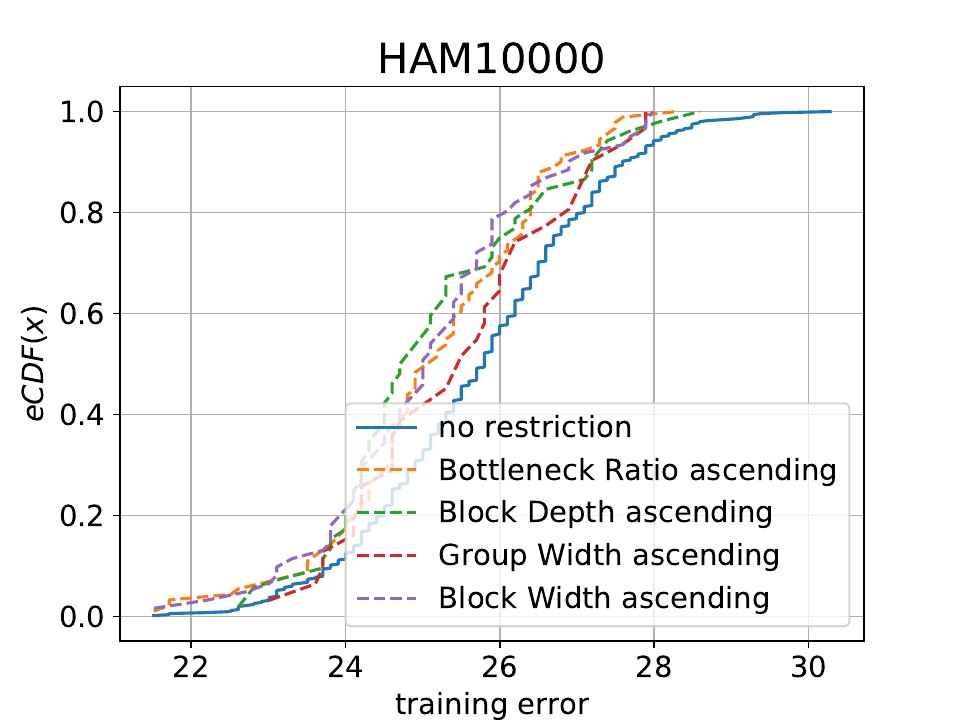} 
    \includegraphics[width=0.29\textwidth]{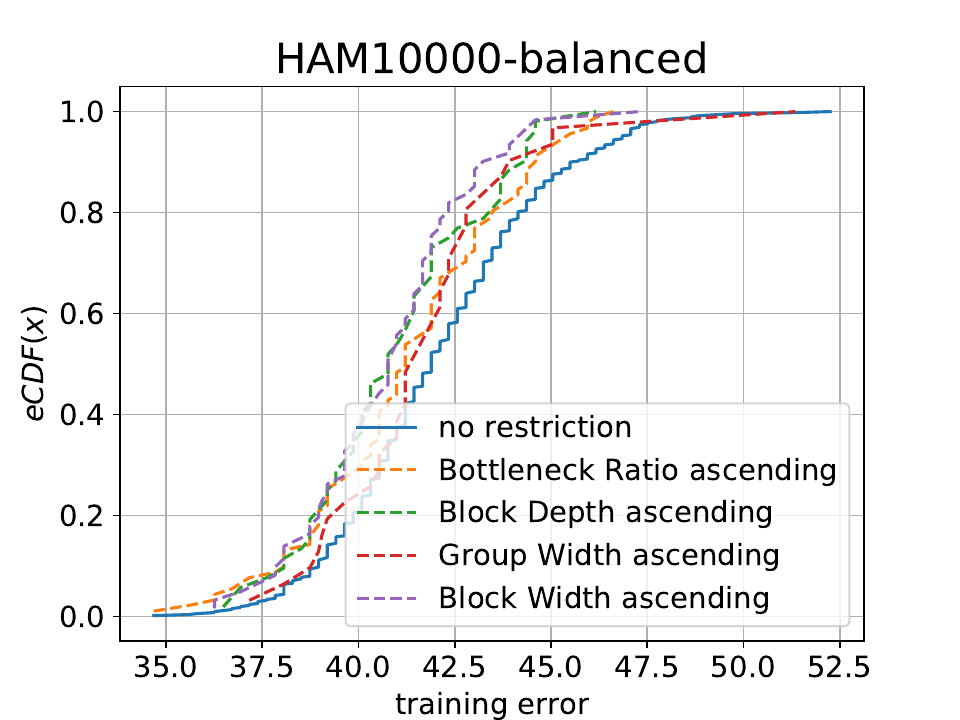} 
    \includegraphics[width=0.29\textwidth]{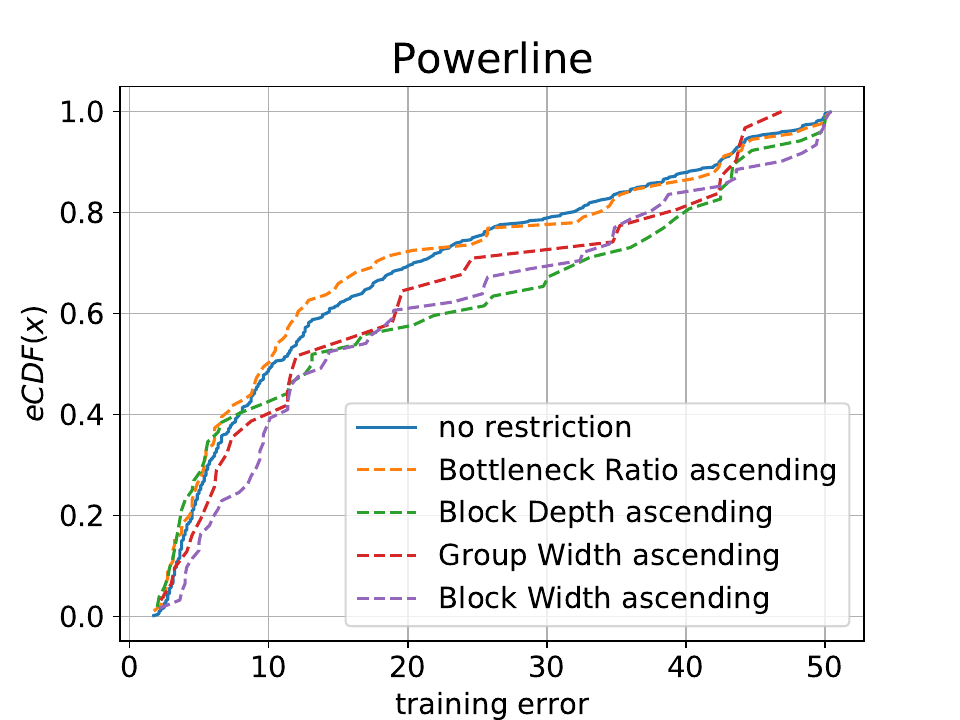} 
    \includegraphics[width=0.29\textwidth]{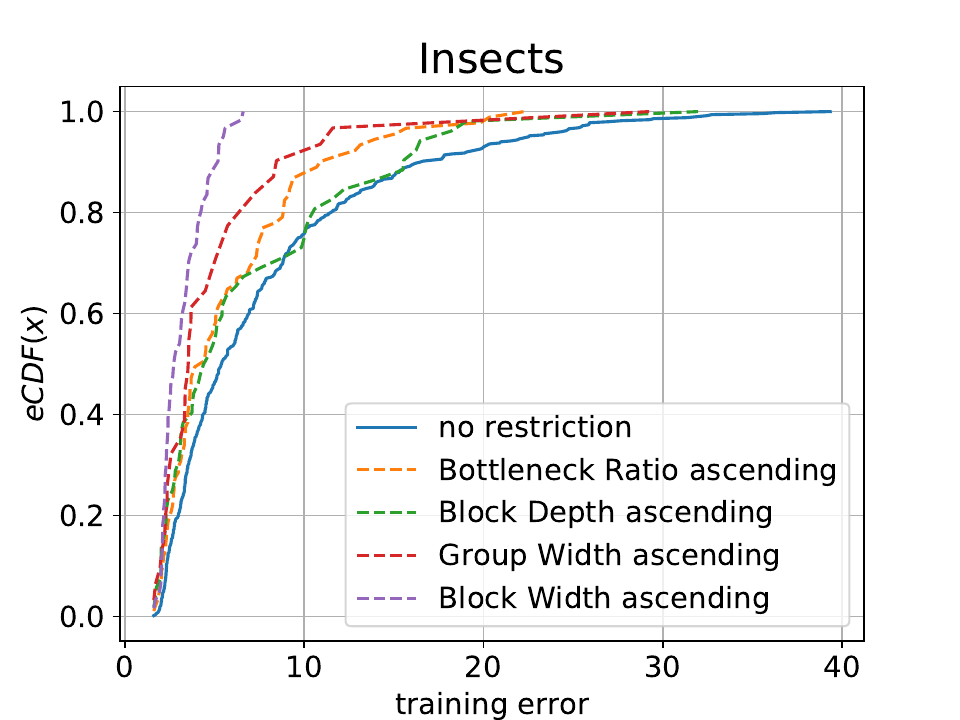} 
    \includegraphics[width=0.29\textwidth]{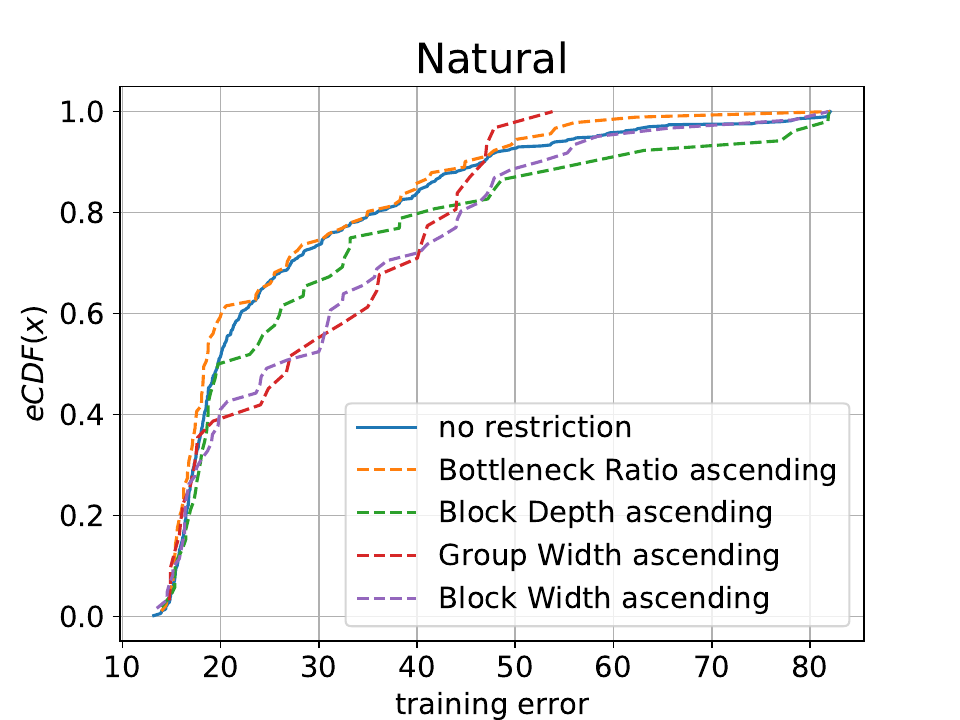} 
    \includegraphics[width=0.29\textwidth]{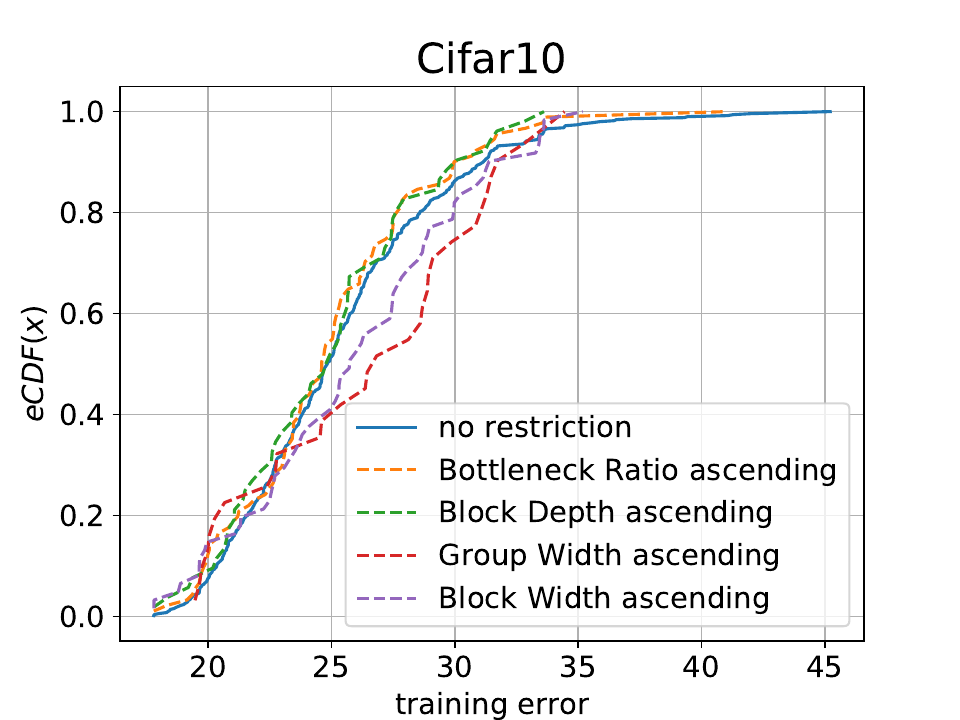} 
    \includegraphics[width=0.29\textwidth]{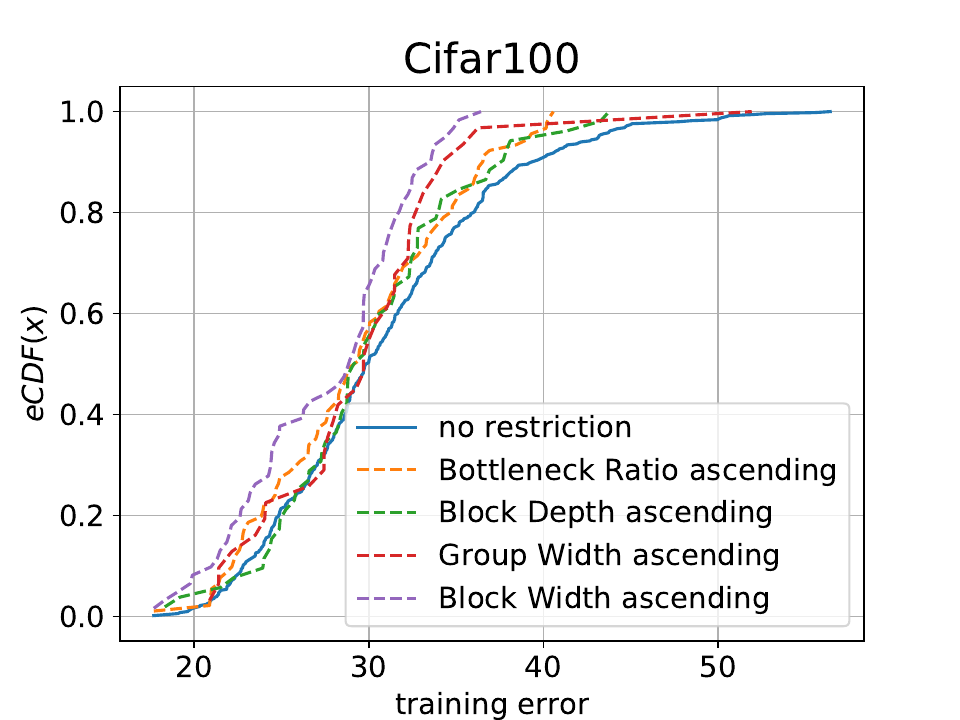} 
    \caption{The $eCDFs$ of the restricted sub-populations in the context of the $eCDF$ of the whole architecture population, for all datasets.}
    \label{fig:treatments}
\end{figure*}

Figure \ref{fig:treatments} shows the empirical cumulative density functions $eCDF$ of the whole pouplation as well as the $eCDFs$ of the restricted sub-populations for every dataset.
For most datasets the restrictions lead to a population with a better error distribution. The effect is most impactful for the models with ascending block width on ImageNet and Insects. Again, Powerline, Natural and to some extent Cifar10 are the odd ones where most restrictions lead to a worse overall performance, \emph{which is further evidence that architecture design is dataset dependent.}

\begin{figure*}[h!]
    \centering
    \includegraphics[width=\textwidth]{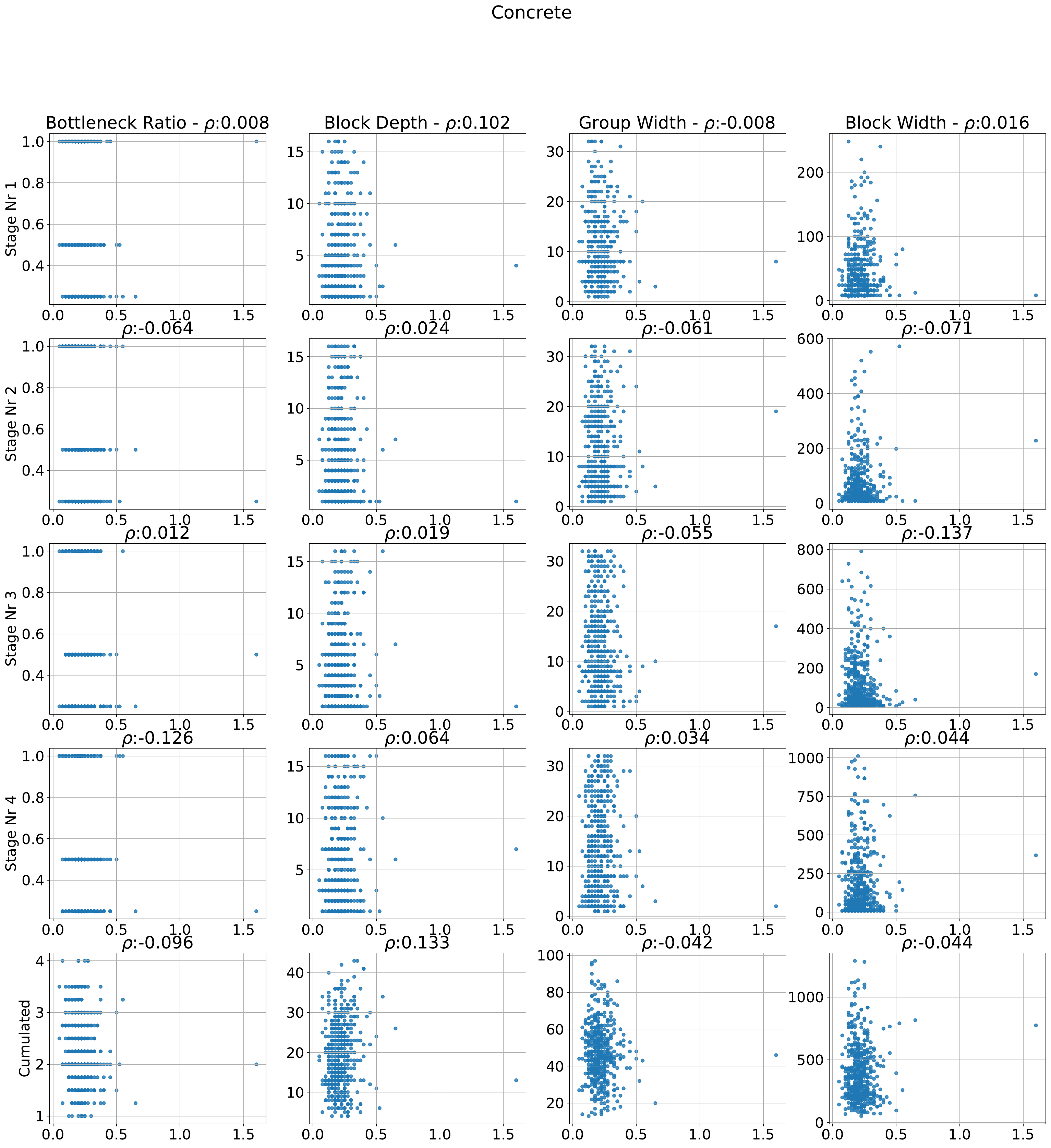}
    \caption{Individual parameter by stage versus error scatterplots for the Concrete dataset. The x-axis gives the error, while the parameter values are given on the ordinate.}
    \label{fig:concrete}
\end{figure*}

\begin{figure*}[h]
    \centering
    \includegraphics[width=\textwidth]{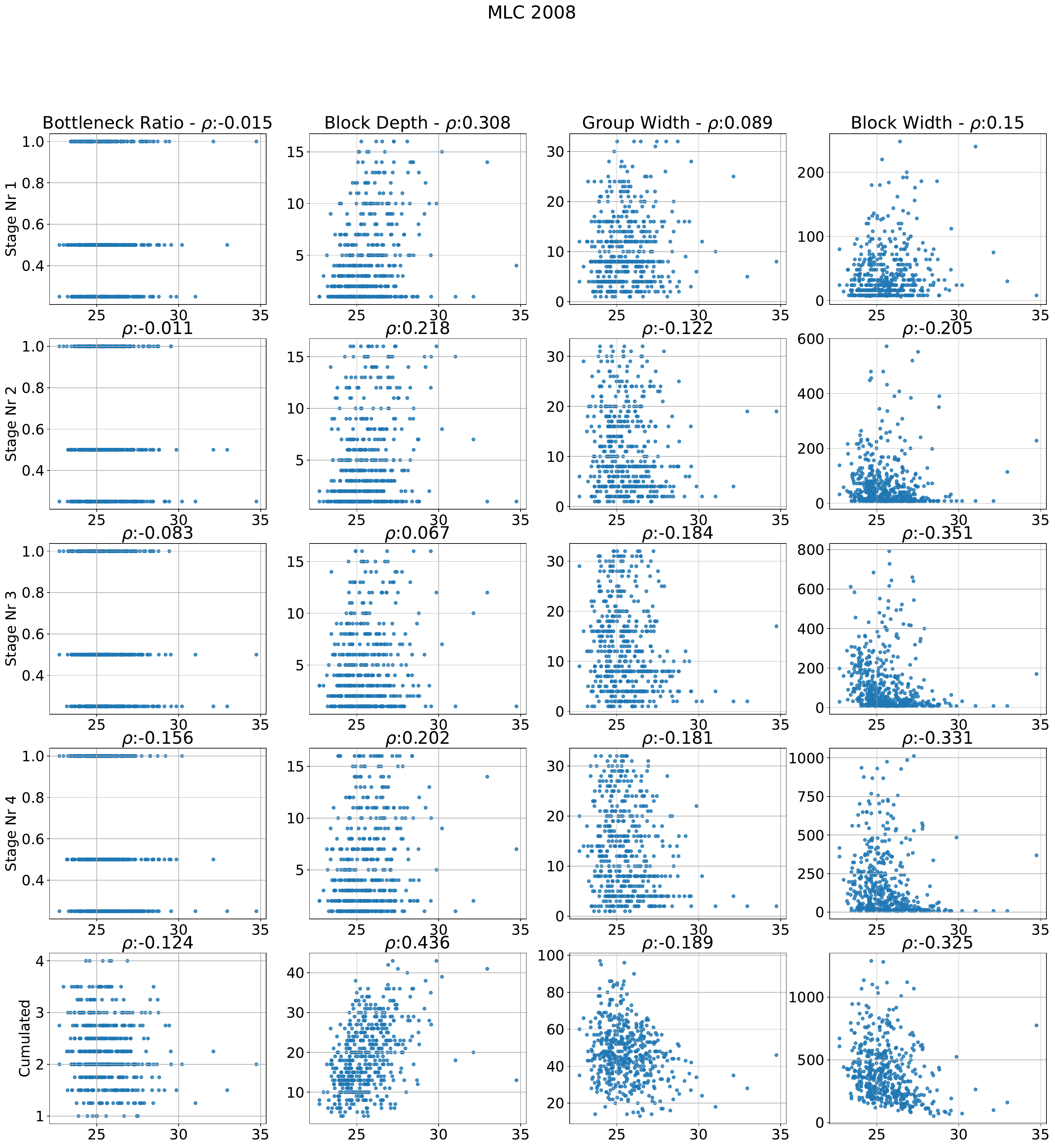}
    \caption{Individual parameter by stage versus error scatterplots for the MLC2008 dataset. The x-axis gives the error, while the parameter values are given on the ordinate.}
    \label{fig:MLC2008}
\end{figure*}
\begin{figure*}[h]
    \centering
    \includegraphics[width=\textwidth]{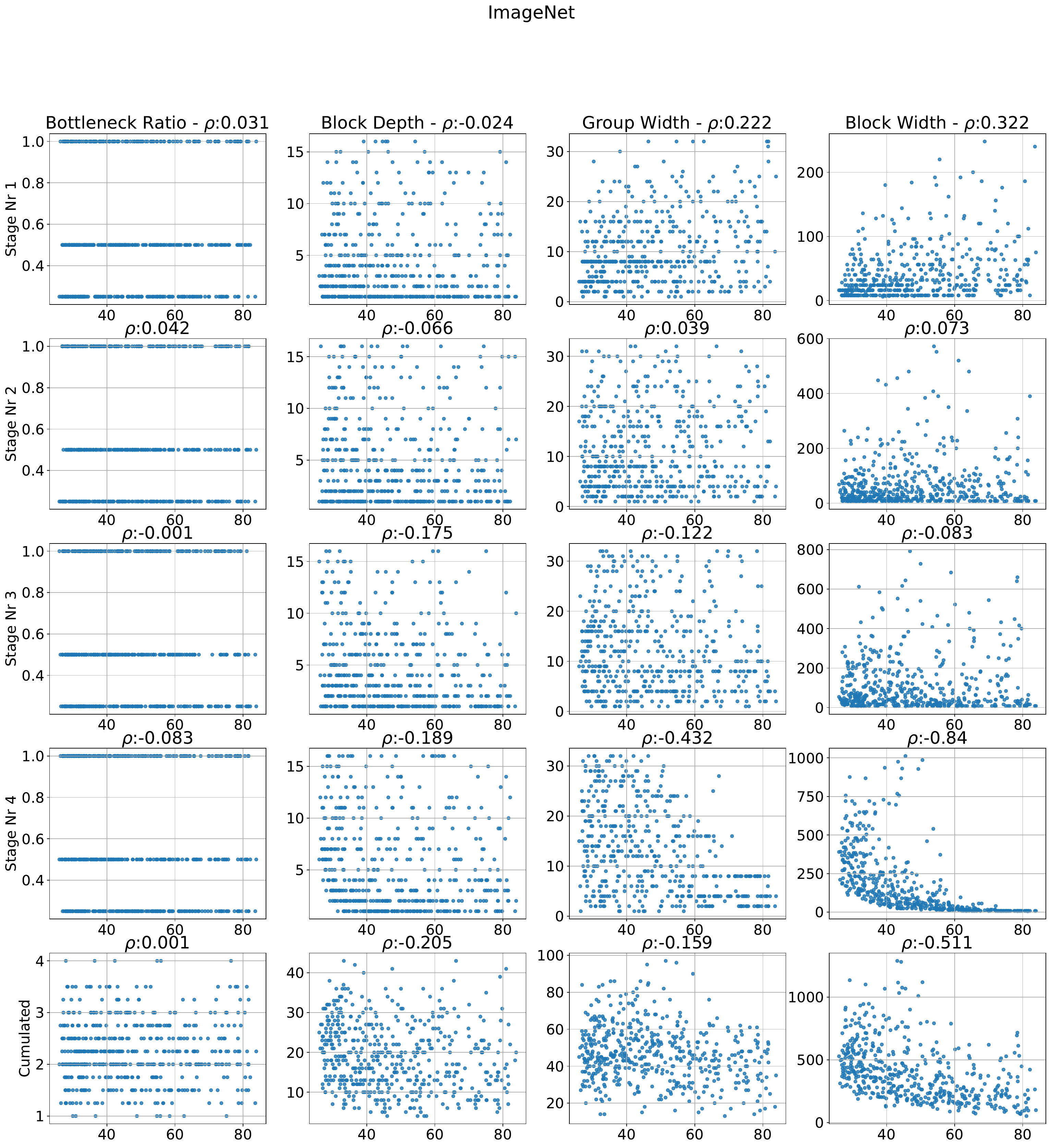}
        \caption{Individual parameter by stage versus error scatterplots for the ImageNet dataset. The x-axis gives the error, while the parameter values are given on the ordinate. The x-axis gives the error, while the parameter values are given on the ordinate.}
    \label{fig:imgnet}
\end{figure*}
\begin{figure*}[h]
    \centering
    \includegraphics[width=\textwidth]{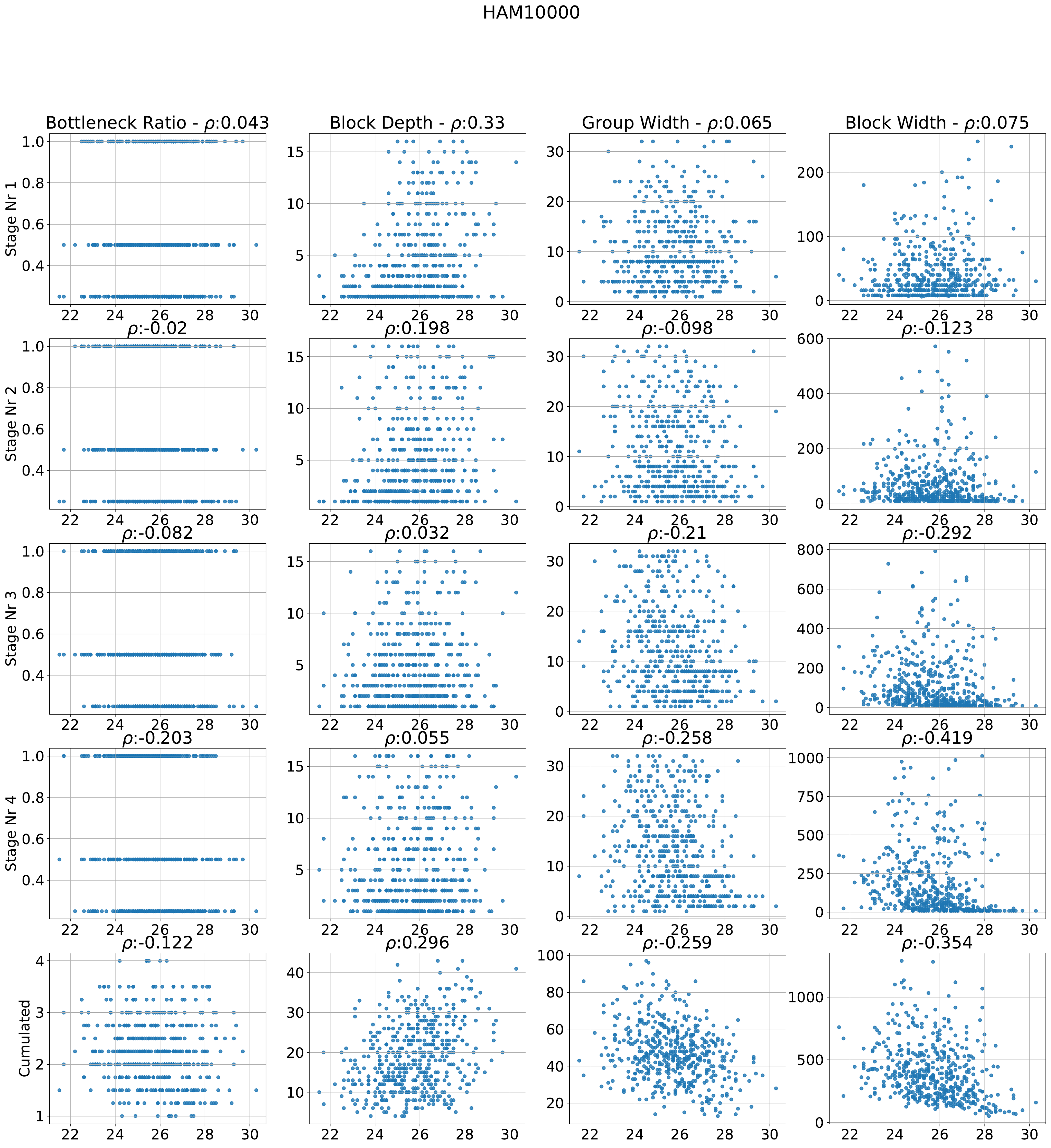}
        \caption{Individual parameter by stage versus error scatterplots for the HAM10000 dataset. The x-axis gives the error, while the parameter values are given on the ordinate.}
    \label{fig:ham}
\end{figure*}
\begin{figure*}[h]
    \centering
    \includegraphics[width=\textwidth]{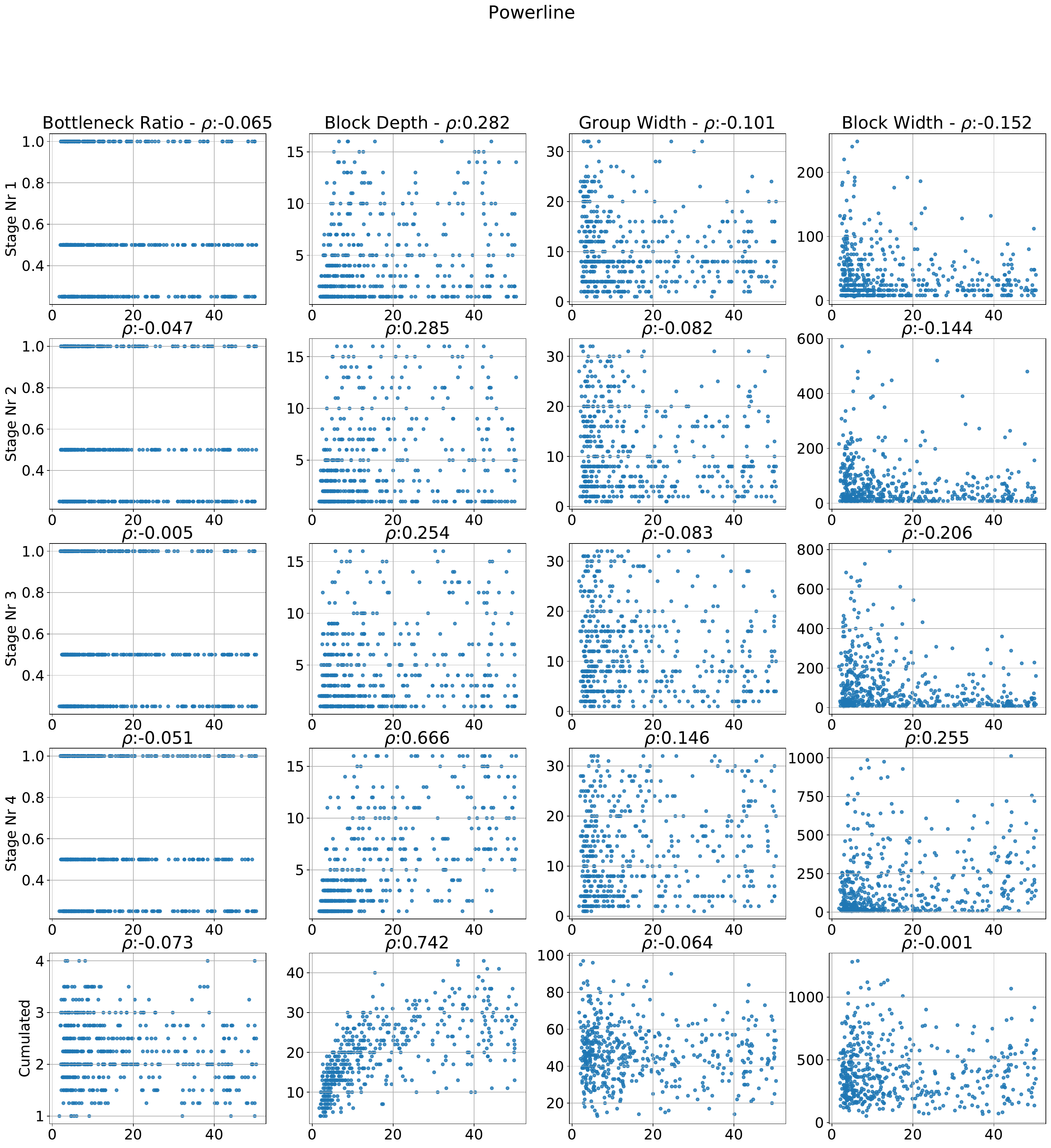}
        \caption{Individual parameter by stage versus error scatterplots for the Powerline dataset. The x-axis gives the error, while the parameter values are given on the ordinate.}
    \label{fig:pow}
\end{figure*}
\begin{figure*}[h]
    \centering
    \includegraphics[width=\textwidth]{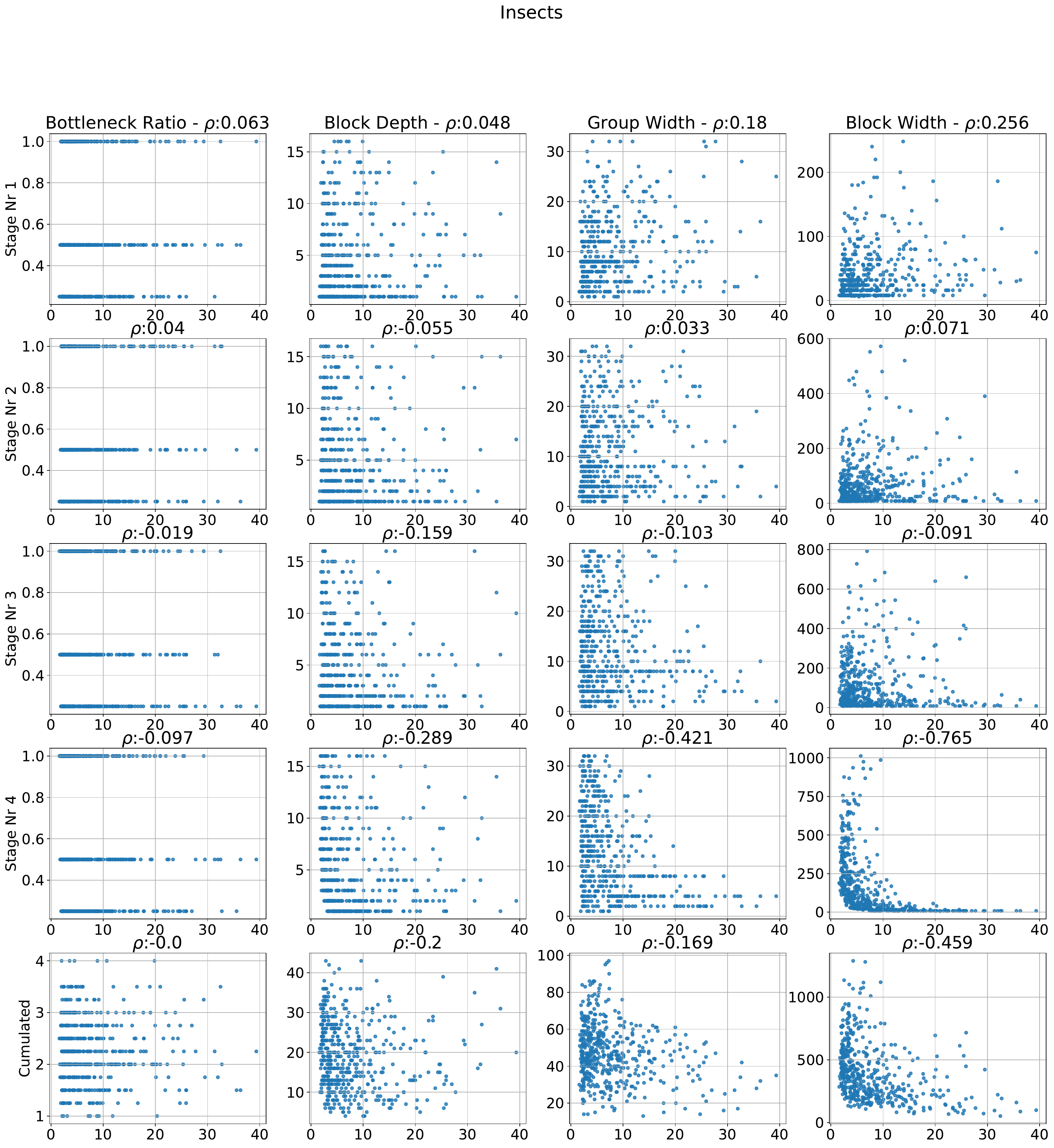}
        \caption{Individual parameter by stage versus error scatterplots for the Insects dataset. The x-axis gives the error, while the parameter values are given on the ordinate.}
    \label{fig:insects}
\end{figure*}
\begin{figure*}[h]
    \centering
    \includegraphics[width=\textwidth]{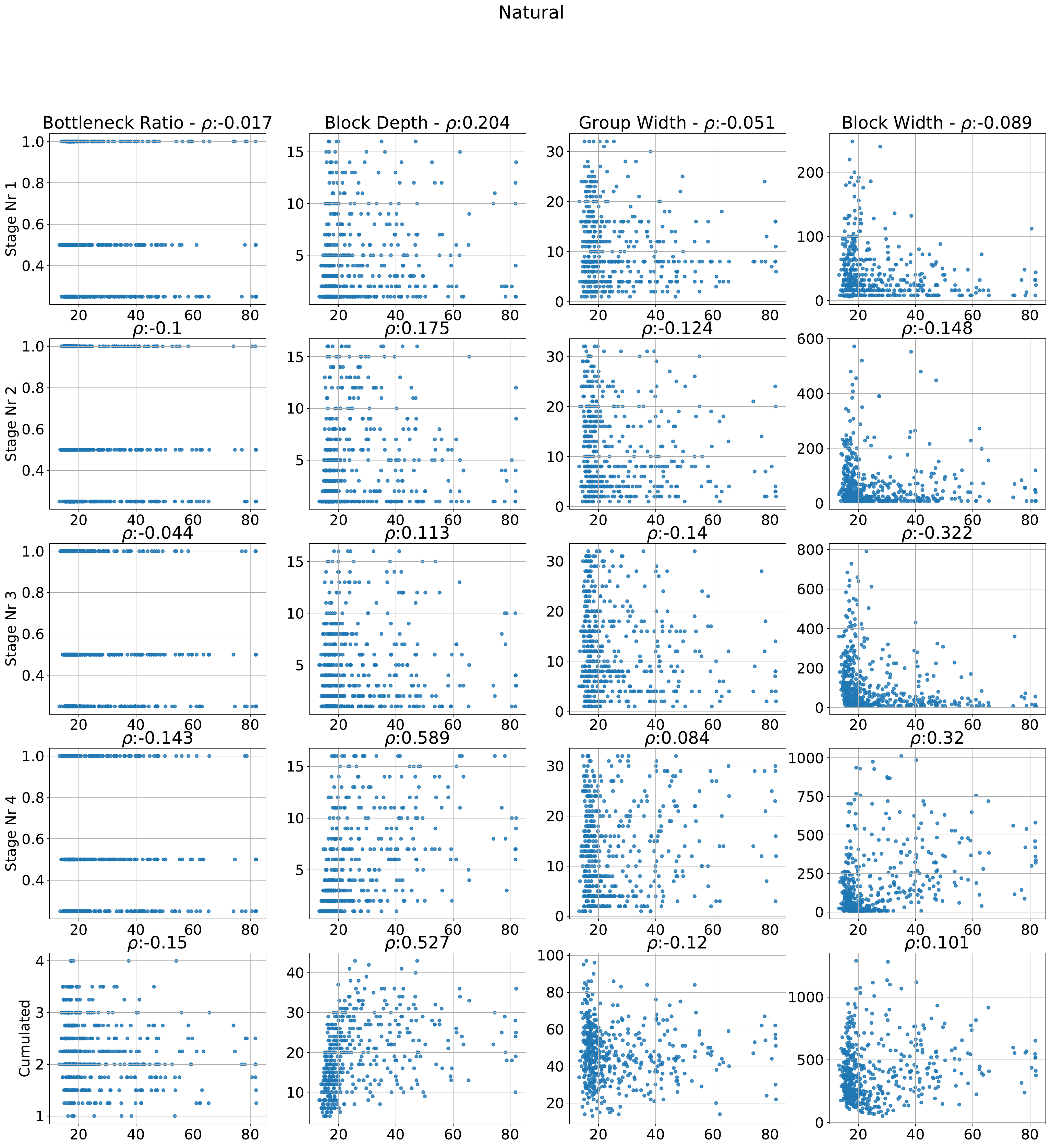}
        \caption{Individual parameter by stage versus error scatterplots for the Natural dataset. The x-axis gives the error, while the parameter values are given on the ordinate.}
    \label{fig:natural}
\end{figure*}
\begin{figure*}[h]
    \centering
    \includegraphics[width=\textwidth]{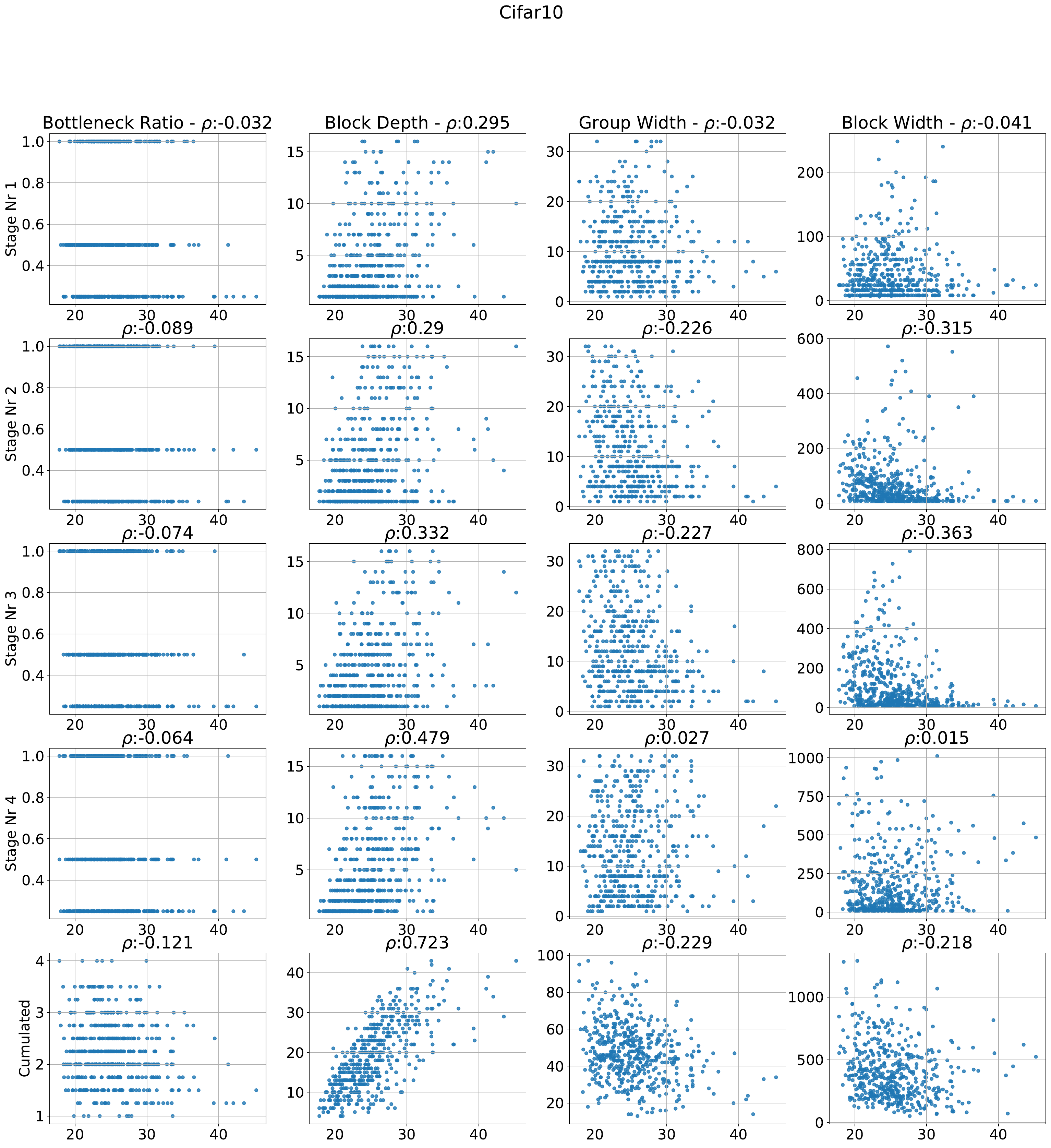}
        \caption{Individual parameter by stage versus error scatterplots for the Cifar10 dataset. The x-axis gives the error, while the parameter values are given on the ordinate.}
    \label{fig:cifar10}
\end{figure*}
\begin{figure*}[h]
    \centering
    \includegraphics[width=\textwidth]{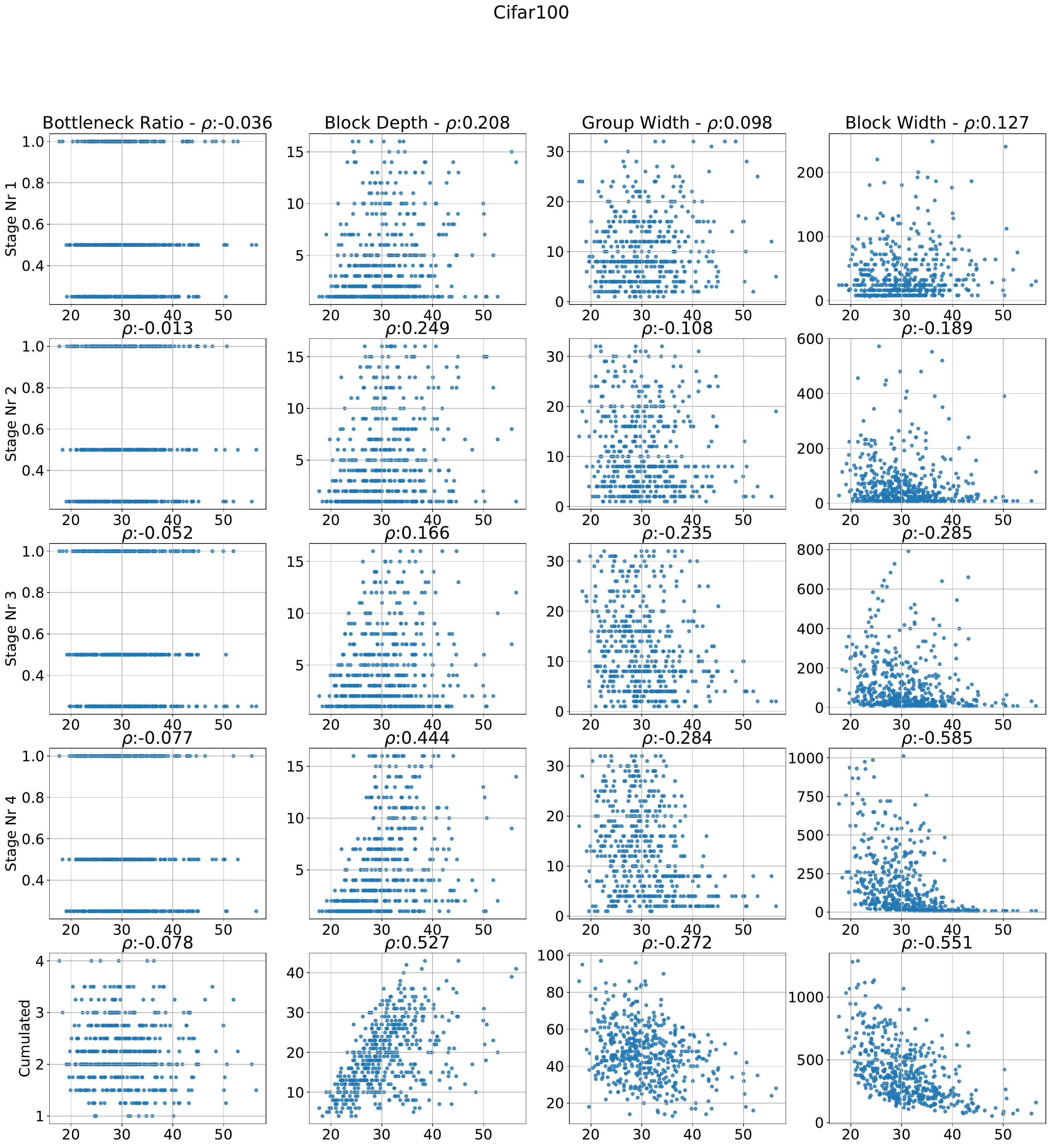}
        \caption{Individual parameter by stage versus error scatterplots for the Cifar100 dataset. The x-axis gives the error, while the parameter values are given on the ordinate.}
    \label{fig:cifar100}
\end{figure*}
\begin{figure*}[h]
    \centering
    \includegraphics[width=\textwidth]{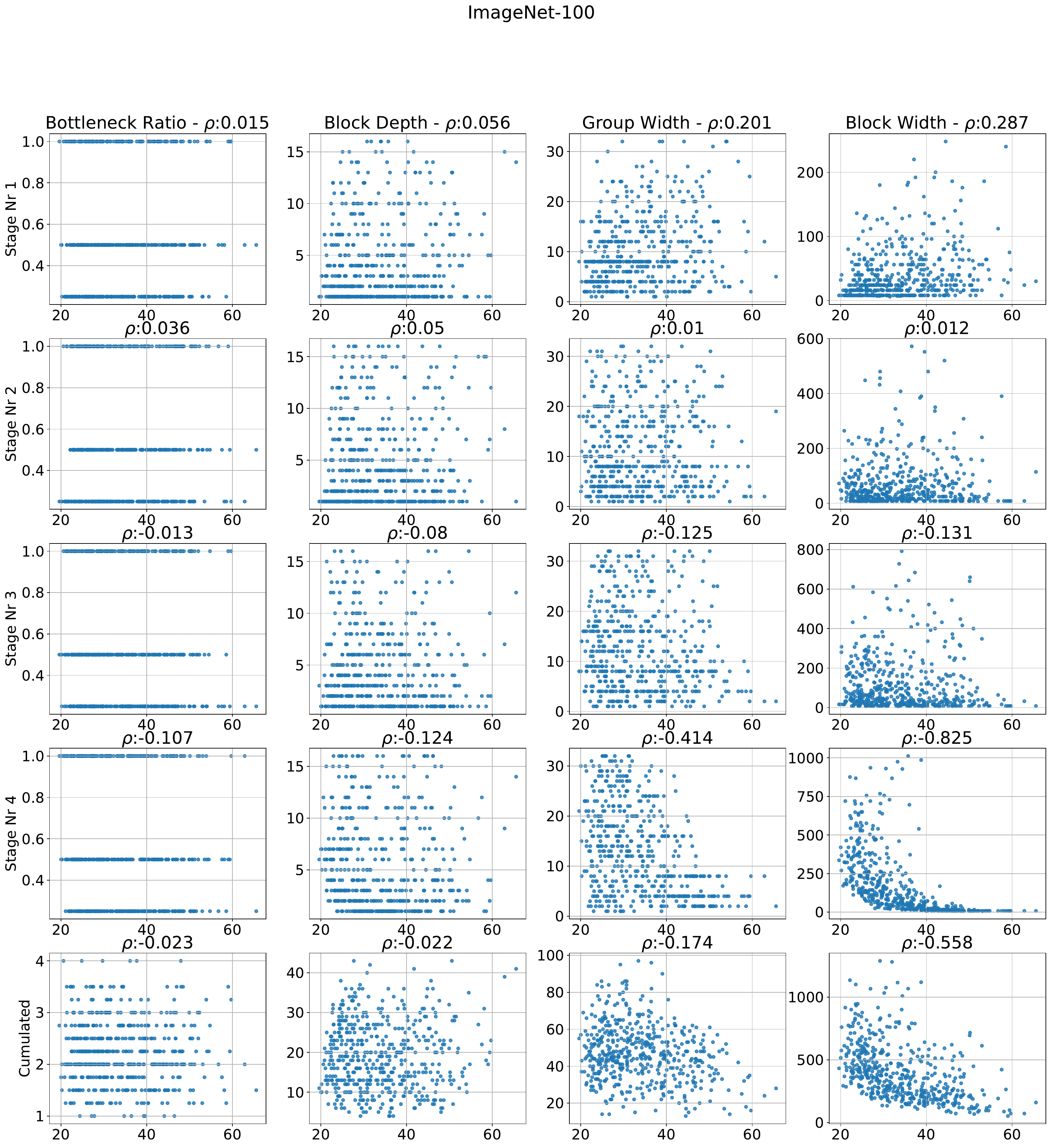}
        \caption{Individual parameter by stage versus error scatterplots for the ImageNet-100 dataset. The x-axis gives the error, while the parameter values are given on the ordinate.}
    \label{fig:imgnet100}
\end{figure*}
\begin{figure*}[h]
    \centering
    \includegraphics[width=\textwidth]{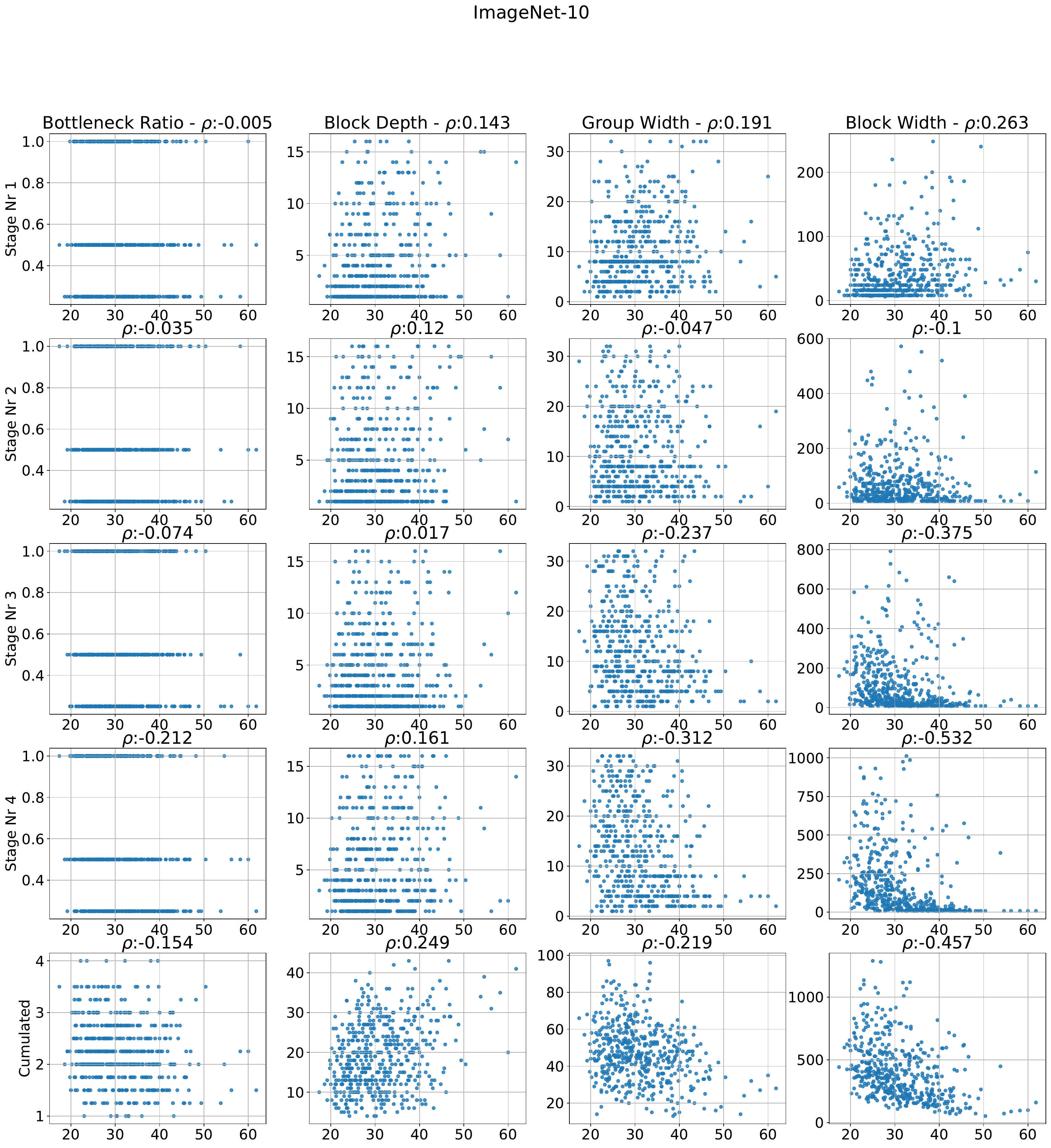}
        \caption{Individual parameter by stage versus error scatterplots for the ImageNet-10 dataset. The x-axis gives the error, while the parameter values are given on the ordinate.}
    \label{fig:imgnet10}
\end{figure*}
\begin{figure*}[h]
    \centering
    \includegraphics[width=\textwidth]{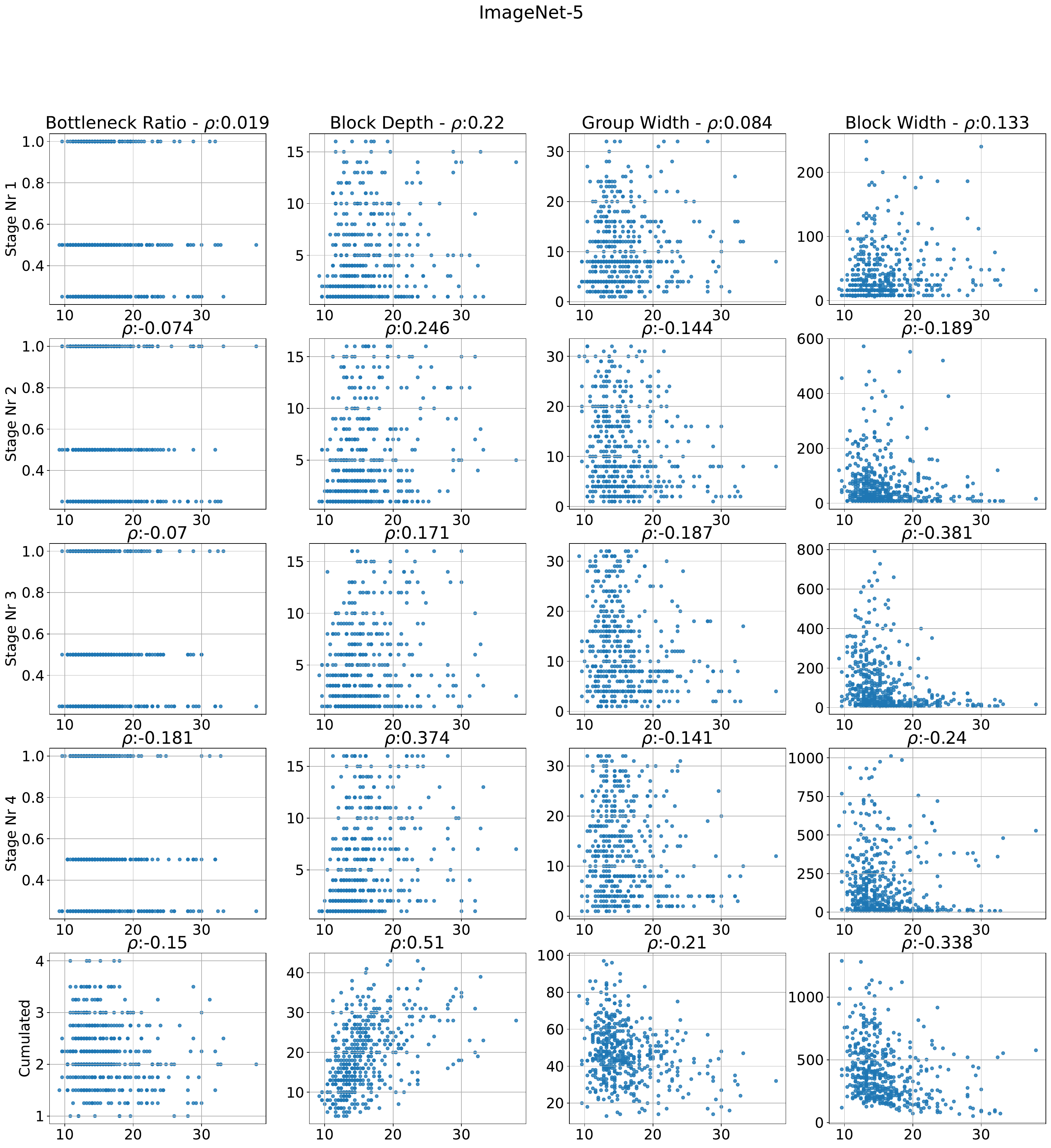}
        \caption{Individual parameter by stage versus error scatterplots for the ImageNet-5 dataset. The x-axis gives the error, while the parameter values are given on the ordinate.}
    \label{fig:imgnet5}
\end{figure*}
\begin{figure*}[h]
    \centering
    \includegraphics[width=\textwidth]{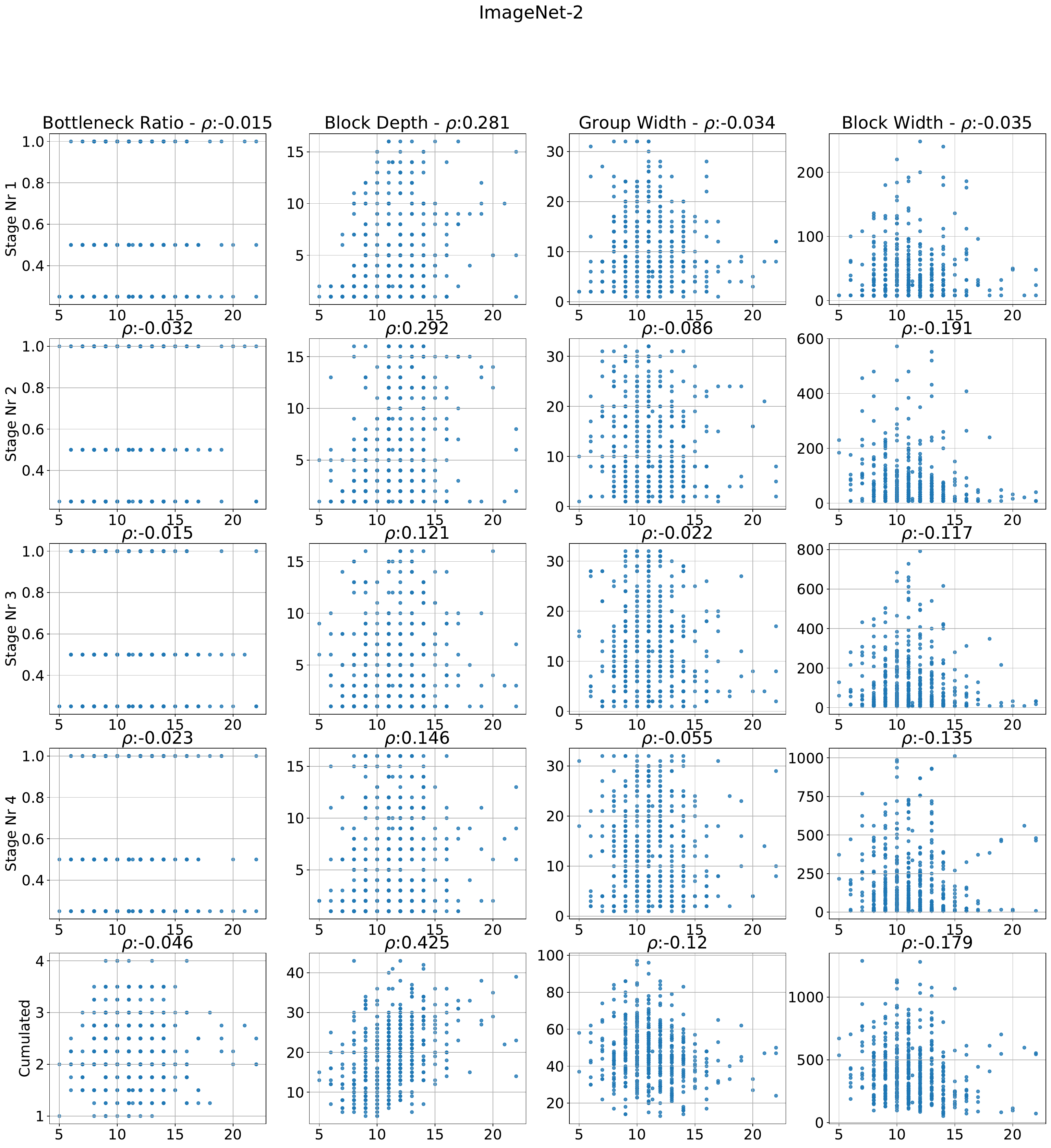}
        \caption{Individual parameter by stage versus error scatterplots for the ImageNet-2 dataset. The x-axis gives the error, while the parameter values are given on the ordinate.}
    \label{fig:imgnet2}
\end{figure*}
\end{appendices}

\end{document}